%% file: main.tex
\documentclass[10pt,twocolumn,letterpaper]{article}

\usepackage[pagenumbers]{cvpr} 

\input{preamble}

\definecolor{cvprblue}{rgb}{0.21,0.49,0.74}
\usepackage[pagebackref,breaklinks,colorlinks,citecolor=cvprblue]{hyperref}

\def\paperID{143} 
\def\confName{3DV\xspace}
\def\confYear{2024\xspace}

\title{Split, Merge, and Refine: Fitting Tight Bounding Boxes via\\Over-Segmentation and Iterative Search}

\author{Chanhyeok Park $\quad$
Minhyuk Sung \\[0.2em]
KAIST \\
{\tt\small \{chpark1111,mhsung\}@kaist.ac.kr}
}

\begin{document}

\twocolumn[{%
\renewcommand\twocolumn[1][]{#1}%
\newcommand{\rulesep}{\unskip\ \vrule\ }
\maketitle
\centering
\captionsetup{type=figure}
\begin{subfigure}[t]{0.49\textwidth}
    \begin{tabular}{>{\centering\arraybackslash}m{1.0\textwidth}}
        \includegraphics[width=\linewidth]{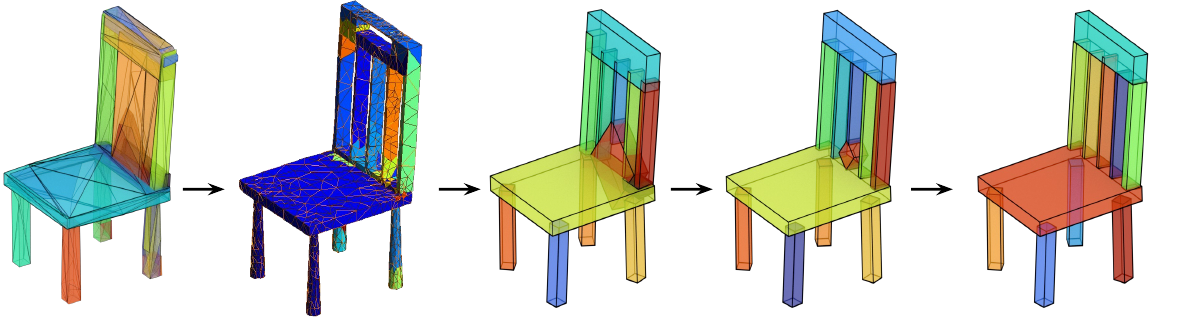} \\
        \includegraphics[width=\linewidth]{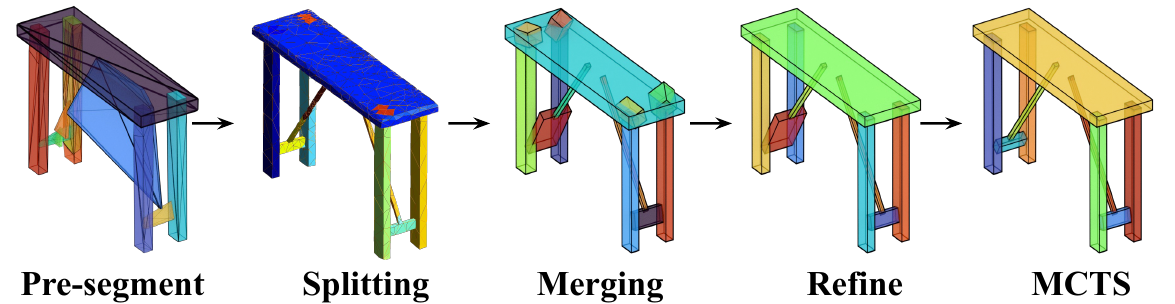} 
    \end{tabular}
    \caption{SMART Overview.}
\end{subfigure}
\hfill
\rulesep
\begin{subfigure}[t]{0.49\textwidth}
    \begin{tabular}{>{\centering\arraybackslash}m{1.0\textwidth}}
        \includegraphics[width=\linewidth]{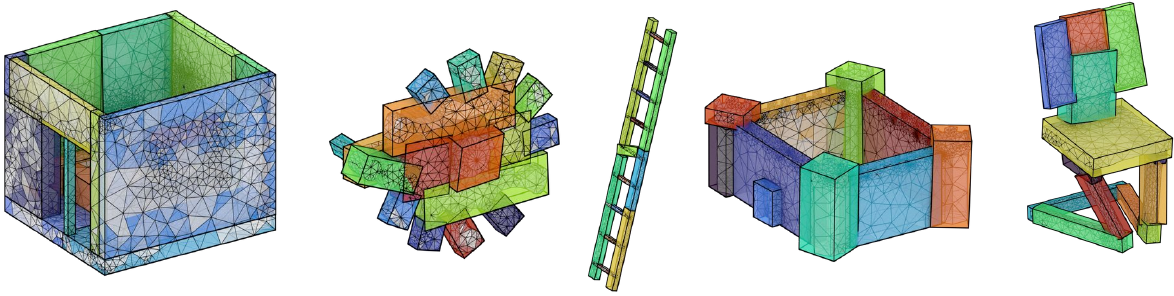} \\
        \includegraphics[width=\linewidth]{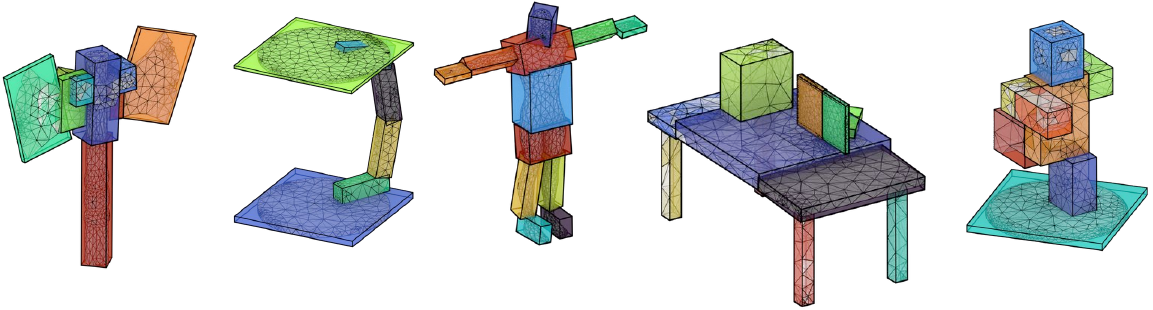} 
    \end{tabular}
    \caption{Results on Objaverse~\cite{objaverse} 3D models.}
\end{subfigure}
\caption{Our framework dubbed~\textbf{\smart{}} finds a set of tight bounding boxes of a 3D shape. (a) Given a 3D shape represented as a volumetric mesh, it first takes \textbf{over-segmentation} by post-processing (second column) any pre-segmentation (first column) and then performs hierarchical \textbf{merging} with tightness-aware criteria (third column). Then, it \textbf{refines} the bounding box parameters with a sequence of actions in a discrete space (fourth column). The results are further improved with long-sighted exploration by our \textbf{MCTS} accelerated with our technique (last column). (b) Results of~\smart{} on Objaverse~\cite{objaverse} overlaid with the input mesh represented as grey.}
\label{fig:teaser}
\vspace{0.5\baselineskip}
}]

\input {sections/0_Abstract}

\input {sections/1_Introduction}

\input{sections/2_Related_Work}

\input{sections/3_Problem_Definition_and_Overview}

\input{sections/4_Method}

\input{sections/5_Results}

\input{sections/6_Conclusion}

{
    \small
    \bibliographystyle{ieeenat_fullname}
    \bibliography{main}
}

\renewcommand{\thesection}{A}
\renewcommand{\thetable}{A\arabic{table}}
\renewcommand{\thefigure}{A\arabic{figure}}

\newif\ifpaper
\papertrue

\clearpage
\newpage

\section*{Appendix}
\input {sections/Supplementary}





\end{document}

%% file: preamble.tex
\usepackage[dvipsnames]{xcolor}
\usepackage[linesnumbered,ruled,vlined]{algorithm2e}
\usepackage{times}
\usepackage{epsfig}
\usepackage{graphicx}
\usepackage{amsmath}
\usepackage{amssymb}
\usepackage{algpseudocode}
\usepackage{adjustbox}
\usepackage{xcolor}
\usepackage{multirow}
\usepackage{makecell} 
\usepackage{booktabs}
\usepackage{tabularx}
\usepackage[font=small]{caption}
\usepackage{amsmath}
\usepackage{enumitem}
\usepackage{float}
\usepackage{longtable}
\usepackage{filecontents,catchfile,pgffor}
\usepackage{dashrule}

\DeclareMathOperator*{\argmax}{arg\,max}
\DeclareMathOperator*{\argmin}{arg\,min}

\SetCommentSty{mycommfont}

\SetKwInput{KwInput}{Input}                
\SetKwInput{KwOutput}{Output}              

\newcommand{\smart}[0]{\textsc{SMART}}


%% file: sections/0_Abstract.tex
\begin{abstract}
\vspace{-\baselineskip}
Achieving tight bounding boxes of a shape while guaranteeing complete boundness is an essential task for efficient geometric operations and unsupervised semantic part detection. But previous methods fail to achieve both full coverage and tightness. Neural-network-based methods are not suitable for these goals due to the non-differentiability of the objective, while classic iterative search methods suffer from their sensitivity to the initialization. We propose a novel framework for finding a set of tight bounding boxes of a 3D shape via over-segmentation and iterative merging and refinement. Our result shows that utilizing effective search methods with appropriate objectives is the \textbf{key} to producing bounding boxes with both properties. We employ an existing pre-segmentation to \textbf{split} the shape and obtain over-segmentation. Then, we apply hierarchical \textbf{merging} with our novel tightness-aware merging and stopping criteria. To overcome the sensitivity to the initialization, we also define actions to \textbf{refine} the bounding box parameters in an Markov Decision Process (MDP) setup with a soft reward function promoting a wider exploration. Lastly, we further improve the refinement step with Monte Carlo Tree Search \textbf{(MCTS)} based multi-action space exploration. By thoughtful evaluation on diverse 3D shapes, we demonstrate full coverage, tightness, and an adequate number of bounding boxes of our method without requiring any training data or supervision. It thus can be applied to various downstream tasks in computer vision and graphics.
\end{abstract}

%% file: sections/1_Introduction.tex
\vspace{-\baselineskip}

\section{Introduction}
\label{sec:introduction}
\vspace{-3pt}

Approximating complex 3D shapes using primitives offers several capabilities, including shape structure analysis, shape abstraction, and efficient geometric computations. To achieve these, many recent self-supervised learning approaches~\cite{tulsiani17abstract, paschalidou19sq, sun19abstract, yang21cubseg, paschalidou21np, chen20bspnet, niu22rimnet} have successfully addressed the problem while exploring different types of primitives.

Despite the recent advances, in this work, we pay attention to some desired yet underinvestigated properties of the bounding primitives of a shape and propose a novel approach aiming to achieve them. The properties are 1) \textbf{full coverage} --- guaranteeing the boundness of the entire shape by the primitives, 2) \textbf{tightness}, and 3) \textbf{adequate number of primitives}.
Attaining these three properties is particularly crucial for the downstream applications requiring \emph{efficient geometric computations}, such as intersection tests~\cite{collision_survey, orientedbbox}, robust transmissions~\cite{robust_transmission}, ray tracing~\cite{ellipsoids_intersection}, or proximity computations~\cite{oriented}. The primitives that just approximate but do not fully cover the shape can result in imprecise computation in such tasks. Also, loosely bounding primitives and too many primitives increase the computation time. Moreover, the bounding primitives satisfying these properties typically provide a better abstraction of the shape aligned with the \emph{human perception} of the shape decomposition.

Although neural-network-based methods have demonstrated their powerful generalizability and expressivity, they typically fail to achieve these three properties mainly due to the \emph{non-differentiable} nature of the objectives. The full coverage and tightness can be computed with the volumetric intersection or difference operations, which are not differentiable. Finding the proper number of parts is also a discrete problem that cannot be easily solved via backpropagation.


As tight bounding primitives have been essential in various applications, there also has been a line of work~\cite{robust_transmission, simari2005extraction, kalaiah2005statistical} addressing the problem before the deep learning era. One notable example is the work by Lu et al.~\cite{2007bvc} that proposes to find the tight bounding boxes by iteratively exchanging point-to-primitive assignments, starting from an initialization. The major drawback of such a method is its dependency and sensitivity to the \emph{initialization}, resulting in a suboptimal output when it starts from a poor initialization. 

To address the problem of achieving the three properties that introduce challenges of \emph{non-differentiability} for neural networks and \emph{initialization sensitivity} for iterative search methods, we propose a framework that performs \textbf{S}plitting, \textbf{M}erging, \textbf{A}nd \textbf{R}efinement \textbf{T}echniques, and is thus dubbed~\textbf{\smart{}}.
We first find that a simple post-processing applied to a pre-segmentation, can provide appropriate \textbf{over-segmentation} of a 3D shape.
Hence, we perform \textbf{hierarchical merging} to find the adequate number of parts.
We introduce \emph{tightness-aware} merging and stopping criteria that enable selection of the optimal number of parts the best number of parts given the trade-off between tightness and parsimony in the decomposition.

While merging over the over-segments already produces promising results, the results are yet dependent on the quality of the over-segmentation (third column of Fig.~\ref{fig:teaser}-a). To overcome the dependency to the initialization, we present the next \textbf{refinement} step that adjusts the bounding box parameters following a sequence of predefined actions. We design a Markov Decision Process (MDP) setup with a \emph{soft} reward function that allows the bounding boxes to break the full coverage in the \emph{middle} of the process. This is the key to having more flexibility in traversing various cases and obtaining better results at the end (fourth column of Fig.~\ref{fig:teaser}-a). Lastly, we extend the refinement step to see not only a single step of actions but multiple steps in a small sequence. We utilize \textbf{MCTS}~\cite{mcts} and introduce acceleration techniques (in the supplementary) to speed up the additional refinement (last column of Fig.~\ref{fig:teaser}-a).

In our experiments with ShapeNet~\cite{shapenet2015}, we demonstrate that our method guaranteeing full coverage provides better tightness and reconstruction compared with the baseline methods while approximating the shapes into similar numbers of cuboids. We additionally show that the decomposition of a shape based on our bounding boxes is better aligned with the semantic parts than other methods. Furthermore, we provide the result of~\smart{} on Objaverse~\cite{objaverse} and OmniObject3D~\cite{wu2023omniobject3d} to show its applicability to real data and various categories of 3D shapes. 

To summarize,
\begin{itemize}
\setlength\itemsep{0.0em}
\item We present a novel framework for finding a set of tight bounding boxes of a 3D shape by optimizing volumetric objectives with iterative search methods.
\item We first propose a hierarchical merging method exploiting over-segmentation from a pre-segmentation and tightness-aware merging and stopping criteria.
\item We also introduce a bounding box refinement process with an effective soft reward function that allows wider exploration in the action space.
\item Finally, we present an MCTS-based efficient exploration of multi-action sequences with acceleration techniques.
\end{itemize}
\vspace{-0.5\baselineskip}

%% file: sections/2_Related_Work.tex
\section{Related Work}
\vspace{-3pt}

\paragraph{Learning-Based Shape Abstraction.}
As neural network-based learning approaches have shown powerful performance, various unsupervised learning-based approaches have attempted to abstract/reconstruct the shape with simple primitives. For primitives, cuboids~\cite{tulsiani17abstract, sun19abstract, yang21cubseg}, superquadrics~\cite{paschalidou19sq}, convexes~\cite{deng20cvxnet, chen20bspnet} and implicit fields~\cite{chen19bae_net, paschalidou21np, niu22rimnet} have been explored. Tulsiani~\etal{}~\cite{tulsiani17abstract} use deep convolutional neural networks, directly predicting volumetric primitives (VP) and their translation parameters. Sun~\etal{}~\cite{sun19abstract} try to discover the adaptive hierarchical cuboid abstraction (HA) to exploit the structural coherence of 3D shapes. Also, Yang and Chen~\cite{yang21cubseg} propose an unsupervised approach that jointly predicts the cuboid parameters and segmentation (CA) of input point clouds giving feedback to each other. Paschalidou~\etal{}~\cite{paschalidou19sq} use superquadrics (SQ) as primitives for abstraction which can represent cylinders, spheres, cuboids, ellipsoids, etc. Furthermore, convexes and implicit fields which are more general representations have been used as primitives to enhance the performance of reconstruction. One such approach by Chen~\etal{}~\cite{chen20bspnet} design a network to learn the associations of binary space partitioning (BSP-Net) as an implicit representation.

\vspace{-\baselineskip}
\paragraph{Volume-Based 3D Shape Decomposition.}
Before learning-based approaches became popular, classical approaches tried to segment 3D shapes by exploiting volumetric information. Lu~\etal{}~\cite{2007bvc} propose a variational formulation using volumetric information to compute a tight bounding volume and segmentation. Also, Attene~\etal{}~\cite{hierconvexmerge} define part concavity with volumetric information to obtain weakly convex decompositions. Asafi~\etal{}~\cite{volweakconvex}'s work is based on visibility (line of sight), which is one of the volumetric properties. Shapira~\etal{}~\cite{sdf_classic} define a shape diameter function that expresses the diameter of the object’s volume in the nearby point on the surface for consistent mesh partitioning. Liu~\etal{}~\cite{partawaremetric} design a part-aware surface metric that considers the volumetric context when encoding part information. Kaick~\etal{}~\cite{convexity} merge weakly convex components taking account of the volumetric profile of the parts to obtain the final segmentation. In the sense of exploiting volumetric information to capture part-level segmentation, ~\smart{} is similar to these approaches. However, such classical approaches usually start without any initialization while ~\smart{} can effectively utilize shape-dependent initialization learned by neural networks (BSP-Net)~\cite{chen20bspnet}. Moreover, we propose an additional step to refine the errors and failures of these approaches.

\vspace{-\baselineskip}
\paragraph{Efficient Search for 3D Applications.}
Efficient search is crucial for tackling high-dimensional search spaces since we cannot exhaustively search all the possible solutions. Continuity and extra dimensionality in 3D data make the search space extremely large, requiring dedicated algorithms for efficient searches. For this purpose beam search~\cite{beamsearchscene, sharma18csg}, monte carlo tree search (MCTS)~\cite{hampali21monte, stekovic21montefloor, mctsrefine, wei22approximate} and learning~\cite{sharma18csg, lin20modeling} approaches have been explored to solve various 3D computer vision tasks such as shape reconstruction, scene detection and convex decomposition. Sharma~\etal{}~\cite{sharma18csg} propose a neural shape parser that learns to generate CSG grammar to challenge the huge search spaces. After obtaining the results from the shape parser, they apply beam search to find the best result. Similarly, Lin~\etal{}~\cite{lin20modeling} propose a learning-based approach to reconstruct the given shape that effectively searches the large discrete action spaces. Hampali~\etal{}~\cite{hampali21monte} use MCTS to detect 3D objects in the scenes to guide the solution faster. Wei~\etal{}~\cite{wei22approximate} also utilize MCTS to overcome the limitations of one-step greedy search. In ~\smart{}, we change the optimization problem of finding tight bounding boxes to a search problem and apply MCTS to effectively tackle the huge search space to overcome the limitations of our one-step greedy heuristic search.

%% file: sections/3_Problem_Definition_and_Overview.tex
\section{Problem Definition and Overview}
\vspace{-3pt}



Our goal is to determine a set of bounding boxes, denoted as $\{B_i\}_{i=1}^M$, for a given 3D shape $S$ represented as a tetrahedral mesh. It is worth noting that any watertight mesh can be converted into a tetrahedral mesh using an off-the-shelf technique~\cite{ftetwild}. To define the desired set of bounding boxes, we first specify three criteria: (1) {\bf coverage}, (2) {\bf tightness}, and (3) the {\bf number of boxes}.

The measure of the {\bf coverage} (Cov) is defined as follows:
\vspace{-2pt}
\begin{align}
    \text{Cov}(S, \{B_i\}) = 1 - \cfrac{\text{vol}\left(S \; \backslash \; (\bigcup_{i=1}^M{B_i}) \right)}{\text{vol}(S)},
\end{align}
where $\text{vol}(\cdot)$ is a function measuring the volume of either a mesh or a bounding box.
$\text{Cov}(S, \{B_i\})$ measures the proportion of the volume of the input shape $S$ that is covered by the bounding boxes.
Our goal is to find a set of bounding boxes that \emph{fully} cover the given shape, and hence $\text{Cov}(S, \{B_i\})$ needs to be equal to one.

The {\bf tightness} (Tgt) is measured as follows, inspired by the variational formulation by Lu et al.~\cite{2007bvc}:
\vspace{-2pt}
\begin{align}
    \text{Tgt}(S, \{B_i\}) = \sum_{i=1}^M\cfrac{\text{vol}(B_i)}{\text{vol}(S)}.
\end{align}

When the coverage is one, the ideal set of bounding boxes that tightly fits the input shape has the minimum value of $\text{Tgt}(S, \{B_i\})$, which is close to one. Hence, our objective function for finding the tightest set of bounding boxes while fully covering the shape is formulated as follows:
%
\vspace{-2pt}
\begin{align}
\argmin_{\{B_i\}_{i=1}^M} \quad \text{Tgt}(S, \{B_i\}) \,\, \textrm{s.t.} \,\,  \text{Cov}(S, \{B_i\}) = 1.
\label{eq:hard-obj}
\end{align}

There can be multiple configurations of bounding boxes that achieve similar tightness with full coverage. Among these configurations, we aim to identify the set of bounding boxes with the {\bf minimum number of boxes}.

The objective function for coverage and tightness mentioned above involves non-differentiable operations, such as volume union and difference. Consequently, using gradient-descent-based methods to solve the problem is not feasible. Moreover, finding the minimal set of bounding boxes involves minimizing discrete variables, making the problem highly non-trivial and unsuitable for solving using neural network-based methods. As a result, recent learning-based methods fitting bounding boxes~\cite{tulsiani17abstract,sun19abstract,yang21cubseg} fail to achieve at least one of the desired criteria above: full coverage, tightness, or the minimum number of bounding boxes.


To this end, we propose a novel \emph{non-learning} framework that finds a set of bounding boxes satisfying the above criteria through three main steps: {\bf split}, {\bf merge}, and {\bf refinement}. We first {\bf split} the given 3D shape using existing techniques~\cite{chen20bspnet,wei22approximate} and obtain an over-segmentation. It allows us to achieve high performance regardless of the choice of over-segmentation techniques~\cite{chen20bspnet,wei22approximate}.

In the subsequent {\bf merging} phase, we revisit the classical hierarchical merging methods~\cite{2007bvc, convexity, hierconvexmerge} with our volumetric criteria (BAVF) to merge the partitions (Sec.~\ref{sec:cluster}). BAVF allows us to determine the appropriate grouping of the initial segments that provides tightness and an adequate number of parts while guaranteeing full coverage.


We aim to improve the bounding boxes through the {\bf refinement} step, which transforms our optimization problem (Eq.~\ref{eq:hard-obj}) into a search problem by iteratively applying discrete unit actions (Sec.~\ref{sec:refine}). Our key observation in this phase is that guaranteeing full coverage may lead to sub-optimal results in the refinement. Hence, we treat coverage as a soft constraint, penalizing low coverage rather than guaranteeing full coverage. Then, we recover full coverage at the end of the process using a simple heuristic. This approach significantly helps to avoid getting trapped in local minima during the optimization process.

Finally, we additionally address the limitations of the one-step greedy search by proposing a general solution with MCTS, which allows us to perform a multi-step search (Sec.~\ref{sec:mcts}). We also propose techniques to accelerate the MCTS in the supplementary.

%% file: sections/4_Method.tex
\section{\smart{} Framework}
\vspace{-3pt}

\subsection{Initialization via Over-Segmentation}
\label{sec:bspinit}
\vspace{-3pt}

\begin{figure}[ht!]
    \centering
    \includegraphics[width=0.8\columnwidth]{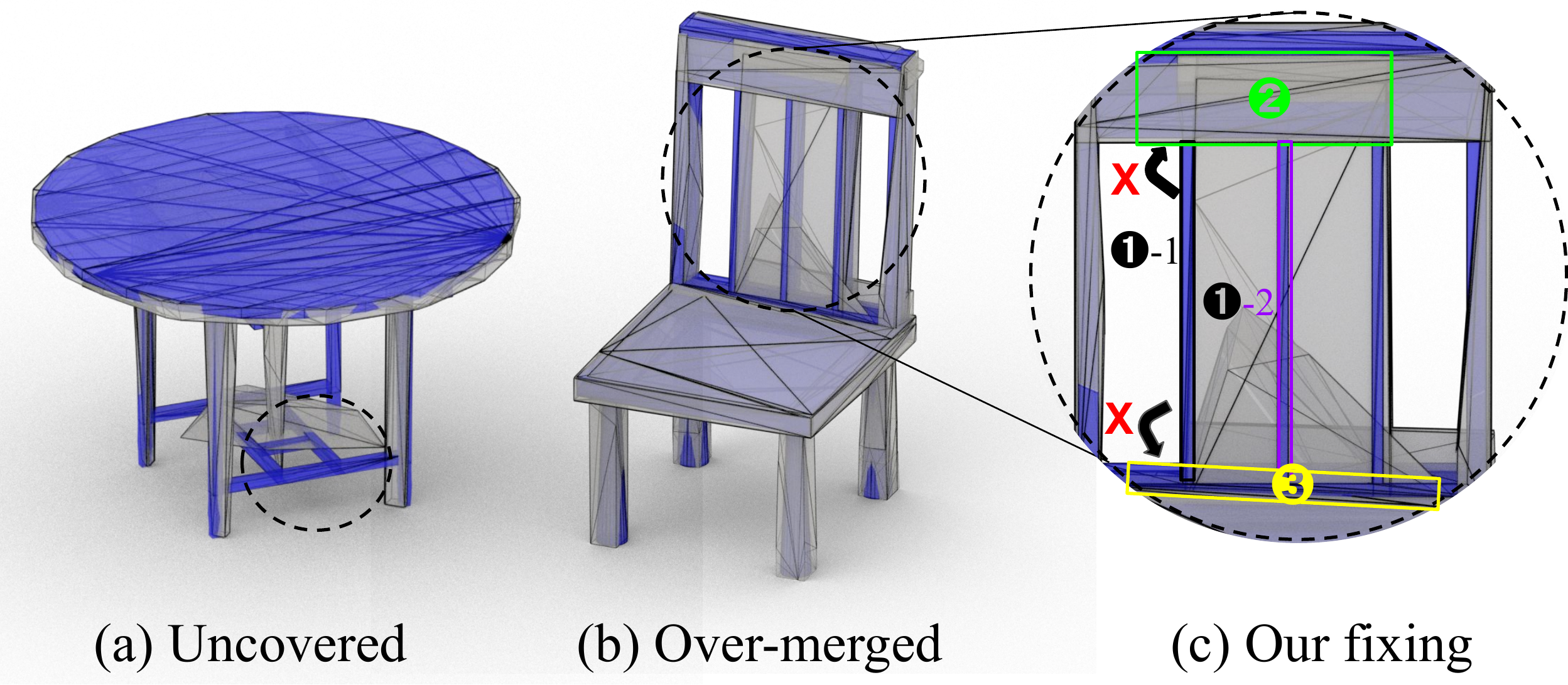}
    \caption{Typical failure cases of pre-segment and our fixing (Blue: Target shape, Grey: BSP-Net). (a) pre-segment that does not fully cover the entire shape, (b) pre-segment that over-merges the closely located parts, and (c) how our initialization can fix the over-merged cases using the flood fill algorithm.}
    \label{fig:bspfail}
    \vspace{-\baselineskip}
\end{figure}

Pre-segments are easier to obtain compared to part-level segmentation or partitions in an unsupervised manner. For example, BSP-Net~\cite{chen20bspnet} or convex decomposition~\cite{wei22approximate} (CoACD) can be used as pre-segments in our initialization. However, using them directly as initial over-segmentation is not straightforward due to their common problems: (1) uncovered parts (Fig.~\ref{fig:bspfail}-a), (2) overlapping segments, and (3) merging of closely located parts (Fig.~\ref{fig:bspfail}-b). Ignoring these problems will cause ambiguity in parsing volumetric information thereby induce inferior result.

To handle these problems and obtain an initial over-segment (partition) for our purpose, we first split the overlaps and regroup the partitions by using our simple flooding algorithm (Fig.~\ref{fig:bspfail}-c). 

Concretely, to resolve the second limitation, we split each overlap generated by pre-segments into different partitions. For the third limitation, we observe that separated parts in the shape merged by pre-segments typically resemble the case shown at Fig.~\ref{fig:bspfail}-b. In such cases, if the upper and lower segments are correctly partitioned, we can separate the parts merged by pre-segments using a flood fill algorithm (Fig.~\ref{fig:bspfail}-c). Lastly, for the first limitation, we merge the nearby uncovered parts to form each partition. The details are described in the supplementary.

\subsection{Hierarchical Merging}
\label{sec:cluster}
\vspace{-3pt}

\paragraph{Bounding-Box-Aware Volume Function.}
We introduce the Bounding-box-Aware Volume Function (BAVF) as our merging criteria in the hierarchical clustering~\cite{patel2015study} process to obtain the part-level segmentation. This volume-based criterion is the key to grouping the over-segments into part-level without requiring any exhaustive search. BAVF calculates the decrease in the bounding volumes that results from merging the two partitions as follows:
\vspace{-2pt}
\begin{align}
\text{BAVF}(S_i^t, S_j^t) &=  \text{Tgt}(\{B_i\}^{t}) - \text{Tgt}(\{B_i\}^{t+1}) \nonumber \\
&= \cfrac{\text{vol}(B_i^t) + \text{vol}(B_j^t) - \text{vol}(B_{ij}^{t+1})}{\text{vol}(S)}.
\label{eq:bavf}
\end{align}
$S_i^t$ is the partition of the tetrahedral mesh $S$ at timestep $t$, $B_i^t$ and $B_j^t$ are the oriented bounding boxes calculated by finding the minimum volume of the vertices of $S_i^t$ and $S_j^t$, respectively, and $B_{ij}^{t+1}$ is the bounding box of $S_i^t \cup S_j^t$.

\begin{figure}
    \centering
    \includegraphics[width=0.8\columnwidth]{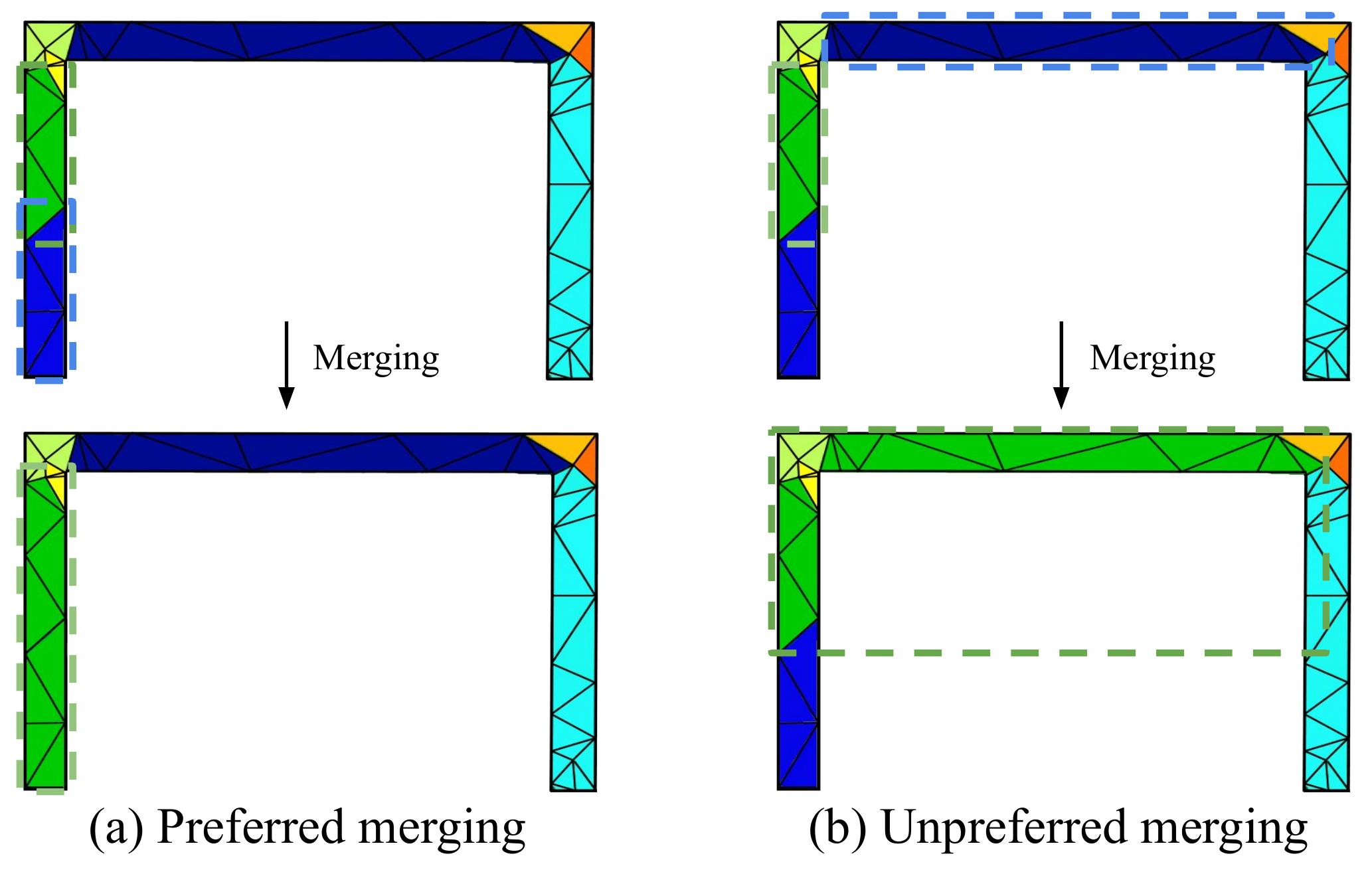}
    \caption{When merging the green and blue partitions, (a) shows a preferred merging case that decreases the bounding volume. (b) shows an unpreferred merging case that significantly increases bounding volume.}
    \label{fig:expfunction}
    \vspace{-2pt}
\end{figure}

\begin{figure}
    \vspace{-\baselineskip}
    \centering
    \includegraphics[width=0.8\columnwidth]{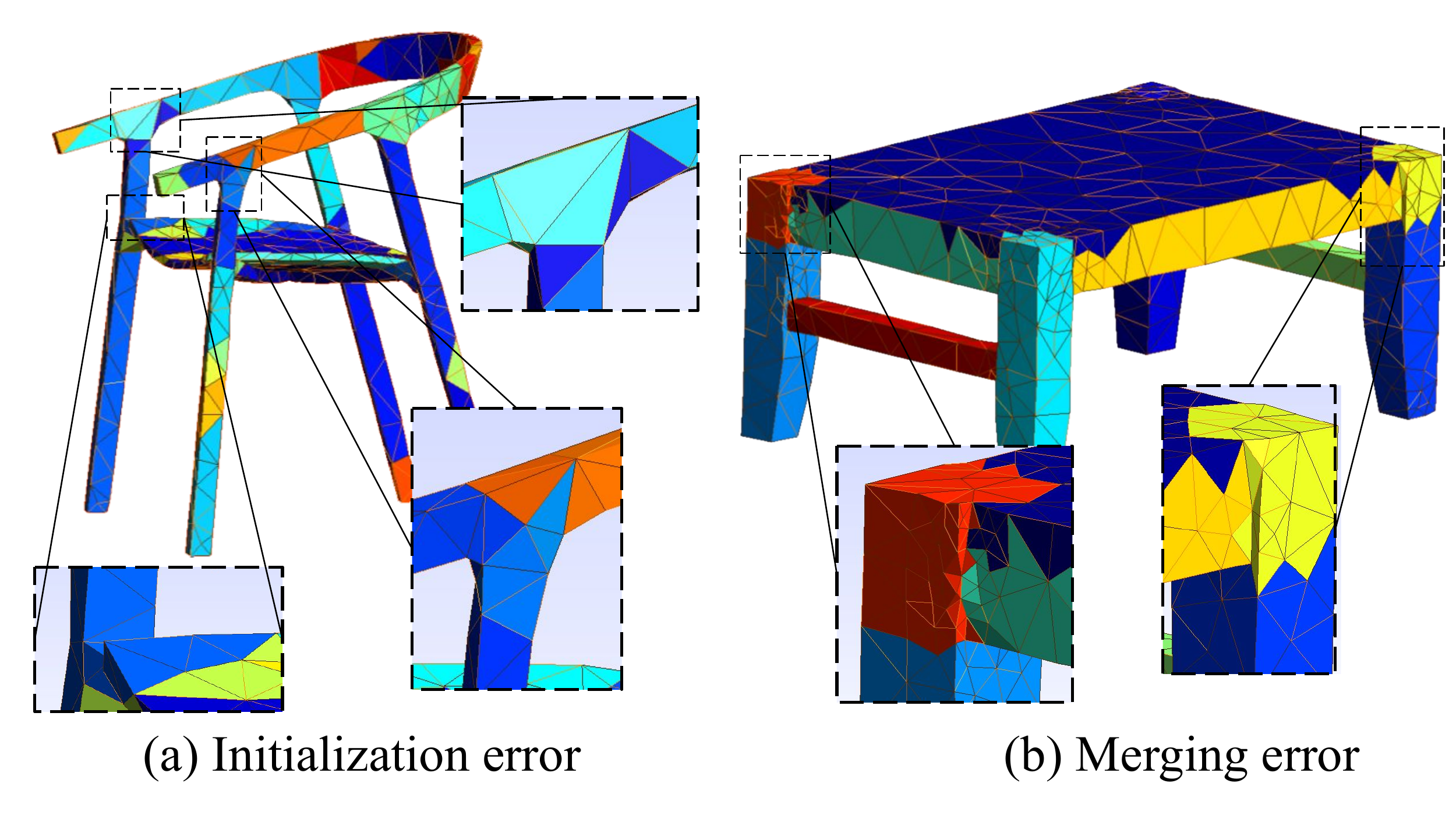}
    \caption{(a) shows the initialization errors where a partition that should not be in the same partition is initialized together. (b) shows the merging errors where partitions that should be merged are not.}
    \label{fig:segerror}
    \vspace{-\baselineskip}
\end{figure}

By utilizing BAVF in hierarchical clustering, we can achieve two objectives simultaneously: minimizing Tgt($S, \{B_i\}$) and reducing the number of partitions, while also being able to distinguish between the part-level merging that should occur and not (as shown in Fig.~\ref{fig:expfunction}). Furthermore, in the early stages of clustering, BAVF tends to prioritize merging nearby partitions rather than distant ones, which aids in achieving the goal of obtaining a part-level segmentation.
\vspace{-\baselineskip}
\paragraph{Hierarchical Clustering.}
Hierarchical clustering starts by identifying the partition pairs with the largest BAVF value, and merging them if the BAVF value is positive. The threshold $\epsilon_{merge}$ is used as a stopping criterion to determine when to stop merging. Because of the low resolution of tetrahedral meshes and errors in initialization, a small negative value is used to force the merging of small errors (Fig.~\ref{fig:segerror}-a). If the BAVF value is negative, the trade-off between decreasing the number of cuboids and reducing the Tgt($S, \{B_i\}$) is considered. When the BAVF value is larger than the threshold, the partitions are still merged. However, if the BAVF value is smaller than the threshold, partitions are no longer merged, and the hierarchical clustering is terminated. The details are described in the supplementary.

After obtaining the part-level segmentation with merging, the oriented bounding box that covers each part-level segmentation with minimum volume is calculated.

\subsection{Bounding Box Refinement}
\vspace{-3pt}

\subsubsection{Greedy Refinement}
\vspace{-3pt}
\label{sec:refine}

Although the bounding boxes obtained from merging (Sec.~\ref{sec:cluster}) hold potential, they are not as tightly fitted as desired. Also, errors can occur during the initialization step (as shown in Fig.~\ref{fig:segerror}-a), causing the bounding boxes to become loose. Additionally, the merging step (Fig.~\ref{fig:segerror}-b) can result in some unmerged bounding boxes, leading to have unnecessary boxes.

Our refinement step targets to fix these errors by formulating the optimization problem into Markov Decision Process (MDP)~\cite{bellman1957markovian}. We can define the state and action space like below and give reward as the improvement of Tgt($S, \{B_i\}$) while $\text{Cov}(S, \{B_i\})$ is 1 following our hard objective (Eq.~\ref{eq:hard-obj}). We iteratively apply greedy actions that minimize Tgt($S, \{B_i\}$) but do not decrease $\text{Cov}(S, \{B_i\})$.

\paragraph{State.}
\vspace{-\baselineskip}
Each bounding box is parameterized with the two diagonal points of the cuboid and a rotation matrix R at time-step t. If one of the coordinates of the left vertex becomes bigger than the right vertex, we treat the bounding box as deleted. Note that $(lx_i, ly_i, lz_i)$ represents the left diagonal point of the bounding box.
\begin{align}
    \{B_i^t\}_{i=1}^M = \{(lx_{i}, ly_{i}, lz_{i}, rx_{i}, ry_{i}, rz_{i}, R_{i})\}_{i=1}^M
\end{align}
\vspace{-5pt}
\vspace{-\baselineskip}
\vspace{-10pt}
\paragraph{Action.}
There are two types of actions available for each bounding box: (1) adding or subtracting a predefined unit scale to one of the parameters in $(lx_{i}, ly_{i}, lz_{i}, rx_{i}, ry_{i}, rz_{i})$ and (2) changing the rotation matrix $R_i$ to orient the bounding box correctly based on the points it covers. The second action is necessary to correct the wrong rotations derived from poor segmentation. In total, there are $6\times2+1=13$ actions for modifying each bounding box.

\begin{figure}
    \centering
    \includegraphics[width=\columnwidth]{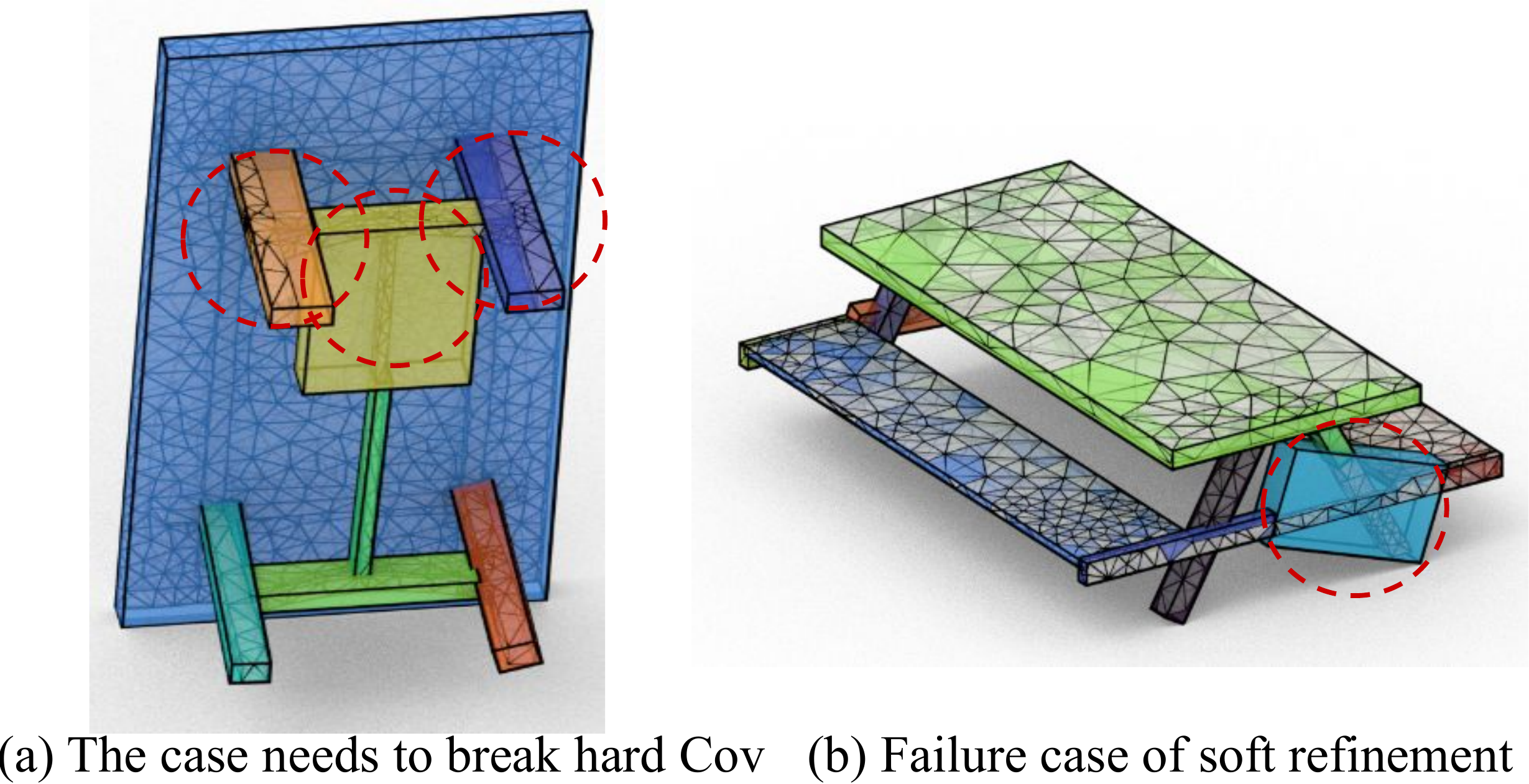}
    \caption{(a) shows the case that needs to perform actions violating Cov constraint. (b) shows the failure case of soft refinement that can not handle the rotation of the bounding box decided by the wrong segmentation.}
    \label{fig:refineexp}
    \vspace{-10pt}
\end{figure}

\vspace{-\baselineskip}
\paragraph{Soft Refinement.}
With the hard constraint, obtaining a satisfactory refinement of cases similar to Fig.~\ref{fig:refineexp}-a is impossible. To address this, we make adjustments to the initial objective function (Eq.~\ref{eq:hard-obj}) by transforming the hard constraint into a flexible soft constraint. However, to ensure $\text{Cov}(S, \{B_i\})$ to be close to 1, we introduce a coefficient denoted as $\alpha$. We iteratively apply one-step greedy actions that minimize the objective function
\begin{align}
    \argmin_{\{B_i\}_{i=1}^M} \quad \text{Tgt}(S, \{B_i\}) - \alpha \text{Cov}(S, \{B_i\}).
    \label{eq:soft}
\end{align}
%
This updated objective function enables a trade-off between coverage and minimizing the bounding volume. Consequently, refinement attempts can be made to break $\text{Cov}(S, \{B_i\})$ = 1, while still finding other actions that guarantee a $\text{Cov}(S, \{B_i\})$ of almost 1, leading to a decrease in Tgt($S, \{B_i\}$). With an appropriate $\alpha$, our soft refinement approach can refine such cases (Fig.~\ref{fig:refineexp}-a), while maintaining $\text{Cov}(S, \{B_i\})$ to be almost 1.

\vspace{-\baselineskip}
\paragraph{Post-Processing.}
The soft refinement may neglect small-volume parts of the shape even with such high $\alpha$. So, to guarantee full coverage, we post-process the outputs by fitting boxes for each nearby uncovered part or using a ternary search to fit the boundaries of the bounding boxes which are smaller than the predefined unit scale.

\vspace{-\baselineskip}
\subsubsection{Monte Carlo Tree Search (MCTS)}
\label{sec:mcts}
\vspace{-5pt}

Fig.~\ref{fig:refineexp}-b demonstrate the insufficiency of one-step greedy search methods (soft refinement) in resolving our problem. Due to the huge search space of size $(M \times 13)^T$ with multiple local minima, it fails to find actions that can fix the wrong rotation. It occurs because rotating typically does not result in an immediate improvement in Tgt($S, \{B_i\}$), making it unfixable through a greedy approach. To mitigate this issue, we propose utilizing MCTS, which allows us to efficiently search the huge search space by simulating multiple steps while saving the explored results and utilizing them for further searches.

\vspace{-5pt}
\paragraph{Tree Structure and Iteration.}
\vspace{-5pt}
In our tree search, each node represents the state of our MDP setup. The state is characterized by the parameters of the bounding boxes $\{B_i^t\}_{i=1}^M$, and each edge represents an action. The root node represents the initial bounding boxes, and we aim to tightly refine the initial bounding boxes using a sequence of actions within a limited time-step $T$. Each node has a child node that corresponds to the state of the bounding boxes after applying possible actions to its node's state. After running pre-defined iterations, we take the best bounding boxes as the result. Refer to the supplementary for the details of MCTS and its acceleration.

%% file: sections/5_Results.tex
\section{Results}
\vspace{-3pt}

\input{tables/tbl_new}






\begin{figure*}[h!]
\centering
{\footnotesize
\begin{tabular}{>{\centering\arraybackslash}m{0.09\textwidth} >{\centering\arraybackslash}m{0.81\textwidth}}
Input  & \includegraphics[width=\linewidth]{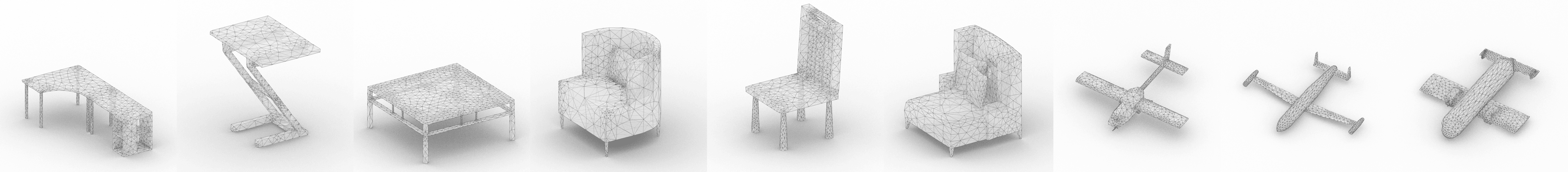} \\
HA~\cite{sun19abstract}  & \includegraphics[width=\linewidth]{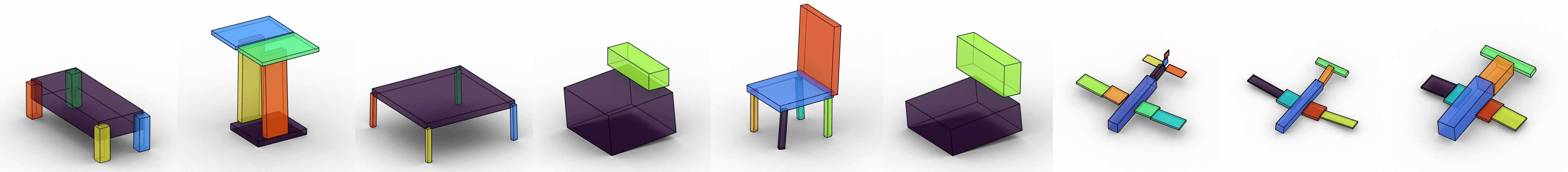} \\
HA~\cite{sun19abstract}+Ref.  & \includegraphics[width=\linewidth]{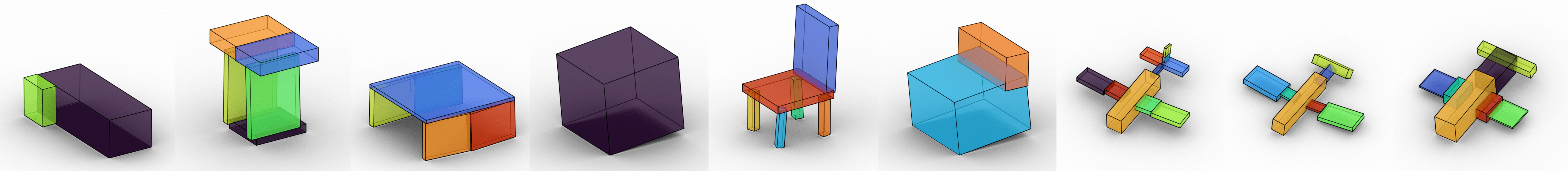} \\
CA~\cite{yang21cubseg}  & \includegraphics[width=\linewidth]{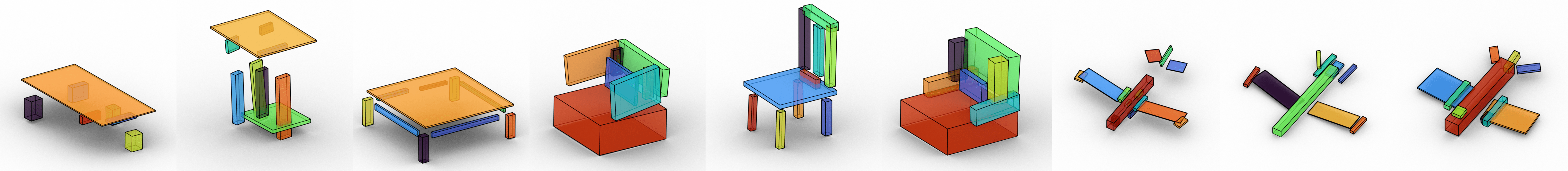} \\
CA~\cite{yang21cubseg}+Ref.  & \includegraphics[width=\linewidth]{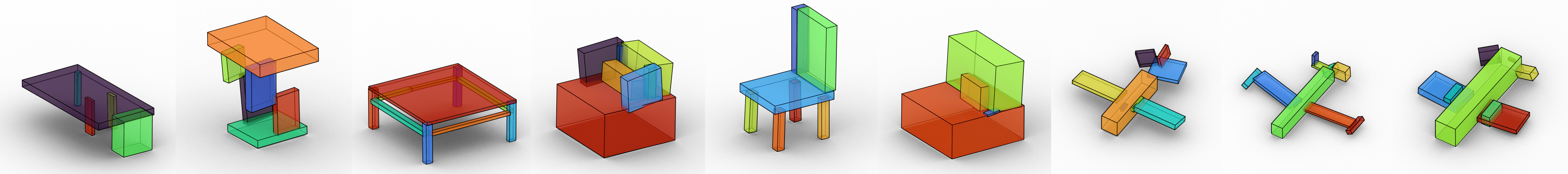} \\
Merge  & \includegraphics[width=\linewidth]{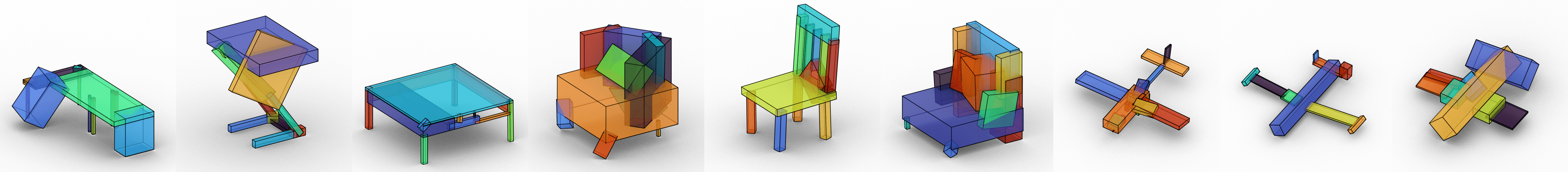} \\
Merge+Ref.  & \includegraphics[width=\linewidth]{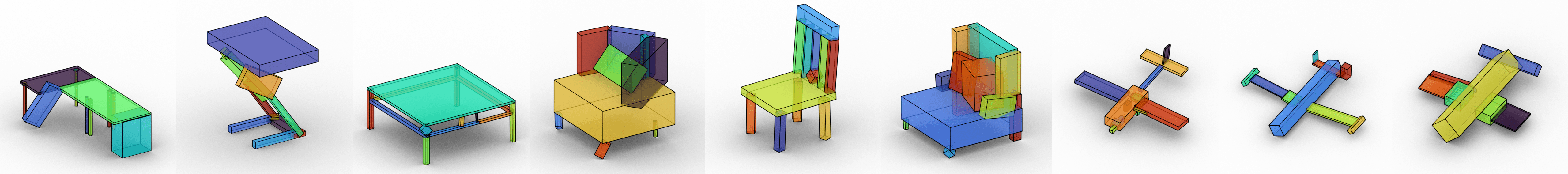} \\
\makecell{Merge+Ref.\\+MCTS}  & \includegraphics[width=\linewidth]{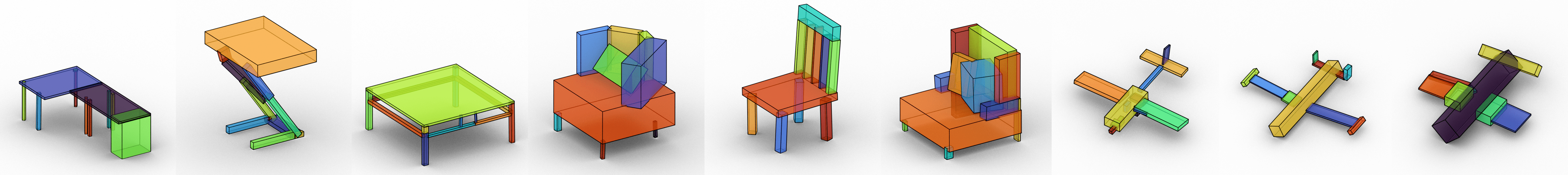} \\
\end{tabular}
}
\vspace{-\baselineskip}
\caption{Qualitative comparison between ~\smart{} and baselines.}
\label{fig:qual}
\end{figure*}

Our bounding boxes are tight bounding volumes that can preserve structural details, resulting in precise reconstruction results without requiring any training data. Also, they provide instance-level segmentation since each box covers each part without overlaps. We comprehensively evaluate our approach's tightness, reconstruction, and instance-level segmentation performance compared to other cuboid abstraction baselines. And we conduct an ablation study to validate our refinement step and MCTS. Furthermore, we test~\smart{} on diverse categories to show its applicability.

\subsection{Implementation Details} 
\vspace{-3pt}

We evaluate~\smart{} on the ShapeNet~\cite{shapenet2015} dataset using 9 categories including 3496 tables, 2500 chairs, and 531 airplanes. Additionally, we randomly selected shapes from Objaverse~\cite{objaverse} and OmniObject3D~\cite{wu2023omniobject3d} to evaluate~\smart{}. BSP-Net~\cite{chen20bspnet} was used as the pre-segment in the initialization step for the ShapeNet meshes, while CoACD~\cite{wei22approximate} was used as the pre-segment for the Objaverse and OmniObject3D meshes. Furthermore, for the volume calculation of meshes, we converted them into watertight tetrahedral mesh by using the algorithm of Huang~
\etal{}~\cite{huang2020manifoldplus} and fTetwild~\cite{ftetwild}. Note that finding an oriented bounding box is implemented with trimesh~\cite{trimesh}. In~\smart{}, we used $\epsilon_{merge}$ as -0.02 and $\alpha$ as 100 and $c$ with 0.001 in all of our experiments.

\subsection{Evaluation Metrics}
\vspace{-3pt}

For evaluating the tightness of our bounding boxes, we use TOV and MOV proposed by Lu et al.~\cite{2007bvc}. Tgt and Cov are also used since they are the main objectives of our problem. Also, to evaluate the reconstruction performance, we adopt the commonly-used chamfer distance~\cite{chamferdistance} (CD) and volumetric IoU (VIoU). Part-level segmentation is evaluated by calculating the mean Average Precision (mAP) using instance labels from PartNet~\cite{mo19partnet}.

\begin{itemize}[itemsep=0ex]
    \item \textbf{Total Outside Volume:}\\
    $\text{TOV}(B) = \cfrac{\text{vol}(\bigcup_i B_i \; \backslash \; S)}{\text{vol}(S)}$
    \item \textbf{Maximum local Outside Volume:}\\
    $\text{MOV}(B, \{S_i\}) = \max_i{\cfrac{\text{vol}(B_i \; \backslash \; S)}{\text{vol}(S_i)}}$
    \item \textbf{Volumetric Intersection over Union:}\\
    $\text{VIoU}(B) = \cfrac{\text{vol}(S \cap \bigcup_i B_i)}{\text{vol}(S \cup \bigcup_i B_i)}$
\end{itemize}

\subsection{Tightness of Bounding Boxes}
\vspace{-3pt}

We first evaluate the tightness of the bounding boxes with Tgt, Cov, MOV~\cite{2007bvc}, and TOV~\cite{2007bvc}. Since there is no baseline work that directly solves the problem of finding tight bounding boxes, we apply our refinement algorithm to the recent state-of-the-art cuboid abstraction works HA~\cite{sun19abstract} and CA~\cite{yang21cubseg} to achieve the $\text{Cov}(S, \{B_i\})$ to be 1. This is necessarily for a fair comparison since TOV, MOV can be properly evaluated only if $\text{Cov}(S, \{B_i\})$ is 1. We report the result of the refined baselines (row 2, 4 in each class) and ~\smart{} (last four row in each class) in Tab.~\ref{tbl:all_comp}. Even compared without refinement on our merging step, ~\smart{} outperforms the baselines in all metrics showing big margins. We can also see that our refinement algorithm successfully tightens the output of our merging step. We give a qualitative comparison in Fig.~\ref{fig:qual}.
\subsection{Shape Reconstruction}

We compare the reconstruction performance of~\smart{} to CA~\cite{yang21cubseg} and HA~\cite{sun19abstract} to demonstrate its superiority. We use CD and VIoU to evaluate the reconstruction performance, where CD was calculated by uniformly sampling 4096 points from the obtained cuboids and the ShapeNet~\cite{shapenet2015} mesh. VIoU was calculated by measuring the IoU between the input mesh and the merged bounding boxes. As shown in Tab.~\ref{tbl:all_comp} (sixth and seventh columns of metric),~\smart{} outperforms all baselines in VIoU while achieving better or comparable performance in CD. It is important to note that, unlike other cuboid abstraction baselines, we did not directly optimize for CD. For qualitative comparison, refer to Fig.~\ref{fig:qual}, where it can be observed that other baselines fail to properly abstract the shape or capture details, while~\smart{} accurately captures them even with the restriction of covering the entire shape ($\text{Cov}(S, \{B_i\})$ = 1).

\vspace{-2pt}
\subsection{Shape Instance Segmentation}
\vspace{-3pt}

Since our bounding boxes are based on part-level segmentation, we also measure its performance by measuring the mean average precision (mAP) of instance segmentation with the labels obtained from PartNet~\cite{mo19partnet}. In addition to HA~\cite{sun19abstract} and CA~\cite{yang21cubseg}, we also compare with BSP-Net~\cite{chen20bspnet} which was used to get our initial segment. We follow the evaluation setting of PartNet~\cite{mo19partnet} to report the instance segmentation results. The segmentation was obtained by identifying the nearest point for each primitive and assigning them to obtain the segmentation mask. Subsequently, we computed the IoU between each segmentation mask and the ground-truth mask and considered a segmentation mask as true positive only when IoU was greater than 0.5. Note that instance segmentation for the airplane category is not reported because PartNet~\cite{mo19partnet} does not have ground-truth labels. The results are reported in the last column of Tab.~\ref{tbl:all_comp}. Compared to other baselines we achieve the best result showing that ~\smart{} captures the individual parts in the shape better than other works having more alignment with the human perception of the shape decomposition.

\vspace{-3pt}
\subsection{Ablation Study}
\vspace{-3pt}

To examine the benefits introduced by the refinement step and MCTS, we provide a quantitative and qualitative comparison. The last four rows of each class in Tab.~\ref{tbl:all_comp} shows the results for these cases. MCTS successfully improves the failure cases of one-step greedy refinement by performing multi-step searches for all the categories. Fig.~\ref{fig:qual} shows the qualitative comparison between our methods. Our soft-refinement fails to handle the rotations, while our MCTS successfully addresses the issue.

\subsection{More Diverse Categories}
\label{sup:diverse}
\begin{figure}[h!]
    \centering
    \includegraphics[width=\linewidth]{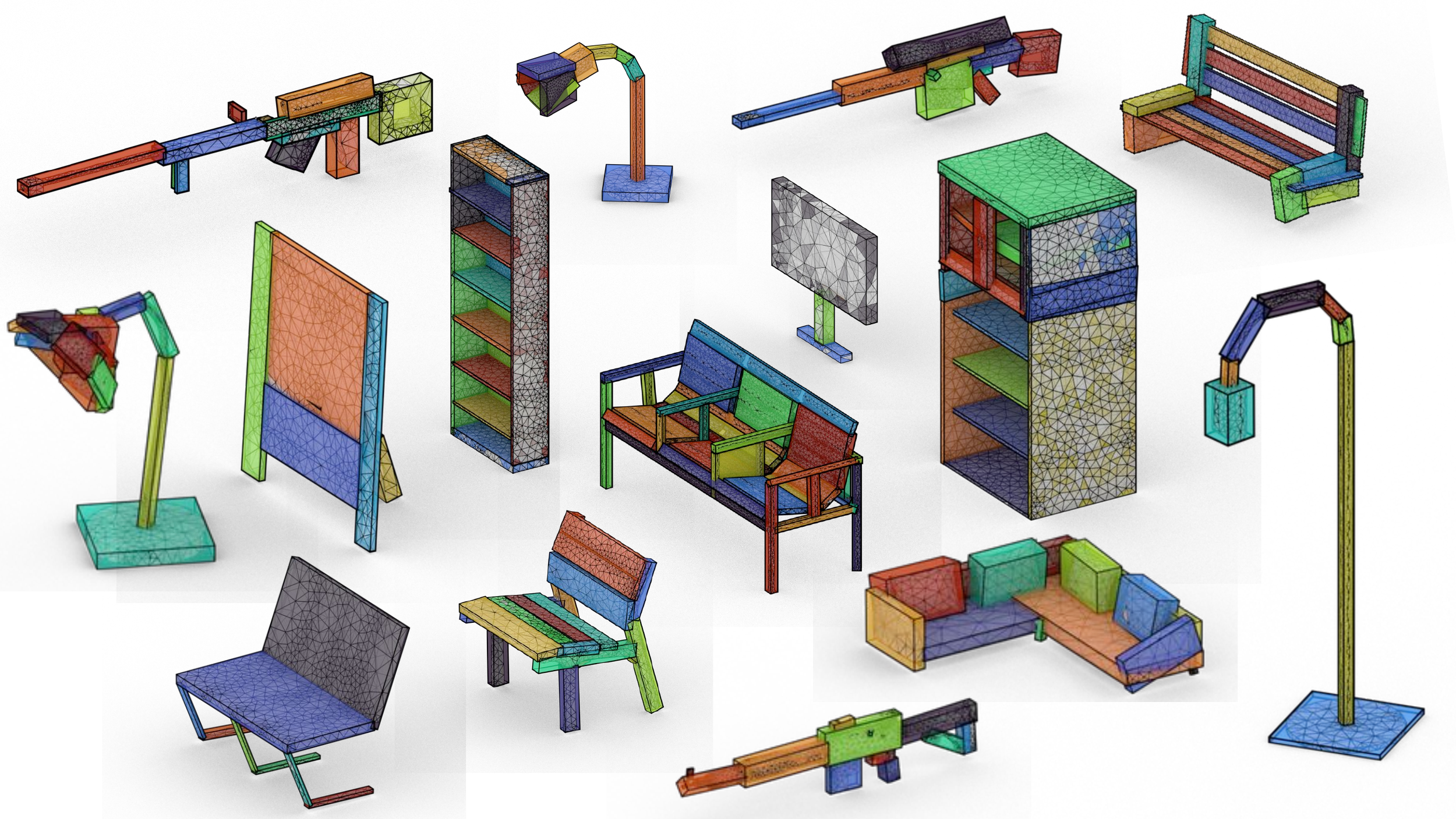}
    \caption{Qualitative result on diverse categories in ShapeNet~\cite{shapenet2015}.}
    \label{fig:div}
    \vspace{-\baselineskip}
\end{figure}

We also provide quantitative and qualitative results for other shape categories in ShapeNet~\cite{shapenet2015} in Fig.~\ref{fig:div} and Tab.~\ref{tbl:other_cat}, including Bench, Cabinet, Couch, Display, Lamp, and Rifle. Note that other baselines such as HA~\cite{sun19abstract} and CA~\cite{yang21cubseg} do not provide any pre-trained models or experiments for other shape categories. Additionally, Fig.~\ref{fig:teaser}-b shows the qualitative result of Objaverse~\cite{objaverse}.
\input{tables/supp/other_cat}

\subsection{Application to Real Data}
\label{sup:real_data}

\begin{figure}[h!]
    \centering
    \includegraphics[width=\linewidth]{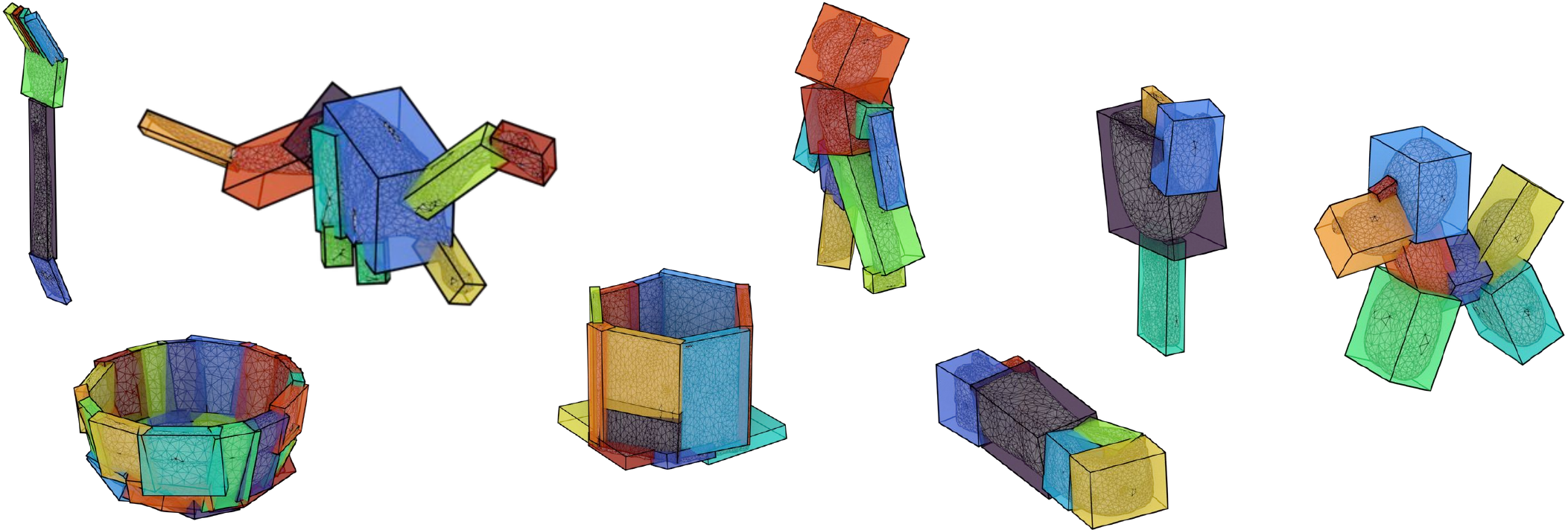}
    \caption{Qualitative result on OmniObject3D~\cite{wu2023omniobject3d}.}
    \label{fig:omniobj}
    \vspace{-\baselineskip}
\end{figure}

For testing applicability and robustness to real data, we show the result of applying~\smart{} to OmniObject3D~\cite{wu2023omniobject3d} dataset at Fig~\ref{fig:omniobj}. Note that OmniObject3D meshes are reconstructed from real 3D objects using multi-view images.

\subsection{Timing and Scalability Analysis}
\label{sup:timing}
We give a timing and scalability analysis of our framework. Using our \emph{unoptimized} coded implemented in \emph{Python}, all the 3496 tables in ShapeNet~\cite{shapenet2015} were processed within a day using 48 CPU cores. For further analysis, we randomly sampled 100 shapes from the ShapeNet table category and ran the entire pipeline. For 90\% of the shapes, it had fewer than 15 bounding boxes after processing. With a single CPU core, it took 8.48 minutes to run the entire pipeline for each shape on average. Within this time, the \emph{optional} MCTS step took 6.22 minutes per shape on average.




%% file: tables/tbl_new.tex
\begin{table}[t!]
\centering

\scriptsize{
\setlength{\tabcolsep}{0.4em}
\renewcommand{\arraystretch}{1.0}
\newcolumntype{Y}{>{\centering\arraybackslash}X}
\newcolumntype{Z}{>{\centering\arraybackslash}m{1.1cm}}
\begin{tabularx}{\linewidth}{c|>{\centering}m{1.57cm}|YYYYYYYY}
  \toprule
   \multicolumn{2}{c|}{Class} & $\#$Box & Tgt$\downarrow$ & Cov$\uparrow$ & MOV$\downarrow$ & TOV$\downarrow$ & CD$\downarrow$ & VIoU$\uparrow$ & mAP$\uparrow$\\
  
   \midrule
   \multirow{8}{*}{\rotatebox[origin=c]{90}{Table}} & HA~\cite{sun19abstract} & 4.59 & - & 0.69 & - & - & 1.41 & 0.41 & 0.31\\
   & HA~\cite{sun19abstract}+Ref. & 4.39 & 3.69 & 0.99 & 6.78 & 2.64 & 2.05 & 0.39 & - \\
   & CA~\cite{yang21cubseg} & 7.49 & - & 0.61 & - & - & 1.19 & 0.45 & 0.51 \\
   & CA~\cite{yang21cubseg}+Ref. & 5.31 & 3.23 & \textbf{1.00} & 6.05 & 2.21 & 1.75 & 0.40 & - \\
   & BSP-Net~\cite{chen20bspnet} & 12.9 & - & - & - & - & - & - & 0.51 \\
   & Ours (Merge) & 8.89 & 2.21 & \textbf{1.00} & 3.95 & 1.04 & 0.76 & 0.60 & 0.58\\
   & Ours (H Ref.) & 8.43 & 2.00 & \textbf{1.00} & 4.72 & 0.95 & 0.74 & 0.63 & 0.59\\
   & Ours (Ref.) & 7.93 & 1.72 & \textbf{1.00} & 2.82 & 0.69 & 0.64 & 0.67 & \textbf{0.60}\\
   & Ours (MCTS) & 7.87 & \textbf{1.69} & \textbf{1.00} & \textbf{2.58} & \textbf{0.67} & \textbf{0.63} & \textbf{0.68} & \textbf{0.60}\\
   \midrule

   \multirow{8}{*}{\rotatebox[origin=c]{90}{Chair}} & HA~\cite{sun19abstract} & 6.34 & - & 0.76 & - & - & 1.42 & 0.49 & 0.38\\
   & HA~\cite{sun19abstract}+Ref. & 5.70 & 3.23 & \textbf{1.00} & 5.82 & 2.19 & 2.76 & 0.41 & -\\
   & CA~\cite{yang21cubseg} & 9.10 & - & 0.72 & - & - & \textbf{0.81} & 0.56 & 0.57\\
   & CA~\cite{yang21cubseg}+Ref. & 7.78 & 2.54 & \textbf{1.00} & 5.14 & 1.50 & 1.55 & 0.49 & -\\
   & BSP-Net~\cite{chen20bspnet} & 20.5 & - & - & - & - & - & - & 0.48 \\
   & Ours (Merge) & 11.58 & 2.09 & \textbf{1.00} & 4.29 & 0.83 & 0.96 & 0.63 & 0.58 \\
   & Ours (H Ref.) & 11.28 & 1.86 & \textbf{1.00} & 6.02 & 0.76 & 0.98 & 0.66 & \textbf{0.59}\\
   & Ours (Ref.) & 10.84 & 1.64 & \textbf{1.00} & 2.96 & 0.58 & 0.84 & \textbf{0.69} & \textbf{0.59}\\
   & Ours (MCTS) & 10.81 & \textbf{1.62} & \textbf{1.00} & \textbf{2.73} & \textbf{0.56} & 0.83 & \textbf{0.69} & \textbf{0.59}\\

   \midrule

   \multirow{7}{*}{\rotatebox[origin=c]{90}{Airplane}} & HA~\cite{sun19abstract} & 7.39 & - & 0.68 & - & - & 0.52 & 0.50 & -\\
   & HA~\cite{sun19abstract}+Ref. & 7.00 & 2.91 & \textbf{1.00} & 4.82 & 1.82 & 0.89 & 0.37 & -\\
   & CA~\cite{yang21cubseg} & 9.48 & - & 0.59 & - & - & 0.45 & 0.46 & -\\
   & CA~\cite{yang21cubseg}+Ref. & 8.16 & 2.83 & \textbf{1.00} & 5.50 & 1.72 & 0.69 & 0.38 & -\\
   & Ours (Merge) & 13.2 & 2.62 & \textbf{1.00} & 3.82 & 1.21 & 0.42 & 0.47 & -\\
   & Ours (H Ref.) & 13.0 & 2.34 & \textbf{1.00} & 4.56 & 1.14 & 0.45 & 0.48 & -\\
   & Ours (Ref.) & 12.6 & 2.06 & \textbf{1.00} & 3.81 & 0.93 & 0.39 & \textbf{0.53} & -\\
   & Ours (MCTS) & 12.6 & \textbf{2.05} & \textbf{1.00} & \textbf{3.77} & \textbf{0.92} & \textbf{0.38} & \textbf{0.53} & -\\
   
  \bottomrule
\end{tabularx}
}
\vspace{-0.5\baselineskip}
\caption{Comparison of~\smart{} and baselines. Ref. stands for soft refinement (Eq.~\ref{eq:soft}) in our framework and H Ref. stands for refinement using the hard objective (Eq.~\ref{eq:hard-obj}). $\#$ Box denotes average number of primitives. Note that all CD is scaled by 1000.}
\vspace{-\baselineskip}
\label{tbl:all_comp}
\end{table}

%% file: tables/supp/other_cat.tex
\begin{table}[ht!]
    \centering
    {\footnotesize
    \newcolumntype{Y}{>{\centering\arraybackslash}X}
    \newcolumntype{Z}{>{\centering\arraybackslash}m{1.1cm}}
    \begin{tabularx}{\linewidth}{>{\centering}m{1.0cm}|ZYYYYYYY}
        \toprule 
        Category & $\#$ Box & Tgt$\downarrow$ & Cov & MOV$\downarrow$ & TOV$\downarrow$ & CD$\downarrow$ & VIoU$\uparrow$\\
        \midrule
        Bench & 10.59 & 1.92 & 1.00 & 3.28 & 0.87 & 0.49 & 0.61\\
        Cabinet & 5.5 & 1.45 & 1.00 & 3.85 & 0.43 & 1.11 & 0.81\\
        Couch & 6.59 & 1.30 & 1.00 & 3.59 & 0.25 & 1.04 & 0.82\\
        Display & 4.45 & 1.37 & 1.00 & 1.24 & 0.34 & 0.82 & 0.76\\
        Lamp & 8.93 & 2.30 & 1.00 & 5.75 & 1.20 & 1.49 & 0.54\\
        Riffle & 9.06 & 1.70 & 1.00 & 2.79 & 0.62 & 0.23 & 0.63\\
        \bottomrule
    \end{tabularx}
    \centering 
    }
    \vspace{-5pt}
    \caption{Quantitative result on diverse categories in ShapeNet. Note that all CD is scaled by 1000 and $\#$ Box denotes average number of primitives.}
    \label{tbl:other_cat}
\end{table}

%% file: sections/6_Conclusion.tex
\vspace{-3pt}
\section{Conclusion}
We presented~\smart{}, a novel framework for finding tight bounding boxes of 3D shapes with pre-segment-based over-segmentation and multiple iterative searches. The framework consists of four steps: 1) \textbf{over-segmentation} using pre-segment and post-processing, 2) \textbf{merging} in a hierarchical way with tightness-aware criteria, 3) \textbf{refinement} with a discrete action space with a soft reward function, and 4) \textbf{additional refinement} via multi-action space exploration using MCTS and its acceleration. The experimental results showed the tightness and parsimony of our bounding boxes that also guarantee the full coverage of the shape.

\paragraph{Limitation.}
\vspace{-\baselineskip}

We have successfully demonstrated the robustness and versatility of our framework across various categories and datasets, all without necessitating any training data. However, we acknowledge certain limitations. As our framework relies on volumetric information to fit cuboids, it requires watertight tetrahedral mesh as input to accurately define the volume of the parts. Additionally, due to the nature of MCTS optimization, it solely can not guarantee $\text{Cov}(S, \{B_i\})$ to be 1 requiring additional post-processing.

\paragraph{Acknowledgement.}
This work was partly supported by NRF grant (RS-2023-00209723) and IITP grant (2022-0-00594, RS-2023-00227592) funded by the Korean government (MSIT), Seoul R\&BD Program (CY230112), and grants from ETRI, KT, NCSOFT, and Samsung Electronics.



%% file: sections/Supplementary.tex


\ifpaper
    \newcommand{\refofpaper}[1]{\unskip}
    \newcommand{\refinpaper}[1]{\unskip}
\else
  \makeatletter
  \newcommand{\manuallabel}[2]{\def\@currentlabel{#2}\label{#1}}
  \makeatother
  \manuallabel{sec:bspinit}{4.1}
  \manuallabel{sec:cluster}{4.2}
  \manuallabel{sec:refine}{4.3}
  \manuallabel{sec:mcts}{4.4}
  \manuallabel{fig:bspfail}{3}
  \manuallabel{fig:refineexp}{6}
  \manuallabel{fig:qual}{7}
  \manuallabel{eq:soft}{6}
  \manuallabel{eq:bavf}{4}
  \manuallabel{eq:hard-obj}{3}
  \newcommand{\refofpaper}[1]{of the main paper}
  \newcommand{\refinpaper}[1]{in the main paper}
\fi

\ifpaper
\else
In this supplementary material, we first demonstrate the effect of using different pre-segments as initialization (Sec.~\ref{sup:diff_init}). Then, we check the result of changing the merging threshold $\epsilon_{merge}$ that results in a trade-off between the number of bounding boxes and Tgt while showing its insensitivity to other metrics (Sec.~\ref{sup:eps}). We also illustrate the effect of $\alpha$ in the soft refinement objective by perturbing it to show the trade-off between coverage and tightness (Sec.~\ref{sup:refine}). Moreover, we provide a more detailed explanation of our algorithms used in~\smart{} (Sec.~\ref{sup:algo}). Then, we explain our acceleration techniques used in MCTS and experiments to prove them (Sec.~\ref{sup:mcts_exp}). Finally, we give more qualitative results and comparisons (Sec.~\ref{sup:more_qual}).
\fi

\subsection{Comparison of Different Initialization}
\label{sup:diff_init}

We show two possible pre-segment initialization method introduced in Sec.~\ref{sec:bspinit}~\refofpaper{} and it's quantitative result. One is the output of BSP-Net~\cite{chen20bspnet} and the other is convex decomposition method~\cite{wei22approximate}. Tab.~\ref{tbl:diff_init} shows the quantitative results with Tables, Chairs, and Airplanes in ShapeNet~\cite{shapenet2015} averaged all together. It demonstrates that our method is \emph{not} sensitive to the choice of the pre-segment in the initialization and other pre-segments can also be used with our algorithm.
\input{rebuttal/tables/init}
\vspace{-\baselineskip}

\subsection{Effect of $\epsilon_{merge}$ in Hierarchical Merging (Sec.~\ref{sec:cluster} in the Main Paper)}
\label{sup:eps}

To investigate the trade-off between reducing the number of boxes and minimizing Tgt in the merging step (Sec.~\ref{sec:cluster}~\refinpaper{}), we varied $\epsilon_{merge}$ to $(0, -0.004, -0.02, -0.1)$ on the Table category of ShapeNet~\cite{shapenet2015}. The quantitative results are presented in Tab.~\ref{sup:tbl:eps}. It is observed that reducing $\epsilon_{merge}$ decreases the number of bounding boxes while slightly worsening other volumetric metrics showing the trade-off. We chose our $\epsilon_{merge}$ as $-0.02$ because it has the best mAP value meaning that it has the best alignment with human perception giving us more plausible part-level segmentation with adequate number of boxes. However, we can notice that other metrics remain similar by changes of $\epsilon_{merge}$ indicating that intensive hyper-parameter tuning is not necessary for our merging step.

\input{tables/supp/epsilon}

\subsection{Effect of $\alpha$ in Bounding Box Refinement (Sec.~\ref{sec:refine} in the Main Paper)}
\label{sup:refine}

We examine the benefits of converting the hard constraint objective (Eq.~\ref{eq:hard-obj}~\refinpaper{}) to the soft constraint objective (Eq.~\ref{eq:soft}~\refinpaper{}) during the refinement step. We begin by formally introducing hard refinement, which uses a one-step greedy approach with the hard constraint. We then present the outcomes of both hard refinement and soft refinement using different values of $\alpha$ to illustrate their impact. 

\paragraph{Hard Refinement.}
A simple one-step greedy approach, hard refinement can be employed by finding the optimal actions that reduce Tgt($S, \{B_i\}$) where $\text{Cov}(S, \{B_i\})$ is 1. Although this approach can enhance the overlaps of bounding boxes, it cannot address scenarios like Fig.~\ref{fig:refineexp}-a~\refinpaper{}, where breaking the $\text{Cov}(S, \{B_i\})$ constraint is necessary to refine the boxes along the action sequence.

\paragraph{Effect of $\alpha$.}
To investigate the trade-off between coverage and tightness, we demonstrate the result of changing the $\alpha$ to $\{1, 10, 100, 1000, \infty\}$ on ShapeNet~\cite{shapenet2015} table category. Note that $\infty$ represents the result for the hard refinement. Tab.~\ref{sup:tbl:alpha} shows the quantitative results. 

\input{tables/supp/alpha}

By using various $\alpha$, we can analyze the trade-off between Cov and other volumetric metrics, such as Tgt, TOV, MOV, and VIoU. Employing hard refinement with $\alpha=\infty$ ensures a $\text{Cov}(S, \{B_i\})$ of 1 for all the shapes. Soft refinement ($\alpha \neq \infty$), on the other hand, may result in rare failures where $\text{Cov}(S, \{B_i\})$ may not reach exactly 1 for some of the shapes while most of the shapes still satisfy the full coverage. However, it generally leads to an overall increase in quality, as illustrated in Tab.~\ref{sup:tbl:alpha}. Typically, these rare failures occur in situations where bounding boxes fail to cover small, detailed parts with a small volume in the shape. Still, for those rare failure cases, we can simply use the post-processing introduced in the main paper to guarantee that all the shapes have full coverage. This soft refinement objective (Eq.~\ref{eq:soft}~\refinpaper{}) provides a chance to fix the wrong bounding boxes resulting in more tightness.








\subsection{Algorithm Details}
\label{sup:algo}

In this section, we provide a detailed explanation of our initialization (Sec.~\ref{sec:bspinit}~\refinpaper{}) and hierarchical merging (Sec.~\ref{sec:cluster}~\refinpaper{}) and MCTS (Sec.~\ref{sec:mcts}~\refinpaper{}) algorithm in the following sections.

\subsubsection{Initialization via Over-Segmentation (Sec.~\ref{sec:bspinit} in the Main Paper)}
\label{sup:init}
We start by recapping the three limitations of pre-segments. Then, we explain how we address these limitations and give a detailed explanation of our algorithm.

The limitations of the pre-segments are: (1) uncovered parts, (2) overlapping segments, and (3) the merging of closely located parts. To address these issues, our algorithm treats the centroids of tetrahedral mesh covered by the pre-segments (\emph{main-part}) and the uncovered centroids (\emph{sub-part}) differently to deal with the first limitation. We first partition the \emph{main-part} and then handle the \emph{sub-part} partition, as the \emph{sub-part} represents the failure cases of pre-segments. To resolve the second limitation, we separate each overlap generated by pre-segments into different partitions in the \emph{main-part}. For the third limitation, we observe that separated parts in the shape merged by pre-segments typically resemble the case shown at Fig.~\ref{fig:bspfail}-b~\refinpaper{}. In such cases, if the upper and lower segments are correctly predicted, we can separate the parts merged by pre-segments using a depth-first search (Fig.~\ref{fig:bspfail}-c~\refinpaper{}). After partitioning all the \emph{main-part}, we merge the nearby left \emph{sub-part} to form each partition. 

\begin{algorithm}[ht]
\caption{Computing Over-Segmentation using Pre-Segments}\label{alg:init}
\DontPrintSemicolon
    \KwInput{Pre-segments $\{Q_i\}_{i=1}^N$, tetrahedral mesh $P$ having $\{c_j\}_{j=1}^M$ tetrahedrals}
    \KwOutput{Initial over-segmentation $\{S_k\}_{k=1}^K$ of $P$}

    \SetKwFunction{MainDFS}{MainDFS}
    \SetKwProg{Fn}{Function}{:}{}
    \Fn{\MainDFS{$j$, $k$, $mask$}}{
        $visited[j]$ $\leftarrow$ True \;
        $S_k \leftarrow S_k + \{c_j\}$ \;
        \For{$id \in P.nearby(c_j)$}    
        { 
            \If{$mask[id]$ == $mask[j]$ and not $visited[id]$}
            {
                \MainDFS{$id$, $k$}
            }
        }
    }

    \SetKwFunction{SubDFS}{SubDFS}
    \SetKwFunction{GetSDFMask}{GetSDFMask}
    \Fn{\SubDFS{$j$, $k$}}{
        $visited[j]$ $\leftarrow$ True \;
        $S_k \leftarrow S_k + \{c_j\}$ \;
        \For{$id \in P.nearby(c_j)$}    
        { 
            \If{not $visited[id]$}
            {
                \SubDFS{$id$, $k$}
            }
        }
    }
    $k = 1$\;
    \For{$j \in \{1...M\}$}    
    { 
        $visited[j]$ $\leftarrow$ False\\
        $mask[j]$ $\leftarrow$ \GetSDFMask($c_j, Q$)\label{alg:masksdf}
    }
    \For{$j \in \{1...M\}$} 
    {
        \If{not $visited[j]$ and $mask[j] \neq 0$}
        {
            $S_k \leftarrow \emptyset $\;
            \MainDFS{$j$, $k$, $mask$}\;\label{alg:maindfs}
            $k \leftarrow k + 1$\;
        }
    }
    \For{$j \in \{1...M\}$} 
    {
        \If{not $visited[j]$}
        {
            $S_k \leftarrow \emptyset $\;
            \SubDFS{$j$, $k$}\;\label{alg:mergesub}
            $k \leftarrow k + 1$\;
        }
    }
\end{algorithm}

These ideas are directly implemented in Algorithm~\ref{alg:init}. We separate the centroids into \emph{main-part} and \emph{sub-part} by calculating the SDF of centroids $\{c_i\}_{i=1}^M$ of P with each pre-segments $\{Q_j\}_{j=1}^N$. To determine which pre-segments $Q_j$ cover the centroid $c_i$, we obtain a length N boolean mask. We mark the boolean mask to 1 if SDF value of $c_i$ and $Q_j$ is positive (Line~\ref{alg:masksdf} in Algorithm~\ref{alg:init}). These masks separate every overlap generated by pre-segments into a different partition. Then we regroup the \emph{main-part} using \MainDFS that have the same $mask$ (Line~\ref{alg:maindfs} in Algorithm~\ref{alg:init}). Finally, we merge the left nearby \emph{sub-part} using \SubDFS to obtain the initial over-segmentation (Line~\ref{alg:mergesub} in Algorithm~\ref{alg:init}).

\subsubsection{Hierarchical Merging (Sec.~\ref{sec:cluster} in the Main Paper)}
\label{sup:merge}

As shown in Algorithm~\ref{alg:merge}, we first start by calculating Bounding-box-Aware Volume Function (BAVF, Eq.~\ref{eq:bavf}~\refinpaper{}) for all the partition pairs. Then, we select the pair that has the maximum value of BAVF. If such value is larger than $\epsilon_{merge}$, we merge the partitions and if not, we terminate our hierarchical merging algorithm. Since merging of $S_i$ and $S_j$ does not change the BAVF value for other pairs, we effectively cache those values to reduce the time complexity of $\mathcal{O}(N^3)$ to $\mathcal{O}(N^2)$.

\begin{algorithm}
\caption{Hierarchical Clustering with BAVF}\label{alg:merge}
\DontPrintSemicolon
    \KwInput{Initial over-segmentation $S = \{S_i\}_{i=1}^K$}
    \KwOutput{Merged part-level partition $S = \{S_i\}_{i=1}^{N}$}
    \SetKwFunction{BAVF}{BAVF}
    
    \While{True}
    {
        $n \leftarrow len(S)$ \;
        \For{$i,j \leq n$}    
        { 
            Calculate $\BAVF(S_i, S_j)$
        }

        \If{$\max_{i \neq j}\BAVF(S_i, S_j) < \epsilon_{merge}$\label{alg:mergethresh}}
        {
            \textbf{break}
        }
        $i^*, j^* \leftarrow \argmax_{i \neq j} \BAVF(S_i, S_j)$ \;
        Remove $S_{i^*}, S_{j^*}$ from $S$. \;
        Add $S_{i^*} \cup S_{j^*}$ to $S$. \;
    }
\end{algorithm}

\subsubsection{Monte Carlo Tree Search (Sec.~\ref{sec:mcts} in the Main Paper)}
\label{sup:mcts}
\SetKwFunction{Score}{Score}

As Algorithm~\ref{alg:mcts} shows, our MCTS consists of iteratively running $iter$ iterations of the four steps: (1) selection, (2) expansion, (3) evaluation, and (4) backpropagation similar to the original MCTS~\cite{mcts} (Line~\ref{alg:mcts_init} in Algorithm~\ref{alg:mcts}). We start from the root node $n_0$ corresponding to the initial bounding boxes $\{B_i\}^0$ we want to tighten, using less than $T$ actions. In each node, we save the information of $Q$($\cdot$) which is the value function, where $N$($\cdot$) indicates the number of visits and its expanded child nodes and untried actions.

In the selection and expansion step, we select the node to expand and expand it. To select the node, we start from the root node and choose the child node with the best UCB~\cite{ucb}:
\begin{align}
    \text{UCB}(n) = Q(n) + c\sqrt{\frac{2\ln N(n_p)}{N(n)}}
    \label{eq:ucb}
\end{align}
if all the possible actions had been tried (in other words, all of its child nodes are expended) (Line~\ref{alg:mcts_sel_ucb} in Algorithm~\ref{alg:mcts}). Note that $n$ is the node we want to calculate and $n_p$ is the parent node of $n$. If there is an untried action, we randomly select an untried action and expand that node as its child node (Line~\ref{alg:mcts_sel_exp} in Algorithm~\ref{alg:mcts}). When expanding the node, we set the untried actions as all the possible actions except that to go back to the parent state to remove unnecessary expansion.

During the evaluation step, evaluating the expanded node is a non-trivial task when applying MCTS to our problem. To address this challenge, we employ a one-step greedy approach where we try out all possible actions on the expanded node and select the best action iteratively by computing its corresponding $\Score(\{B_i\}^0, \{B_i\}^t)$ (Line~\ref{alg:mcts_grd} in Algorithm~\ref{alg:mcts}, Eq.~\ref{eq:score}), which reflects the improvement from the initial bounding boxes. Note that $\Score(\{B_i\}^0, \{B_i\}^t)$ is calculated as:
\begin{align}
    \Score(\{B_i\}^0, &\{B_i\}^t) = \nonumber \\
    &\text{Tgt}(S, \{B_i\}^0)-\text{Tgt}(S, \{B_i\}^t) \nonumber \\ &-\alpha \text{Cov}(S, \{B_i\}^0) +\alpha\text{Cov}(S, \{B_i\}^t).
    \label{eq:score}
\end{align}

If the action does not improve the $\Score$, we terminate the evaluation process (Line~\ref{alg:mcts_term} in Algorithm~\ref{alg:mcts}).

The last backpropagation step is simple and intuitive. We simply update the number of visits $N$($\cdot$) and value function $Q$($\cdot$) along the path (Line~\ref{alg:backup} in Algorithm~\ref{alg:mcts}).

\subsection{MCTS Acceleration Techniques and Ablation Study}
\label{sup:mcts_exp}

In this section, we discuss our dedicated acceleration techniques used in MCTS and demonstrate its performance through an ablation study.
\vspace{-\baselineskip}
\paragraph{Greedy Bounding Box Pruning.}
The primary time constraint in the MCTS is selecting the greedy action during node evaluation (Line~\ref{alg:mcts_grd} in Algorithm~\ref{alg:mcts}), which involves testing all potential one-step actions. To optimize this process and improve evaluation efficiency, we calculate the greedy action by the unit of each bounding box. If all the actions corresponding to the i-th bounding box fail to yield any improvement, we do not attempt any action on that bounding box until the end of the evaluation.
\vspace{-\baselineskip}
\paragraph{Evaluation Expansion (EE).}
To reduce the number of greedy actions, we not only expand the node in the selection step but also expand the node at the evaluation step (Line~\ref{alg:grdexp} in Algorithm~\ref{alg:mcts}). However, this expansion differs from normal expansion as it only permits the previously taken greedy action at that node. This technique has the same outcome as saving the greedy action at each node, thereby reducing the time of calculating repetitive greedy actions as the time step increases.
\vspace{-\baselineskip}
\paragraph{Prioritized Node Selection (PNS).}
Since we are using discrete unit actions to refine the bounding boxes, there is a possibility that applying an action that results in a $score$ increase may do so again if applied repeatedly. Therefore, prioritizing such actions during node expansion in the selection step can facilitate faster optimization than random node expansion, by deepening the tree instead of widening it.

\begin{algorithm}[h!]
\caption{MCTS for Tight Bounding Boxes}\label{alg:mcts}
\DontPrintSemicolon
    \SetKwProg{Fn}{Function}{:}{}
    \SetKwFunction{MCTS}{MCTS}
    \SetKwFunction{Select}{Select}
    \SetKwFunction{Eval}{Eval}
    \SetKwFunction{Backup}{Backup}
    
    \Fn{\MCTS{$\{B_i\}$, iter, T}}{\label{alg:mcts_init}
        Create root node $n_0$ with initial bounding boxes $B^0 = \{B_i\}^0$\;
        \While{iter iterations}
        {
            $\{B^1, \cdots, B^s \}, n_s \leftarrow$ \Select{$n_0, T$}\;
            $\{B^{s+1}, \cdots, B^t \}, n_t \leftarrow$ \Eval{$n_s, T$}\;
            $r \leftarrow \Score(B^0, B^t)$\label{alg:score} where \Score is Eq.~\ref{eq:score}.\;
            \Backup{$n_t, r$}\;
        }
        $n^{*} \leftarrow \underset{n_c \in \text{ children of } n_0}\argmax Q(n_c)$\;
        $\KwRet$ Action sequence from $n_0$ to $n^{*}$\;
    }
    
    \Fn{\Select{$n$, $T$}}{
        $P \leftarrow \emptyset$ \;
        \While{$depth(n) < T$}
        {
            $t = depth(n)$\;
            \If{all node of $n$ are expanded}
            {
                \label{alg:mcts_sel_ucb}
                $n_{bst}, a_{bst} \leftarrow $ Select the best child and action of $n$ by the UCB($n$) (Eq.~\ref{eq:ucb}).\;
                Apply action $a_{bst}$ to the bounding boxes $B^{t}$ to get $B^{t+1}$.\;
                $P, n \leftarrow P + \{B^{t+1}\}, n_{bst}$\;
            }
            \Else
            {
                \label{alg:mcts_sel_exp}
                Randomly select a untried action $a_{u}$.~\label{alg:rand_action}\;
                Apply action $a_u$ to the bounding boxes $B^{t}$ to get $B^{t+1}$.~\label{alg:a_u}\;
                Create a new child $n_{chd}$ with $B^{t+1}$ to $n$.\;
                $\KwRet \; P + \{B^{t+1}\}, n_{chd}$
            }
        }
        $\KwRet \; P, n$
    }
    \Fn{\Eval{$n, T$}}{
        $P \leftarrow \emptyset$ \;
        \While{$depth(n) < T$}
        {
            $a_{grd} \leftarrow $ Select one-step greedy action on $n$.~\label{alg:mcts_grd}\;
            Apply action $a_{grd}$ to the bounding boxes $B^{t}$ to get $B^{t+1}$.~\label{alg:grd}\;
            \If{$\Score(B^0, B^{t+1}) \leq \Score(B^0, B^t)$}
            {
                \label{alg:mcts_term}   
                \textbf{break}
            }
            Create a new child $n_{grd}$ with $B^{t+1}$ to $n$ while removing all untried actions of $n$.~\label{alg:grdexp}\;
            $P, n \leftarrow P + \{B^{t+1}\}, n_{grd}$\;
        }
        $\KwRet \; P, n$
    }

    \Fn{\Backup{$n, r$}}{
        \label{alg:backup}
        \While{$n$ is not null}
        {
            $N(n) \leftarrow N(n) + 1$\;
            $Q(n) \leftarrow \max(Q(n), r)$\;
            $n \leftarrow$ Parent of $n$\;
        }
    }
\end{algorithm}

To implement this technique, we calculate the $score$ increase for each expansion action ($a_u$ at Line~\ref{alg:a_u} in Algorithm~\ref{alg:mcts}) and increase the probability of random selection for expansion (Line~\ref{alg:rand_action} in Algorithm~\ref{alg:mcts}) proportional to the past score increase history while allowing to skip the untried actions of the node (execute Line~\ref{alg:mcts_sel_ucb} in Algorithm~\ref{alg:mcts} though the condition is false) with some probability if it is on the path to finding the best reward. We use 0.9 for the skipping probability in our PNS.

\begin{figure}[ht]
    \vspace{-5pt}
    \centering
    \includegraphics[width=\columnwidth]{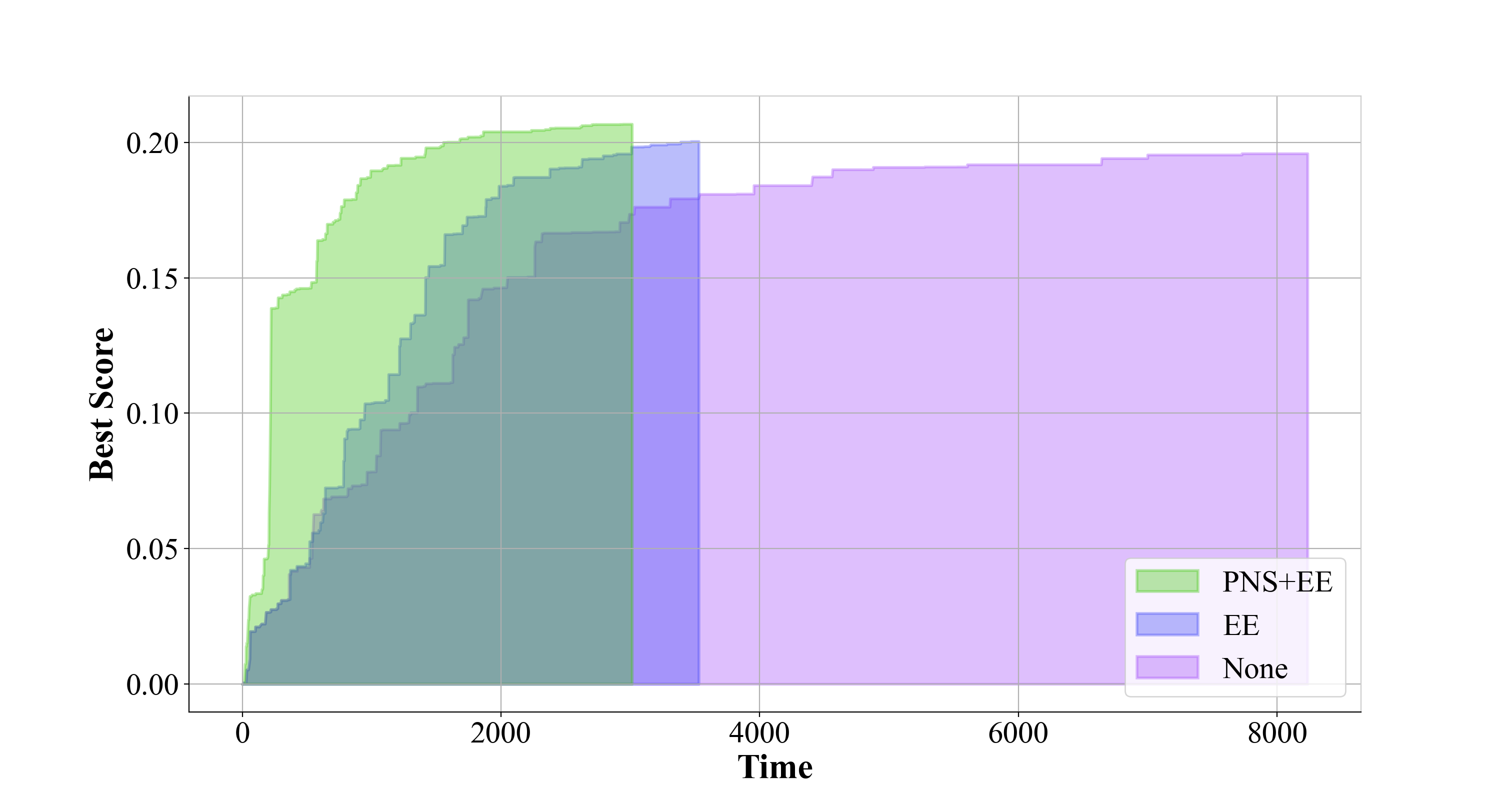}
    \vspace{-15pt}                                                                                                                
    \caption{Ablation study on the techniques used in MCTS. X-axis represents the elapsed time in seconds and Y-axis shows the found best score.}
    \label{fig:mctsalb}
    \vspace{-10pt}
\end{figure}

\paragraph{Does the Acceleration in MCTS Really Help?}
We now give an ablation study of our acceleration techniques used in MCTS. Due to the time-consuming nature without the greedy bounding box pruning, we only conducted ablations on evaluation expansion (EE) and prioritized node selection (PNS). (Greedy bounding box pruning is the default.) Fig.~\ref{fig:mctsalb} illustrates the result of the best score and elapsed time as the iteration of the tree search increases. Our EE technique allows us to reduce the required time, and PNS enables us to identify the best score in earlier iterations than without it. These techniques all bring acceleration and additional improvements.

\clearpage
\newpage
\onecolumn

\subsection{More Qualitative Results}
\label{sup:more_qual}
We provide more results of the comparison with the other methods below, as shown in Fig.~\ref{fig:qual}~\refinpaper{}. We show two lines for each shape for a better view of tightness and coverage: the upper line overlaying the input mesh and the lower line without it.
\input{figures/supp/more_comp_ov/comp}

\clearpage
\newpage
\twocolumn

%% file: rebuttal/tables/init.tex
\begin{table}[ht!]
    {\footnotesize
    \newcolumntype{Y}{>{\centering\arraybackslash}X}
    \begin{tabularx}{\linewidth}{>{\centering}m{2.0cm}|YYYYYY}
        \toprule
        & \multicolumn{5}{c}{Table, Chair, and Airplane} \\
            \cmidrule(lr){2-6} 
        Initialization & Tgt$\downarrow$ & Cov$\uparrow$ & MOV$\downarrow$ & TOV$\downarrow$ & VIoU$\uparrow$  \\
        \midrule
        BSP-Net~\cite{chen20bspnet} & 1.79 & 1.00 & 3.03 & 0.72 & 0.63\\
        CoACD~\cite{wei22approximate} & 1.82 & 1.00 & 2.75 & 0.76 & 0.62\\
        \bottomrule
    \end{tabularx}
    \centering 
    }
    \caption{Comparison of \textsc{SMART} by using different initializations. Both initializations are processed by Merge+Ref.+MCTS.}
    \label{tbl:diff_init}
\end{table}


%% file: tables/supp/epsilon.tex
\begin{table}[ht!]
    \centering
    {\footnotesize
    \newcolumntype{Y}{>{\centering\arraybackslash}X}
    \newcolumntype{Z}{>{\centering\arraybackslash}m{1.2cm}}
    \begin{tabularx}{\linewidth}{>{\centering}m{1.0cm}|ZYYYYYY}
        \toprule 
        $\epsilon_{merge}$ & $\#$ boxes$\downarrow$ & mAP$\uparrow$ & Tgt$\downarrow$ & Cov$\uparrow$ & MOV$\downarrow$ & TOV$\downarrow$ & VIoU$\uparrow$ \\
        \midrule
        0 & 14.25 & 0.57 & 2.20 & 1.00 & 3.89 & 1.01 & 0.61 \\
        -0.004 & 11.41 & 0.56 & 2.19 & 1.00 & 3.95 & 1.02 & 0.61\\
        -0.02 & 8.89 & \textbf{0.58} & 2.21 & 1.00 & 3.95 & 1.04 & 0.60\\
        -0.1 & 6.59 & 0.57 & 2.30 & 1.00 & 4.16 & 1.15 & 0.57 \\
        \bottomrule
    \end{tabularx}
    \centering 
    }
    \vspace{-5pt}
    \caption{Effect of $\epsilon_{merge}$ in merging step (Sec.~\ref{sec:cluster}~\refinpaper{}) at ShapeNet~\cite{shapenet2015} Table category. The best mAP is marked in \textbf{bold} showing the best alignment with human perception.}
    \label{sup:tbl:eps}
\end{table}

%% file: tables/supp/alpha.tex
\begin{table}[ht!]
    \centering
    {\footnotesize
    \newcolumntype{Y}{>{\centering\arraybackslash}X}
    \newcolumntype{Z}{>{\centering\arraybackslash}m{1.2cm}}
    \begin{tabularx}{\linewidth}{>{\centering}m{1.0cm}|ZYYYYY}
        \toprule 
        $\alpha$ & $\#$ boxes$\downarrow$ & Tgt$\downarrow$ & Cov$\uparrow$ & MOV$\downarrow$ & TOV$\downarrow$ & VIoU$\uparrow$ \\
        \midrule
        1 & 8.37 & 1.00 & 0.7628 & $-$ & $-$ & 0.66\\
        10 & 7.77 & 1.43 & 0.9887 & 1.34 & 0.42 & 0.74\\
        100 & 7.90 & 1.72 & 0.9996 & 2.85 & 0.69 & 0.67\\
        1000 & 8.09 & 1.85 & 0.9999 & 4.30 & 0.81 & 0.66\\
        $\infty$(Hard) & 8.41 & 2.02 & 1.0000 & 5.18 & 0.97 & 0.62\\
        \bottomrule
    \end{tabularx}
    \centering 
    }
    \vspace{-5pt}
    \caption{Effect of $\alpha$ in the refinement step (Sec.~\ref{sec:refine}~\refinpaper{}) at ShapeNet~\cite{shapenet2015} Table category.}
    \label{sup:tbl:alpha}
\end{table}

%% file: figures/supp/more_comp_ov/comp.tex
\newcommand{\AllComparisonImages}{\input{figures/supp/more_comp_ov/image_list.tex}}
\graphicspath{{figures/supp/more_comp_ov/}}

\makeatletter
\def\Image#1{%
  \multicolumn{\LT@cols}{l}{\includegraphics[width=\textwidth]{#1}}\\
  \midrule
}
\makeatother

\setlength{\tabcolsep}{0em}
\def\arraystretch{0.0}
\newcolumntype{Z}{>{\centering\arraybackslash}m{0.125\textwidth}}
{\scriptsize
\begin{longtable}{Z|Z|Z|Z|Z|Z|Z|Z}
Input &
HA~\cite{sun19abstract} &
HA~\cite{sun19abstract}+Ref. &
CA~\cite{yang21cubseg} &
CA~\cite{yang21cubseg}+Ref. &
Merge &
Merge+Ref. &
\makecell{Merge+Ref.\\+MCTS} \\
  \midrule
  \endhead
  
  \endfoot

  \input{figures/supp/more_comp_ov/image_list.tex}
\end{longtable}
}

%% file: figures/supp/more_comp_ov/image_list.tex
\Image{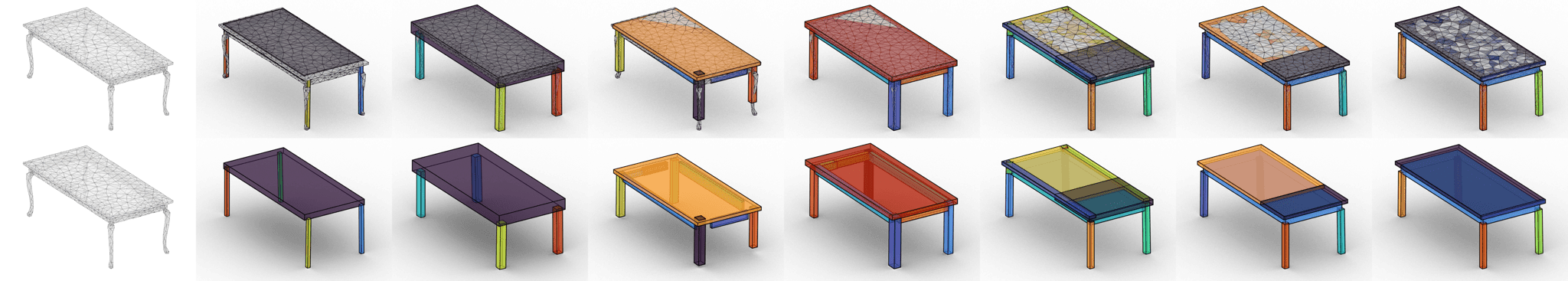}
\Image{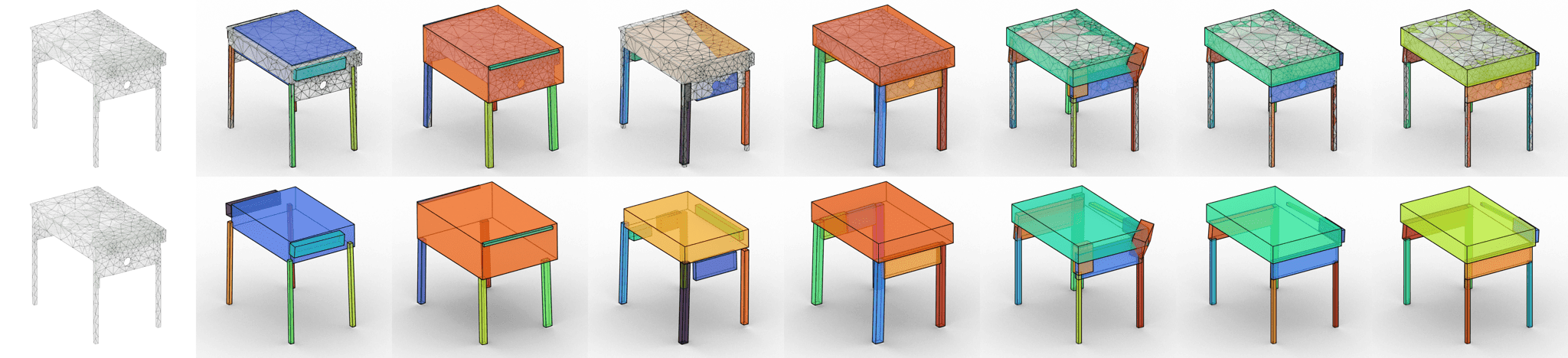}
\Image{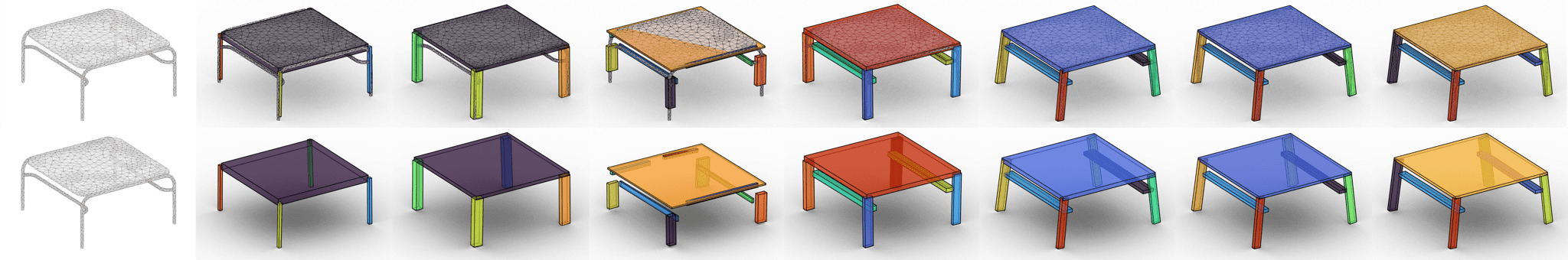}
\Image{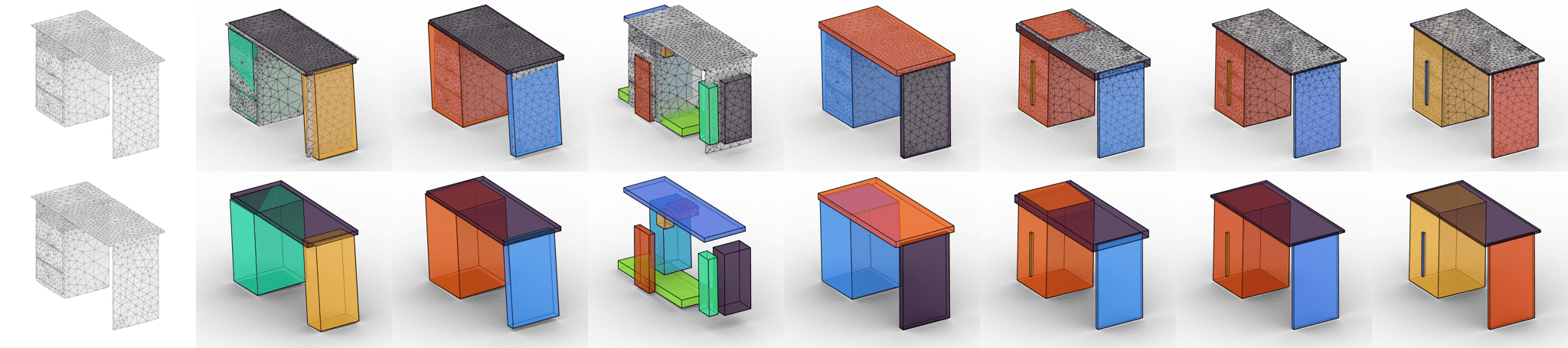}
\Image{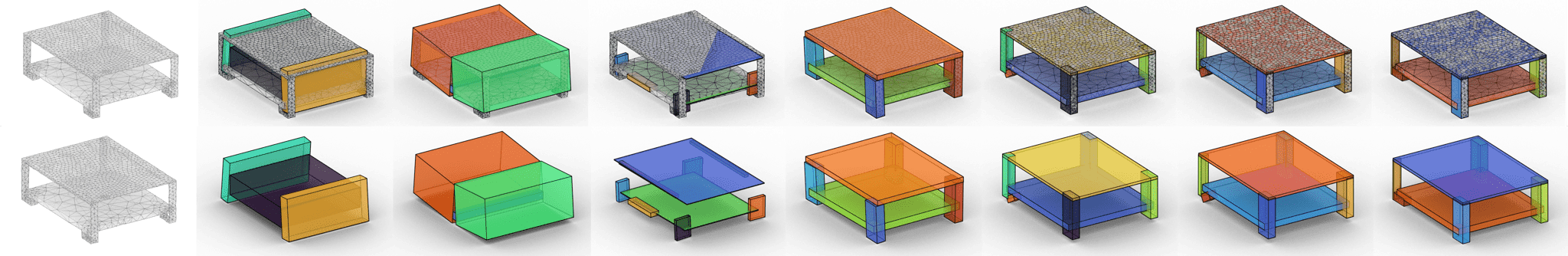}
\Image{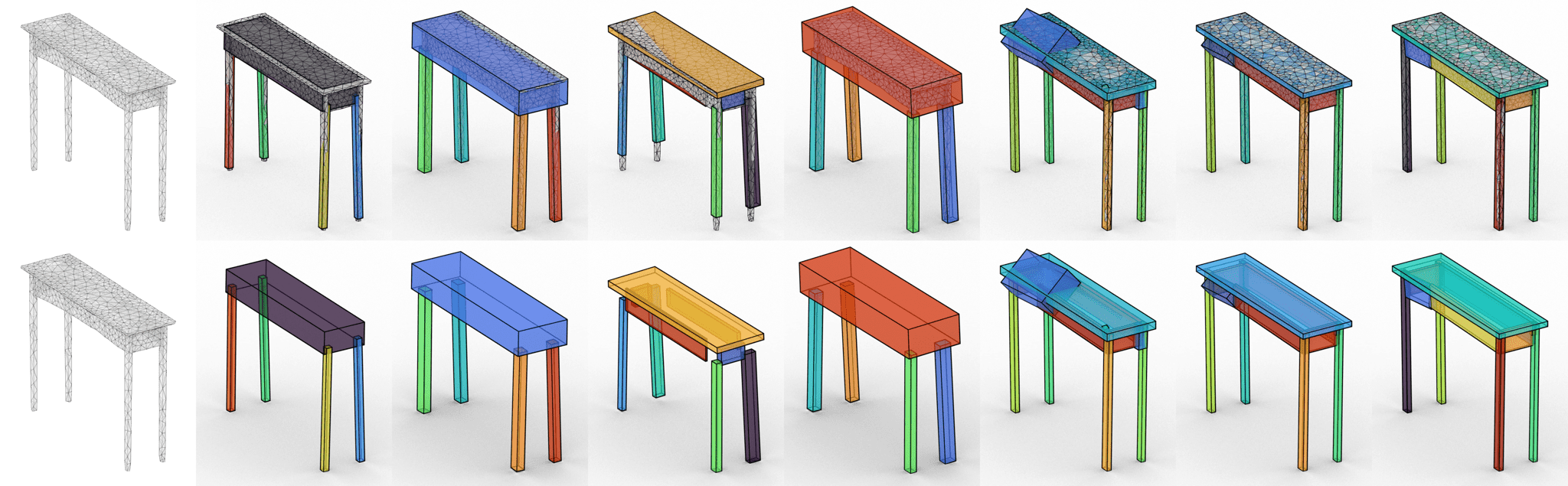}
\Image{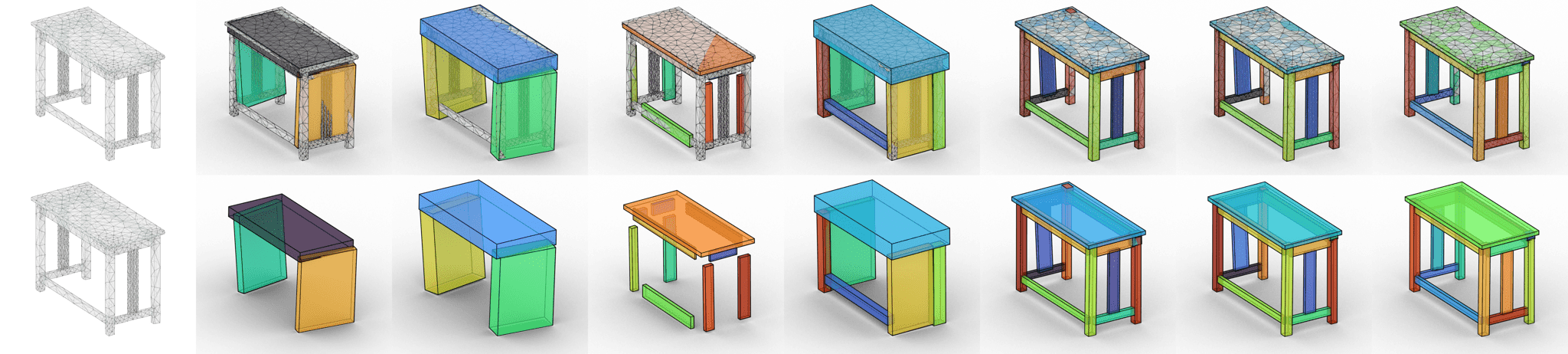}
\Image{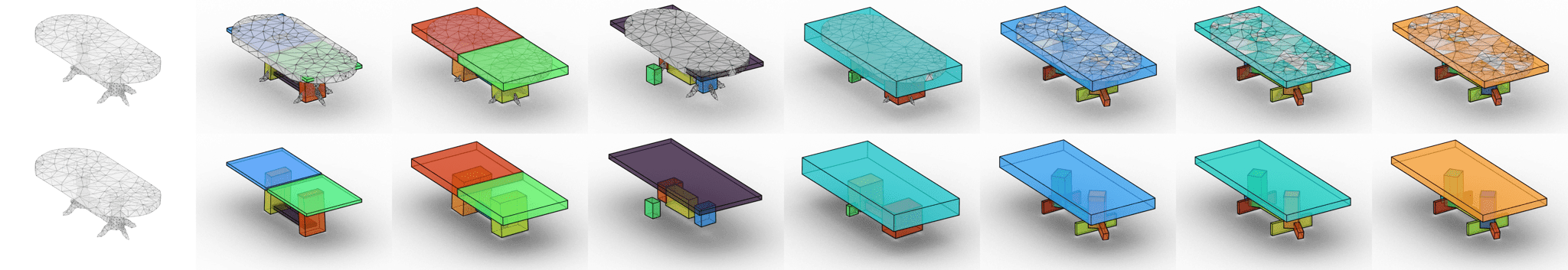}
\Image{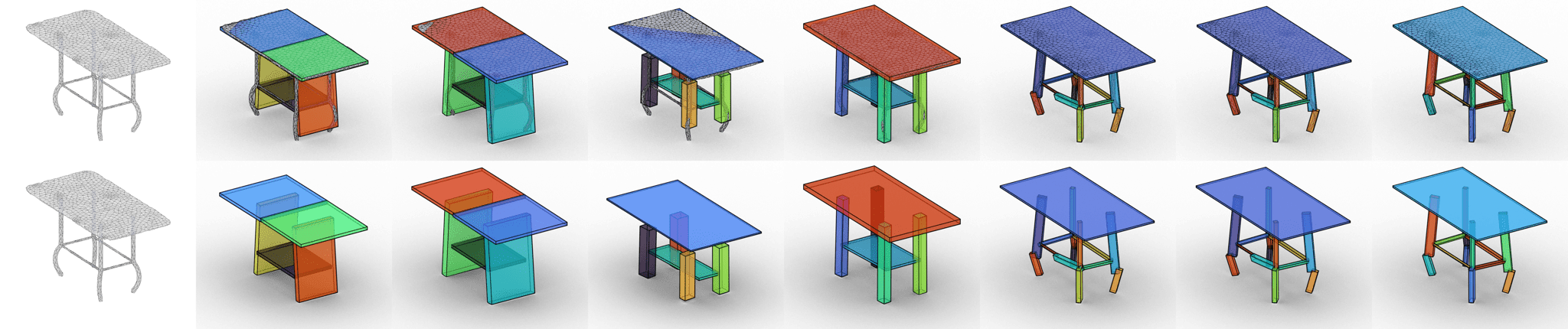}
\Image{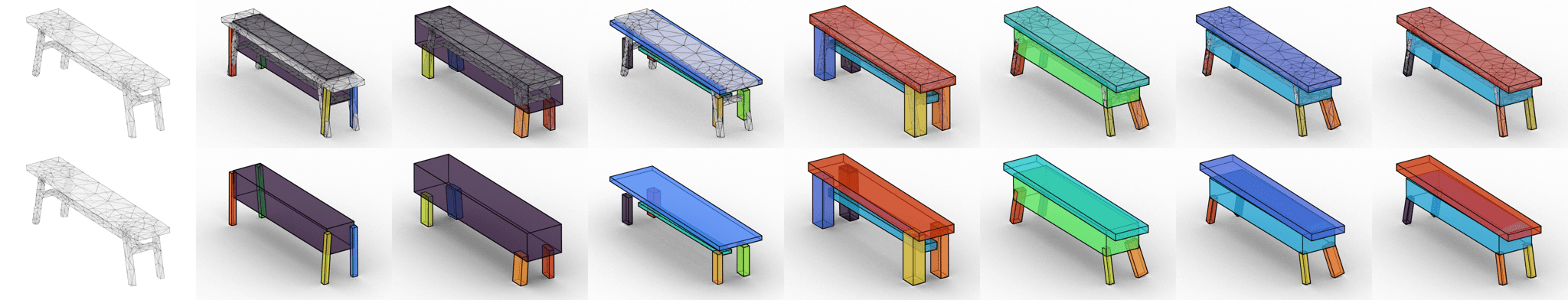}
\Image{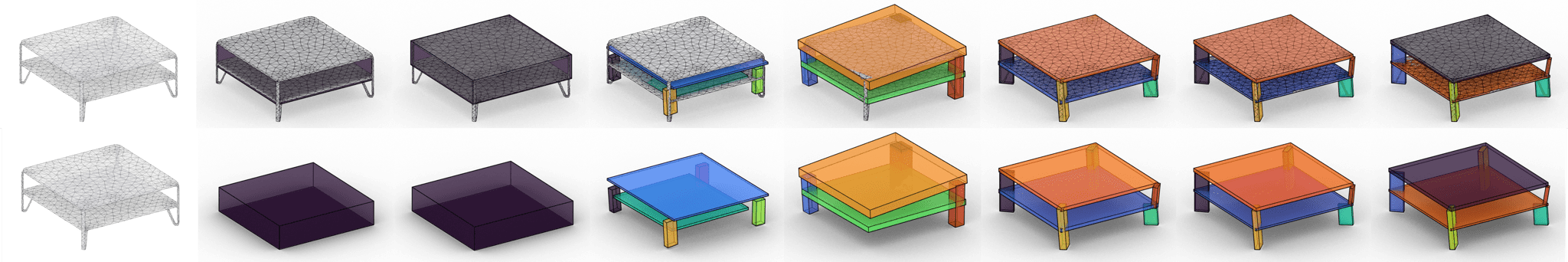}
\Image{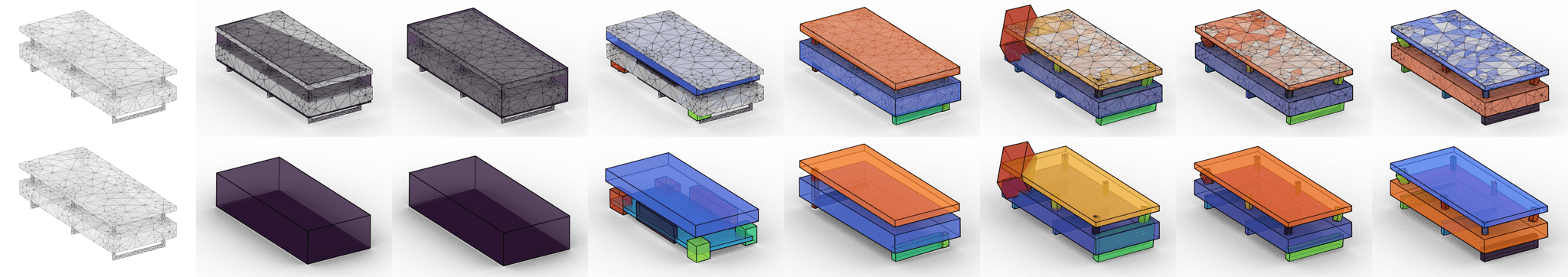}
\Image{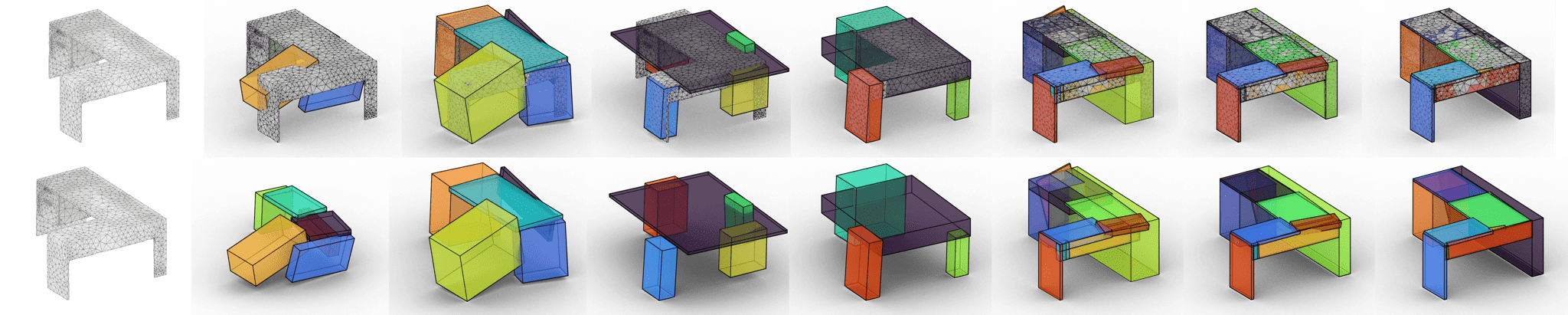}
\Image{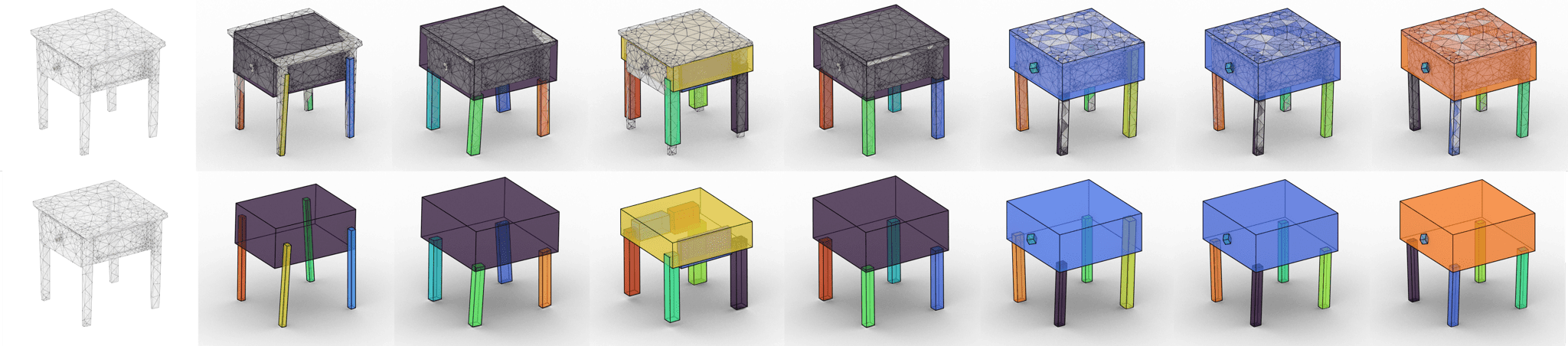}
\Image{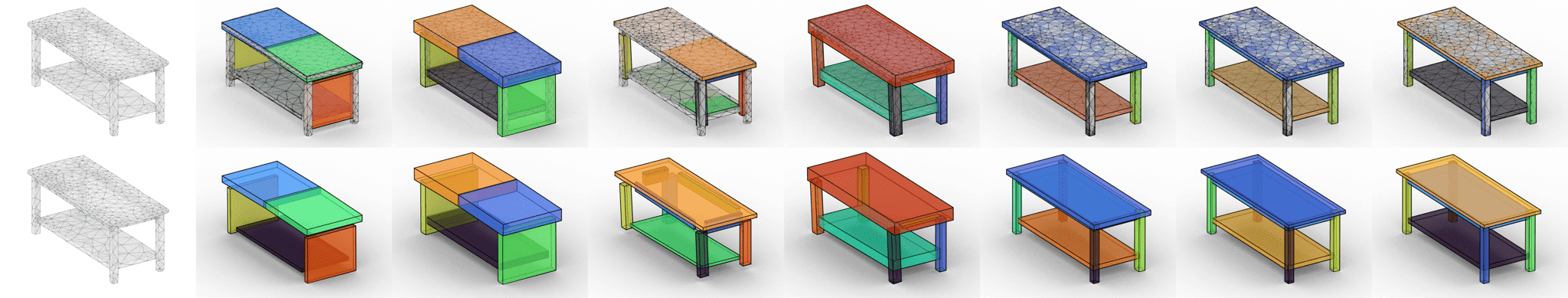}
\Image{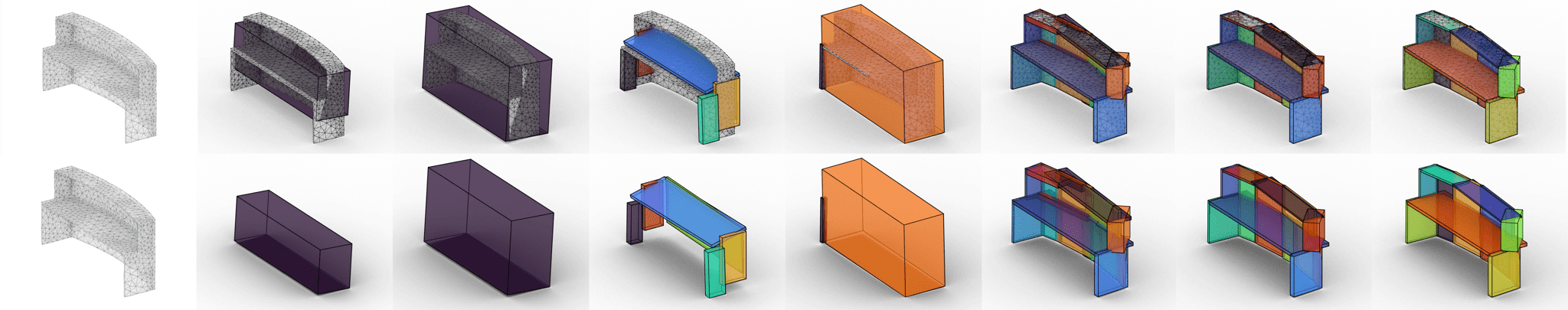}
\Image{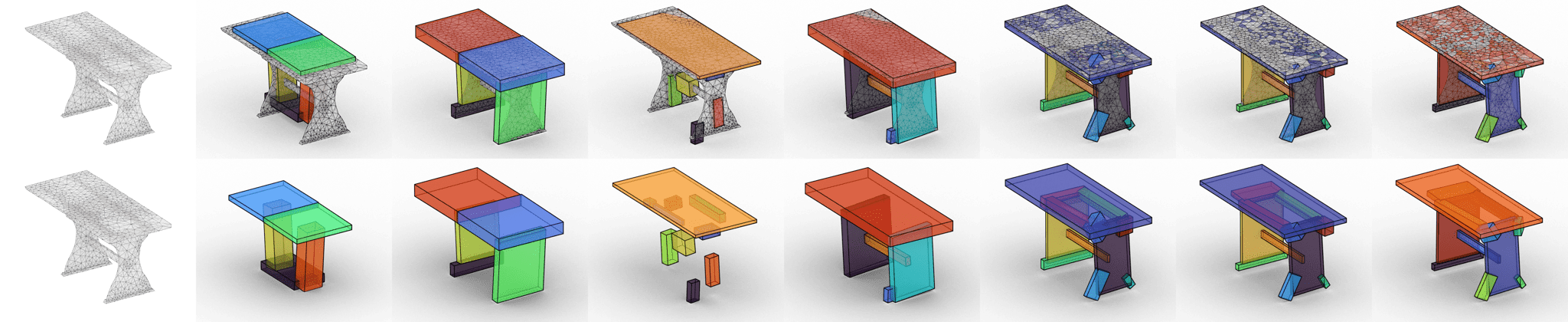}
\Image{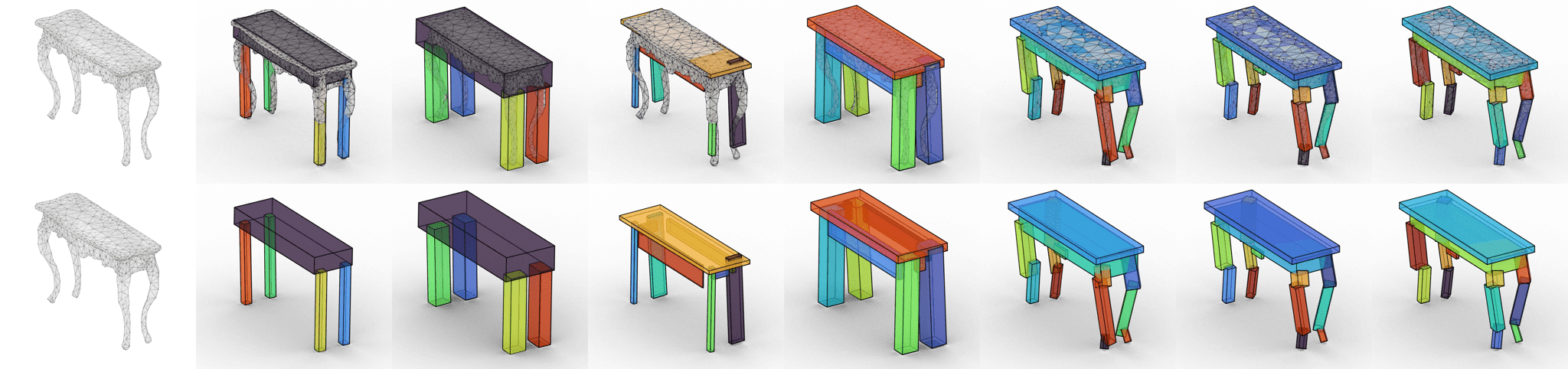}
\Image{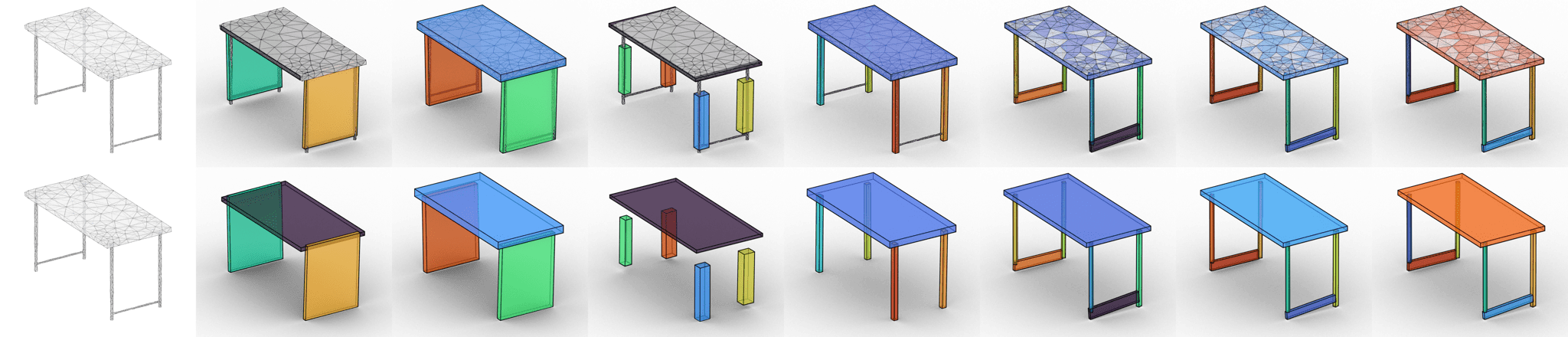}
\Image{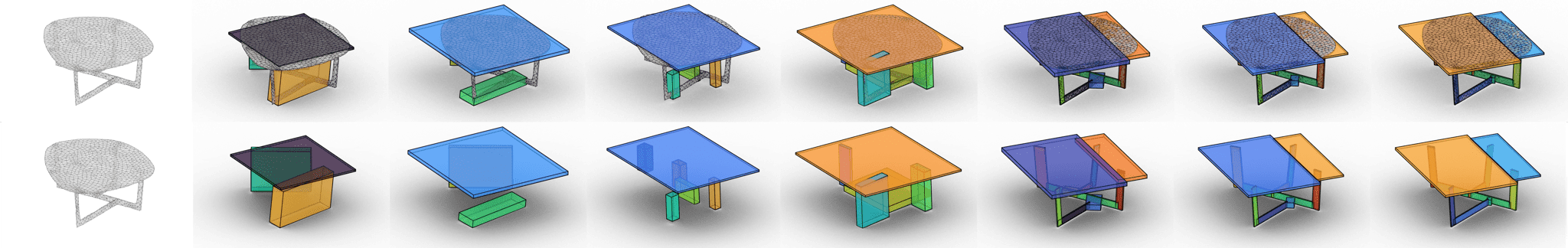}
\Image{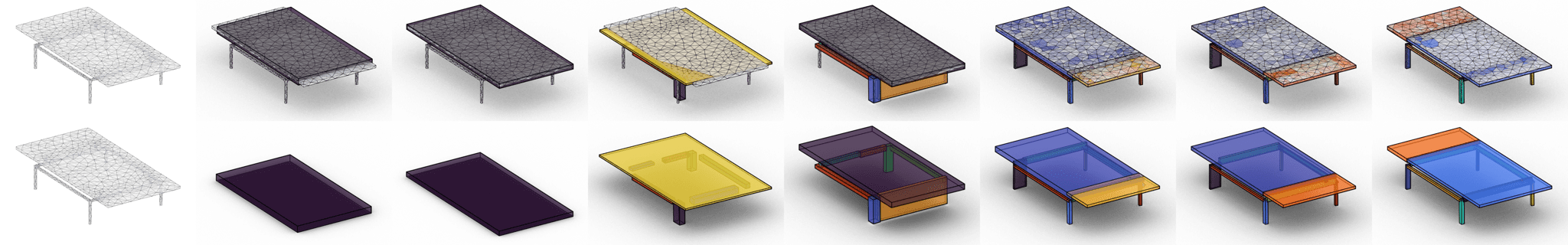}
\Image{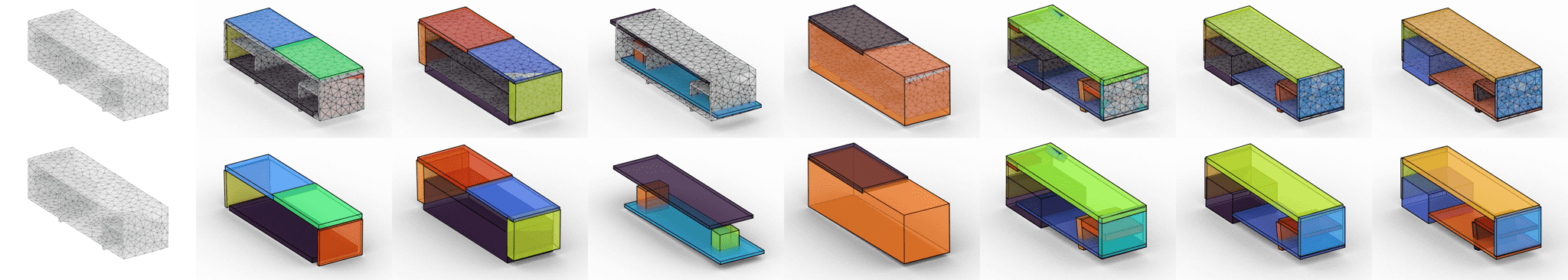}
\Image{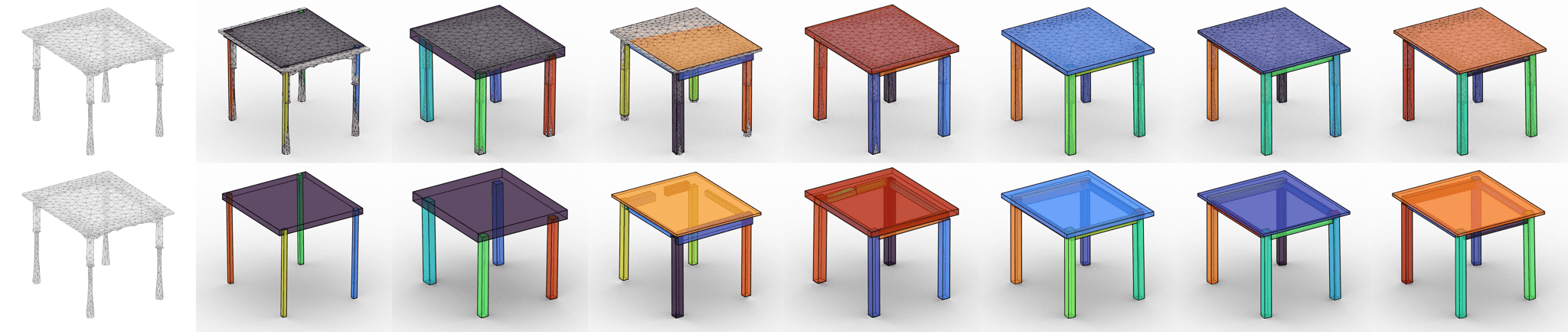}
\Image{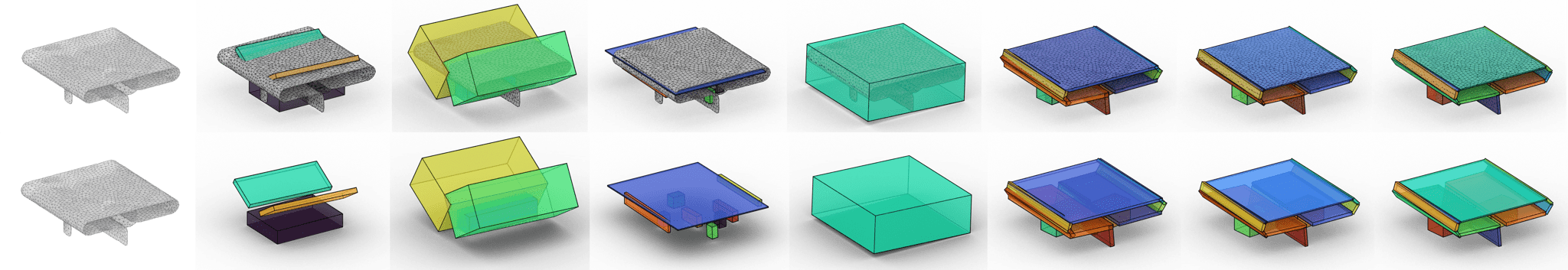}
\Image{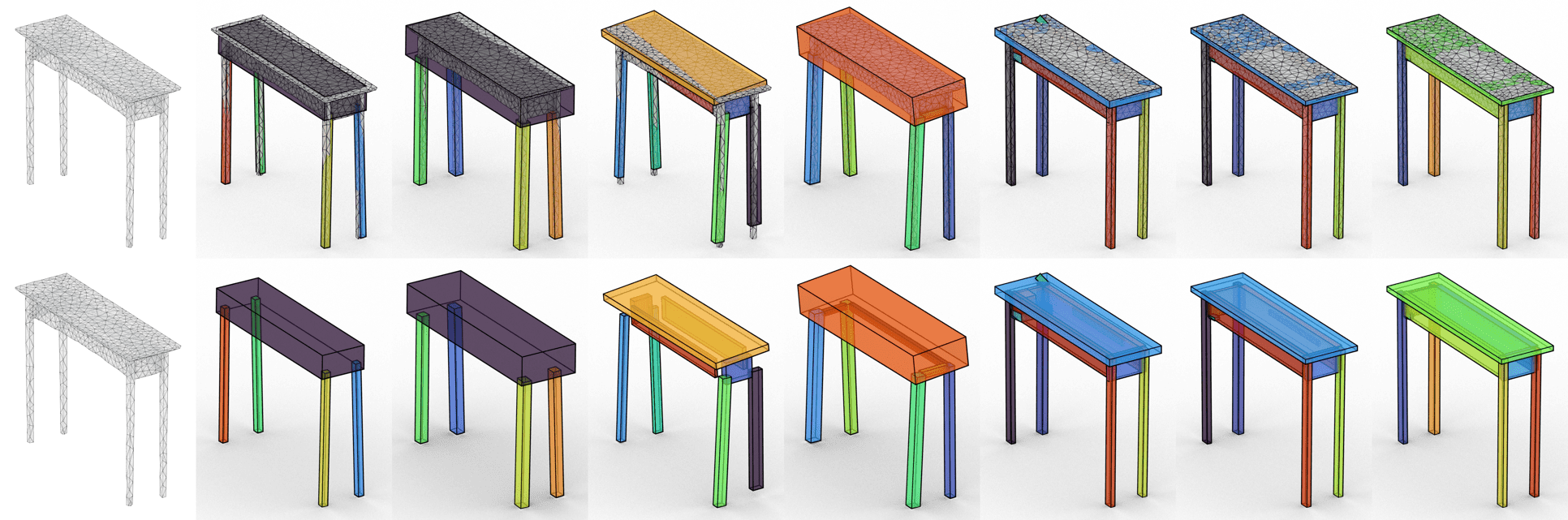}
\Image{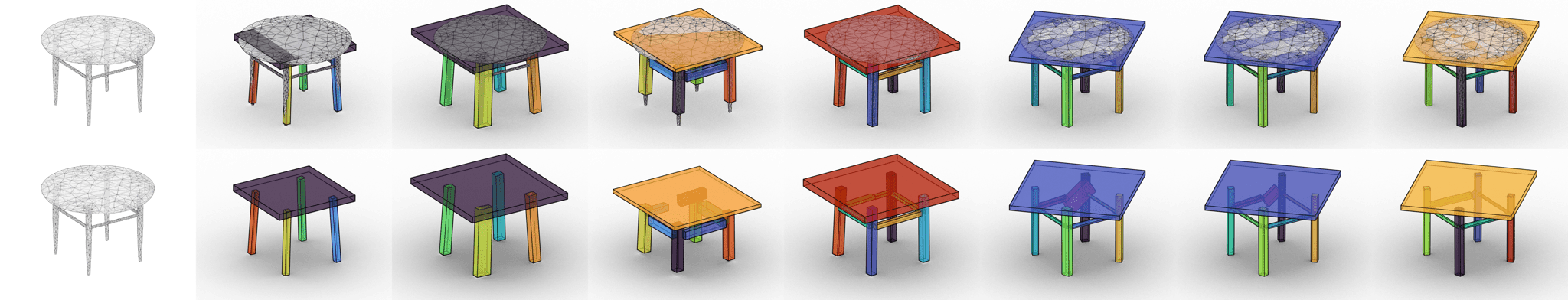}
\Image{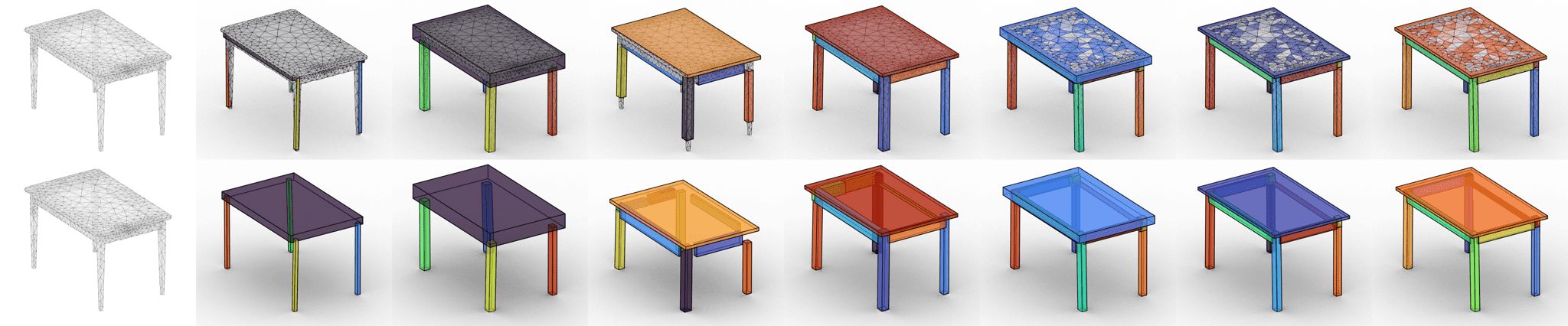}
\Image{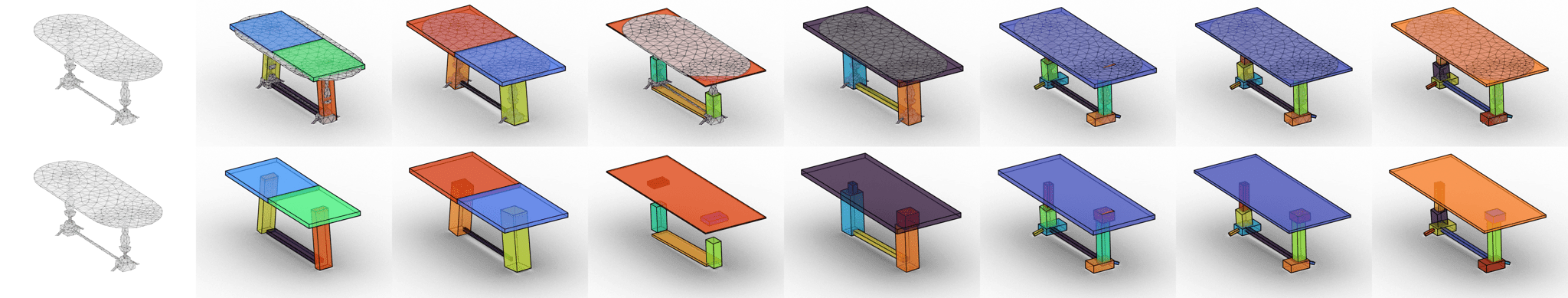}
\Image{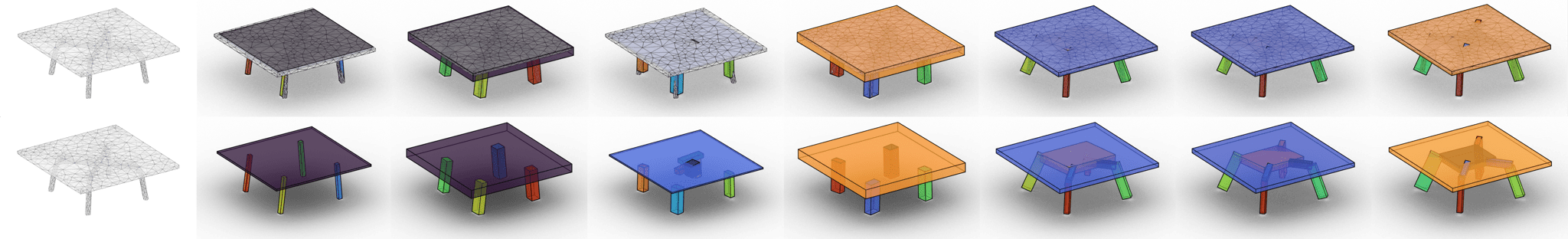}
\Image{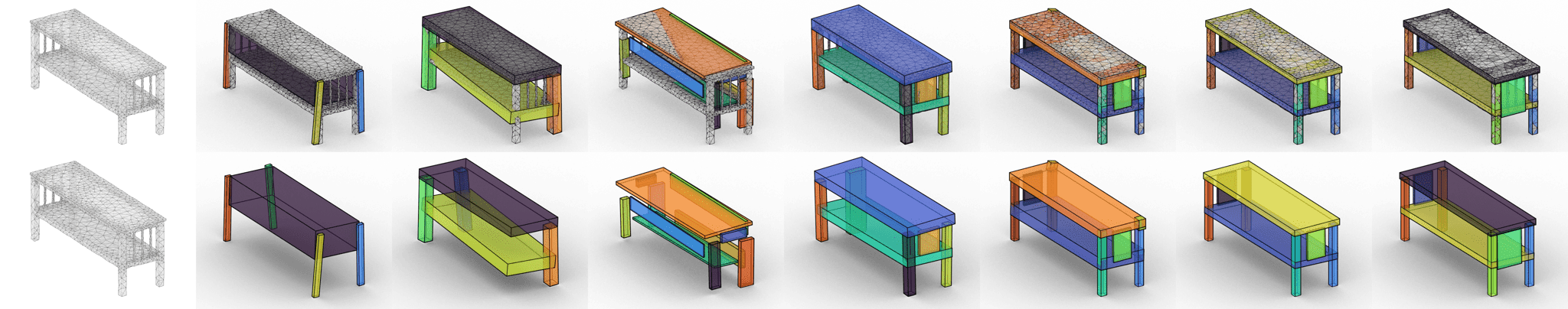}
\Image{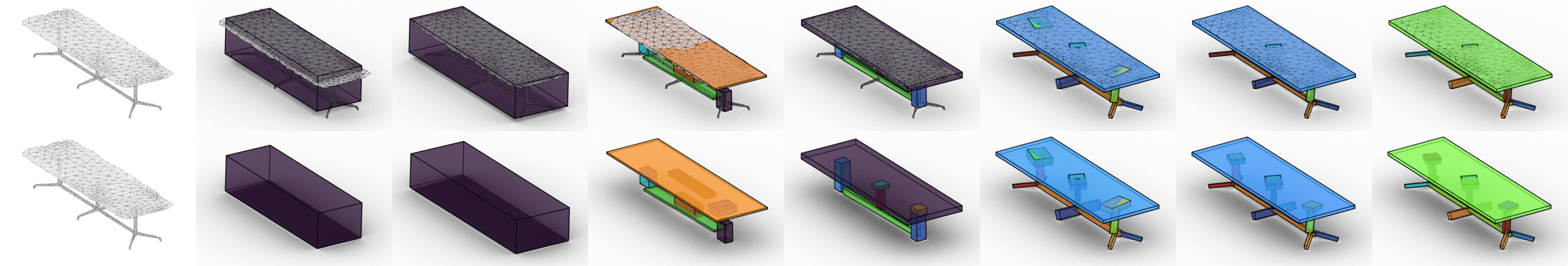}
\Image{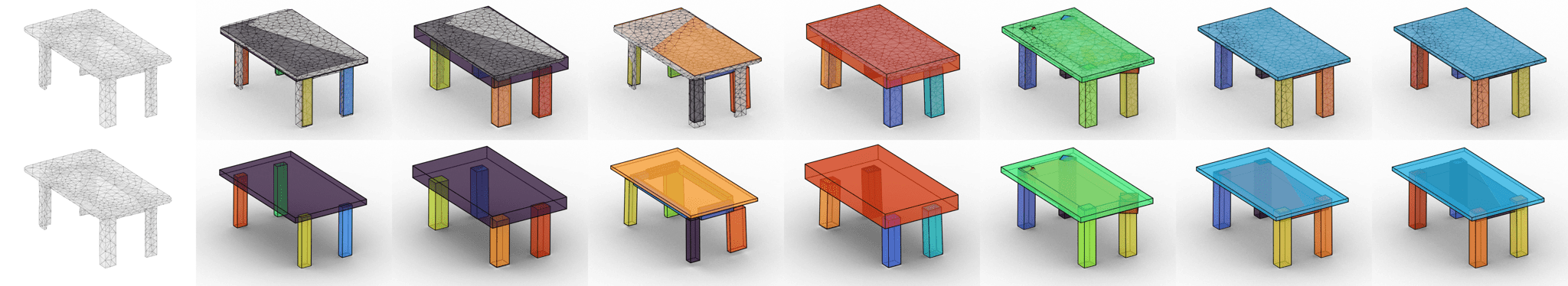}
\Image{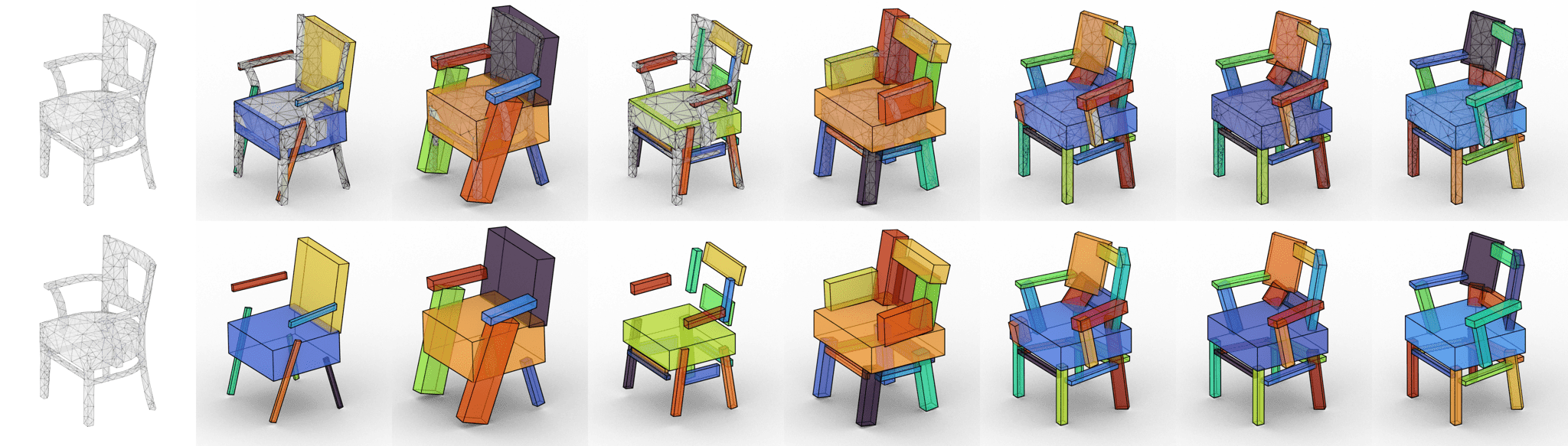}
\Image{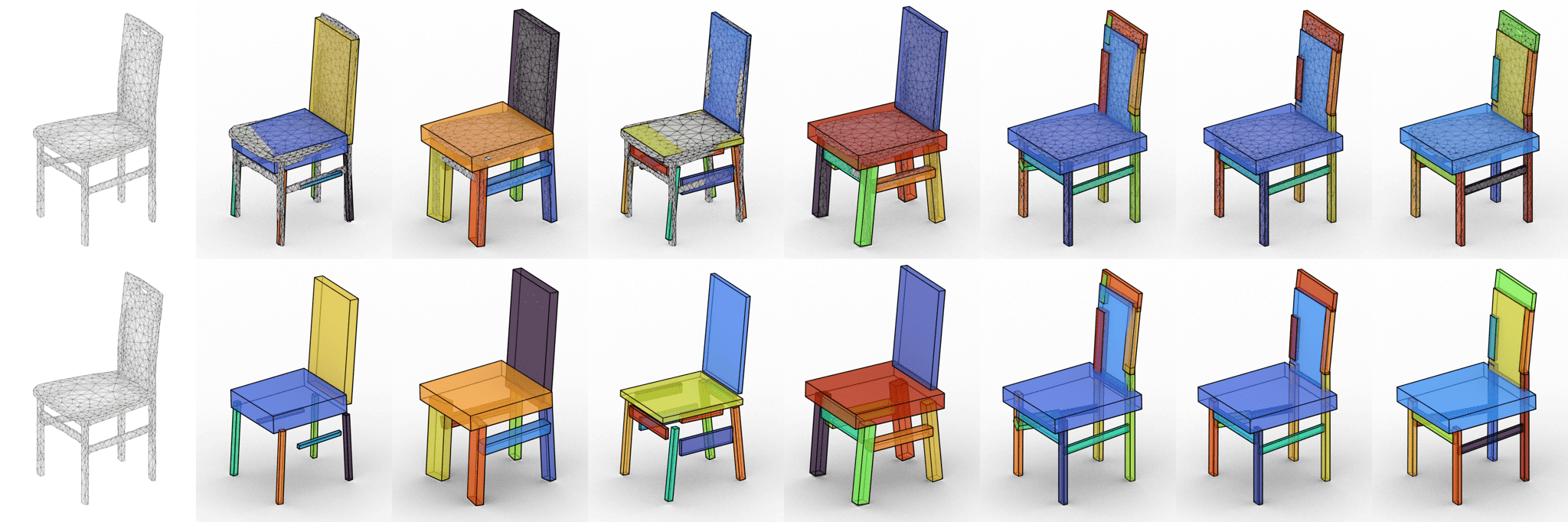}
\Image{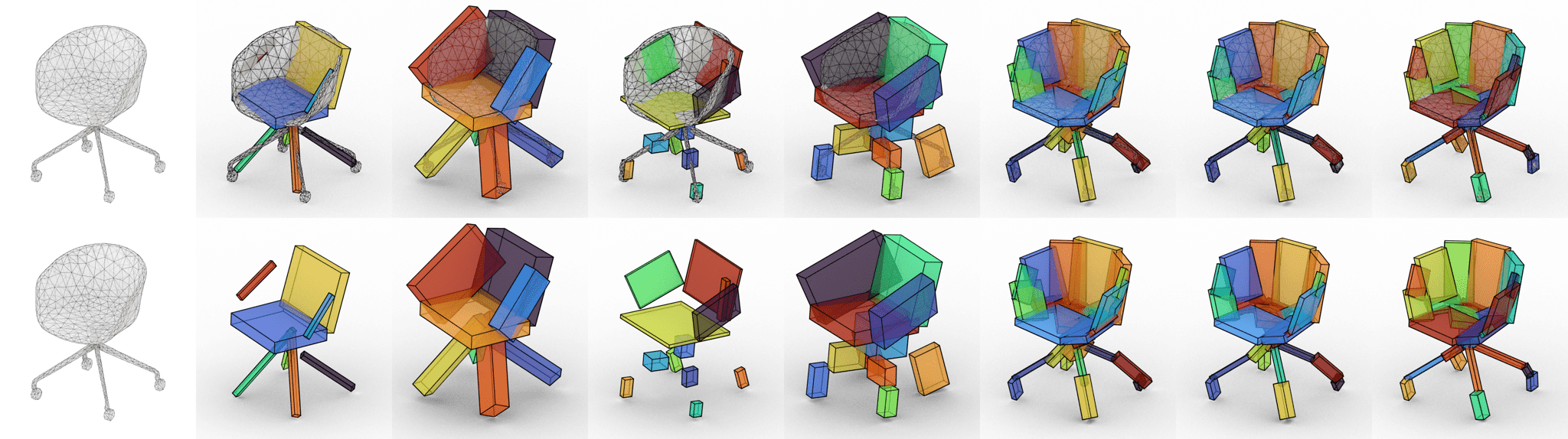}
\Image{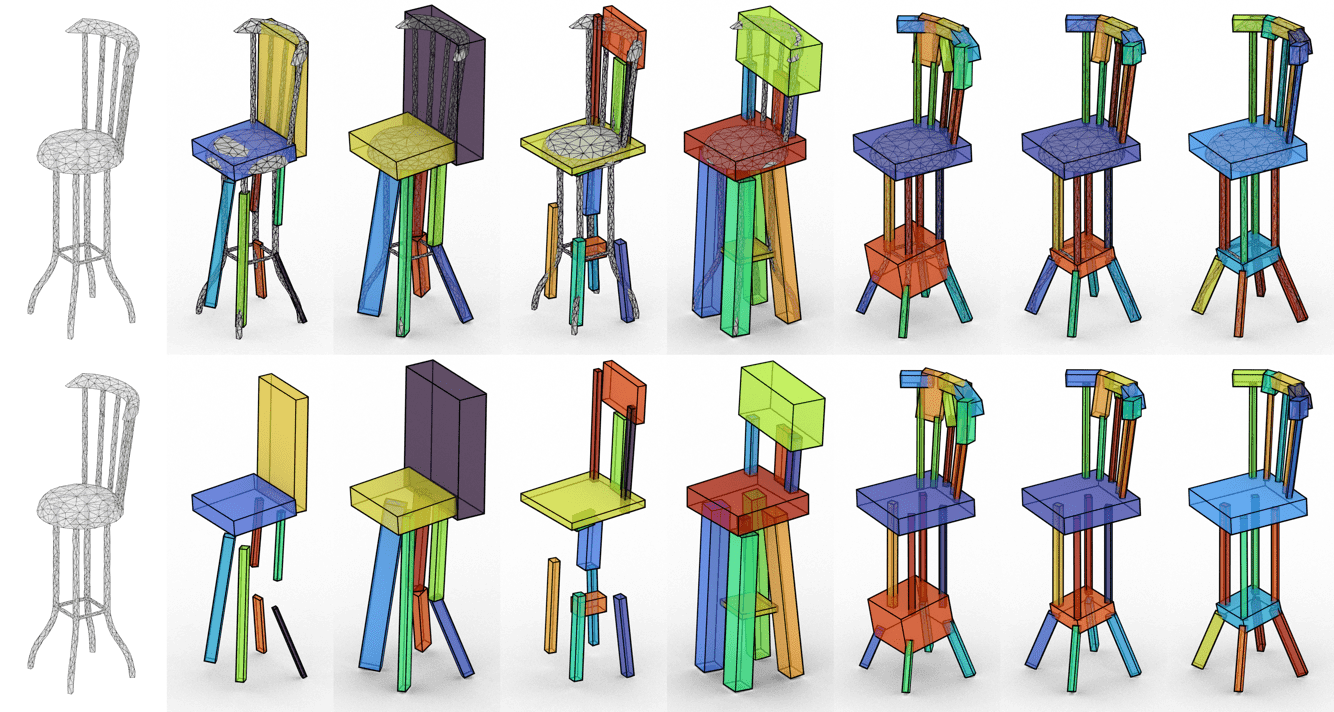}
\Image{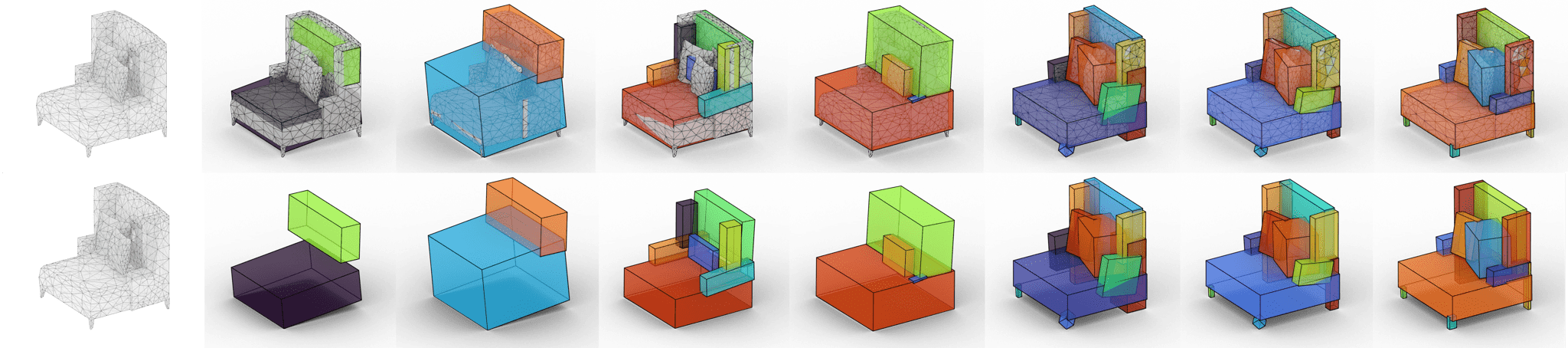}
\Image{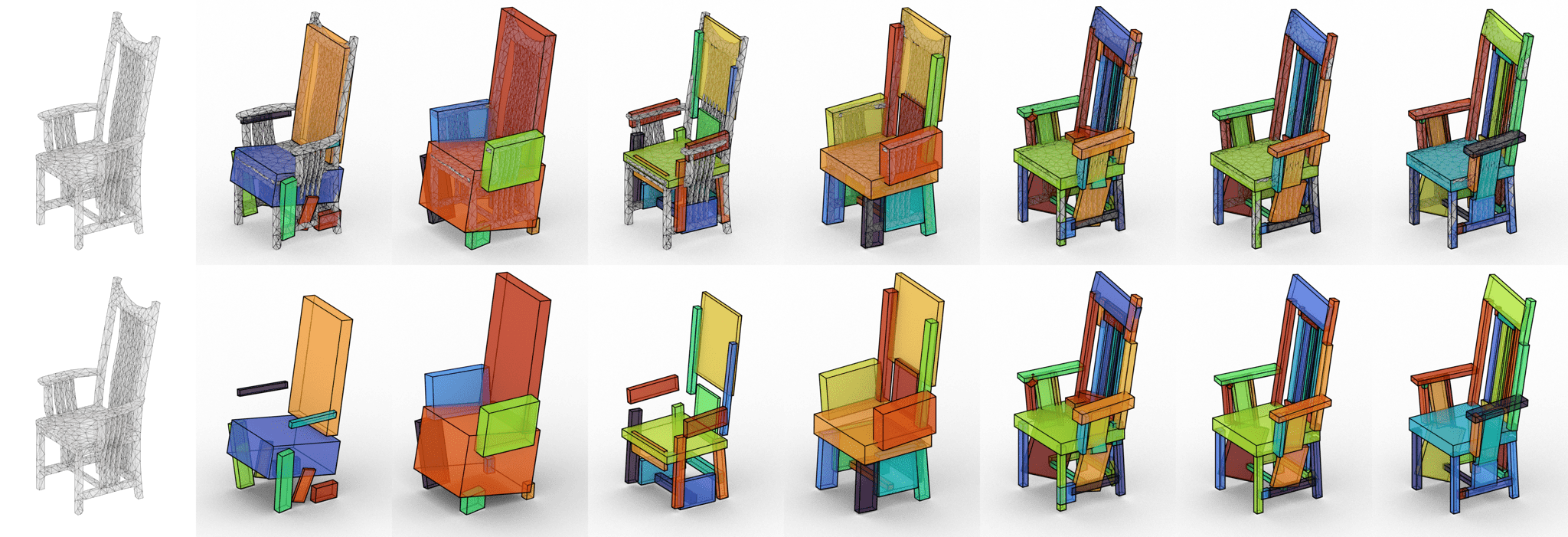}
\Image{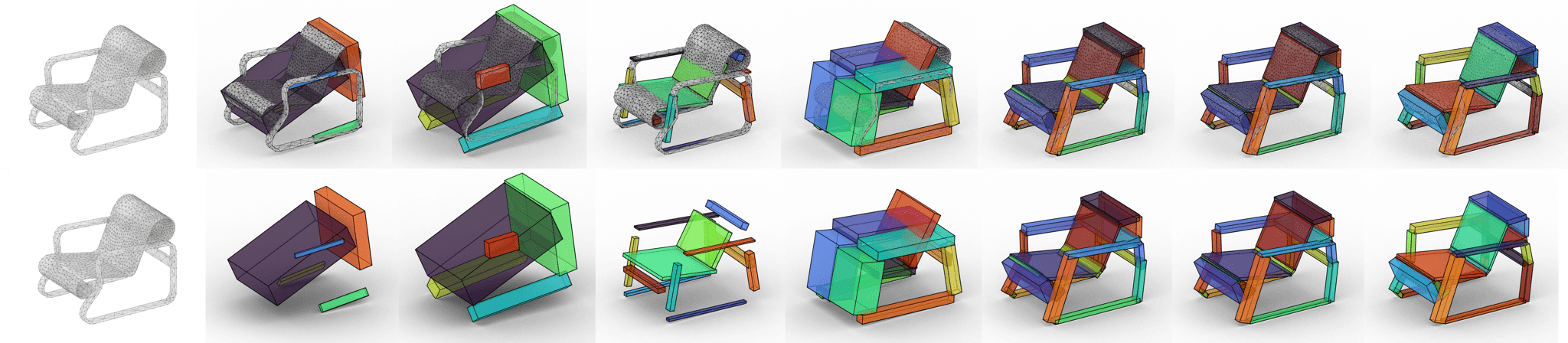}
\Image{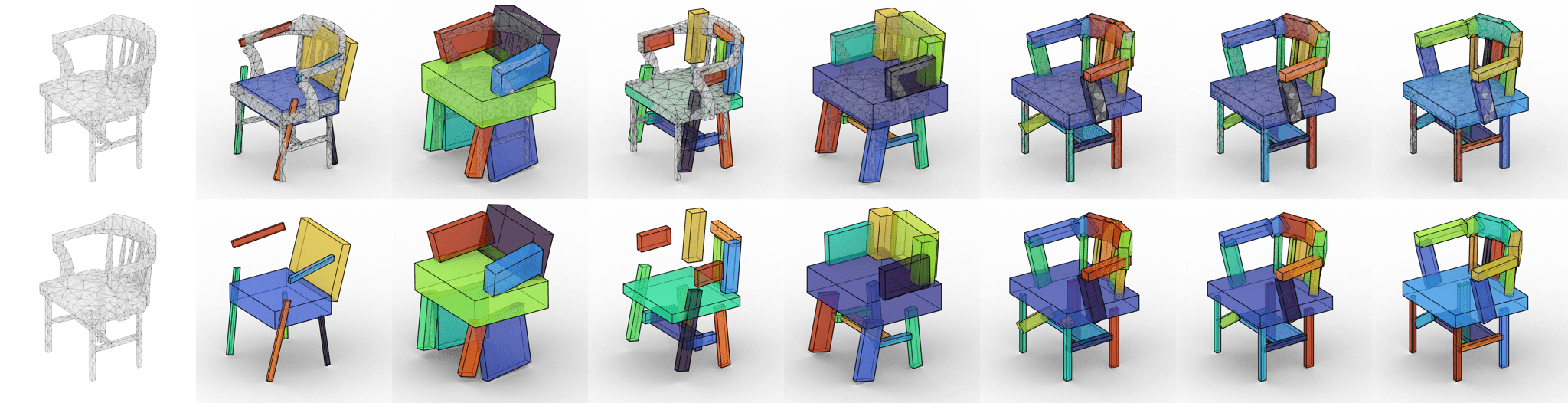}
\Image{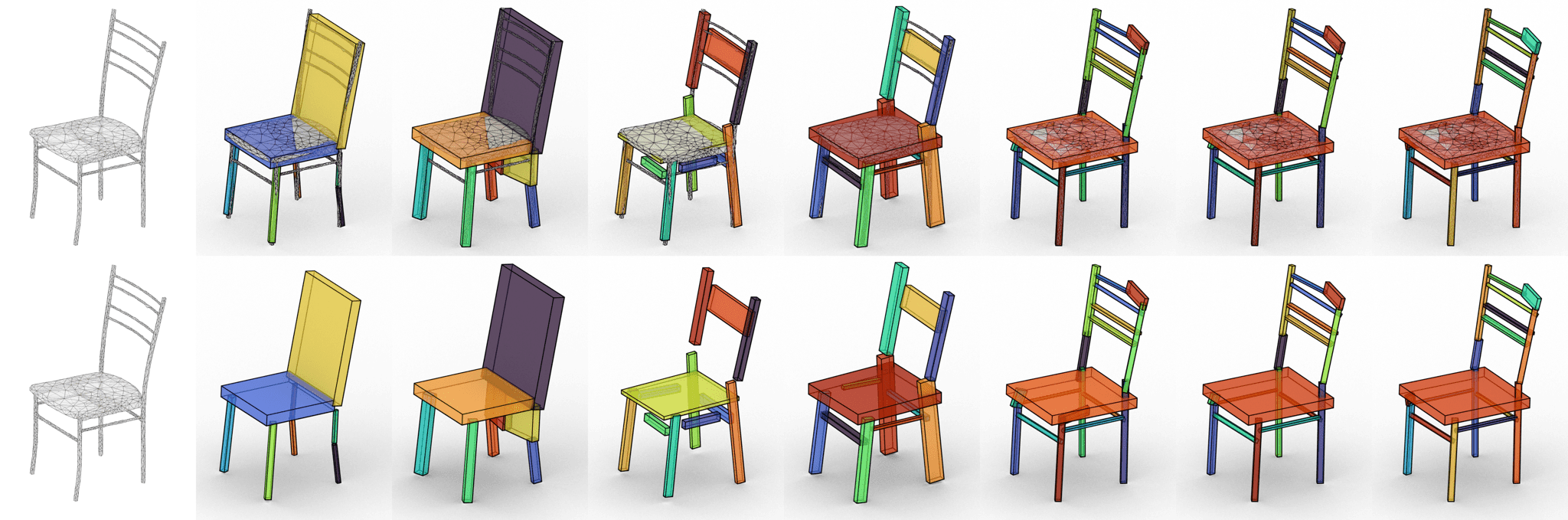}
\Image{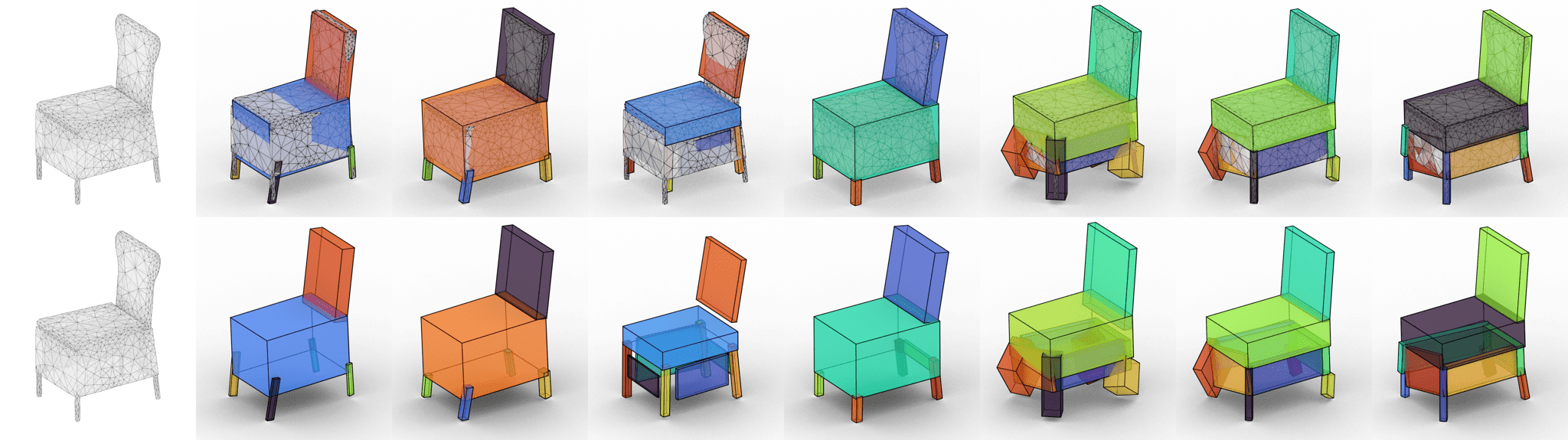}
\Image{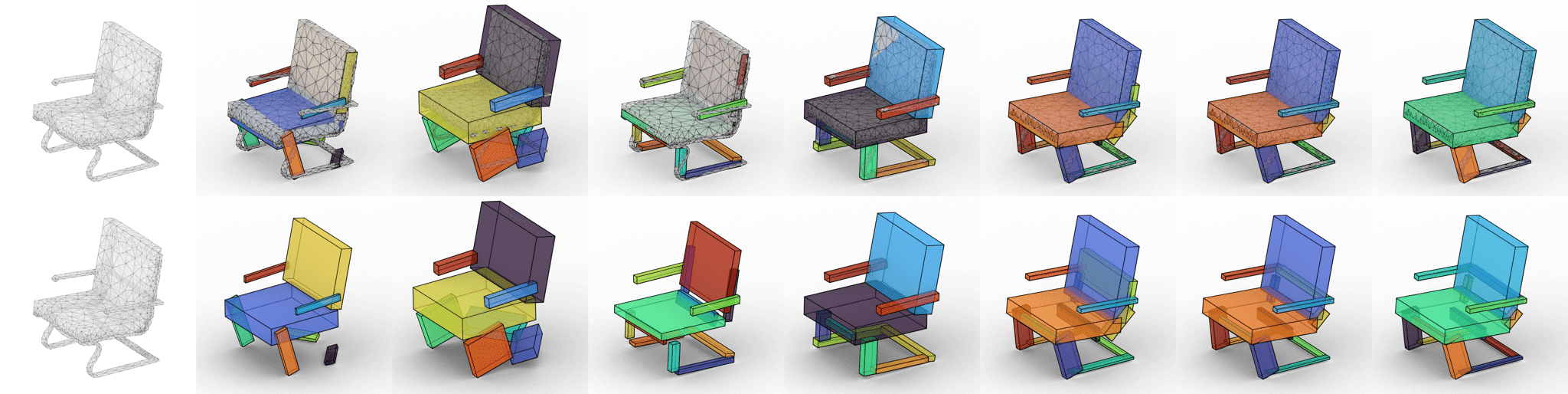}
\Image{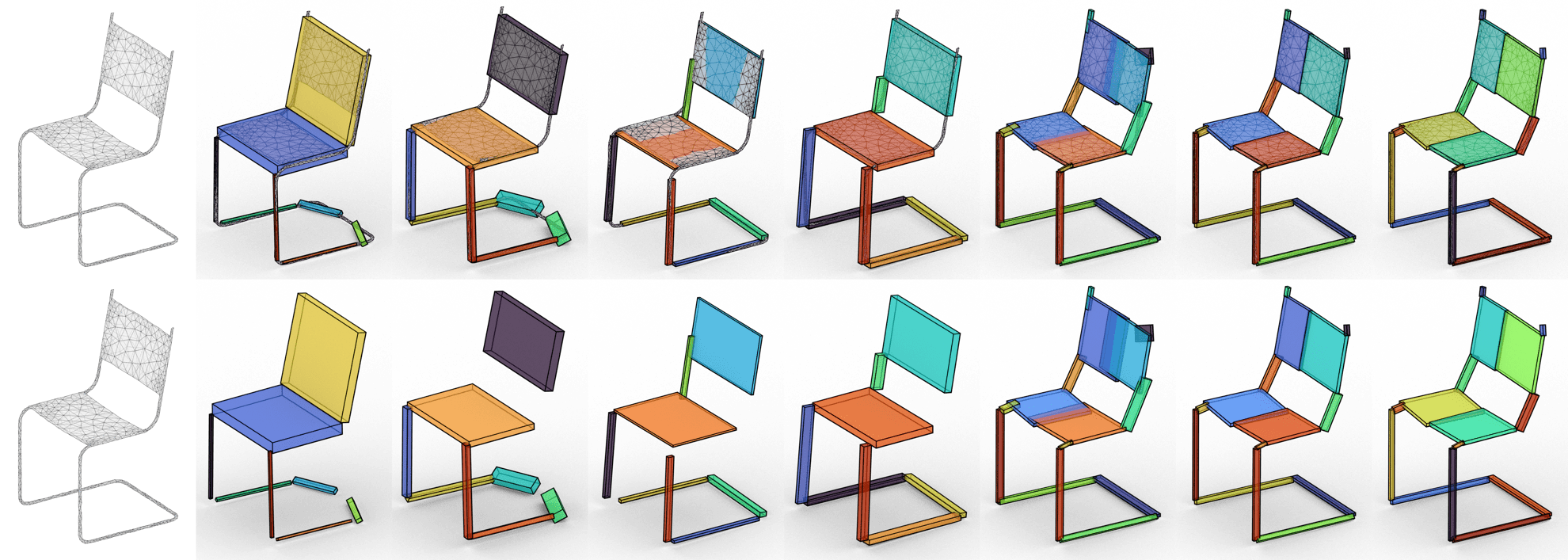}
\Image{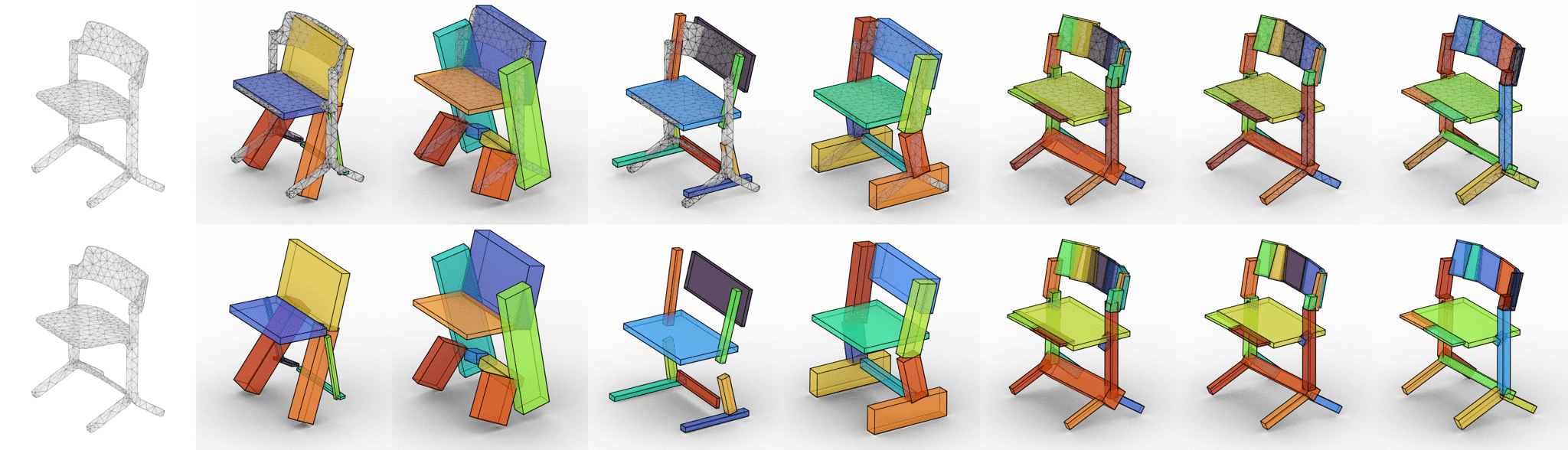}
\Image{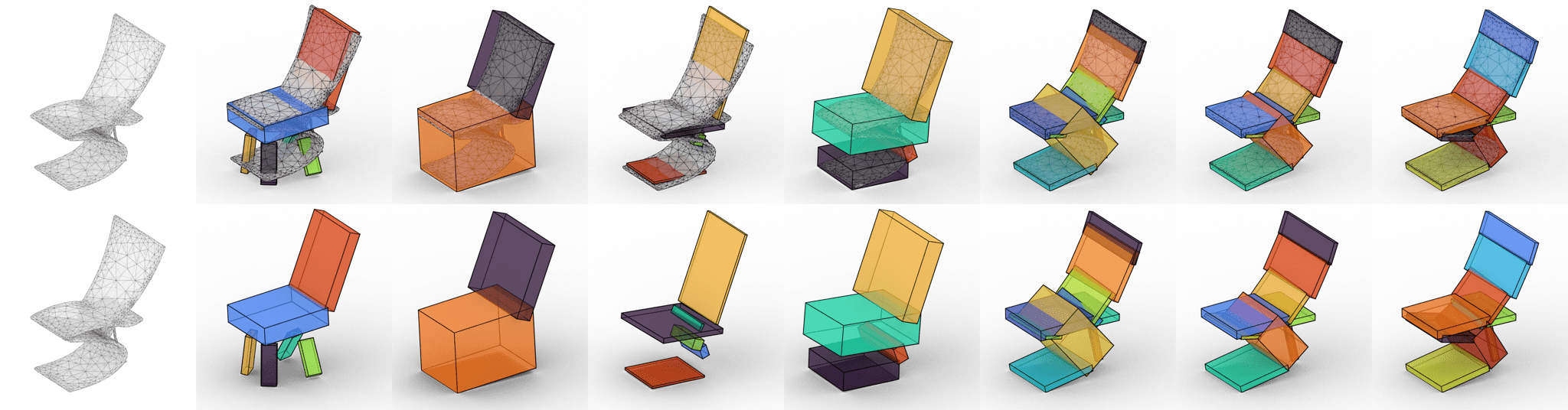}
\Image{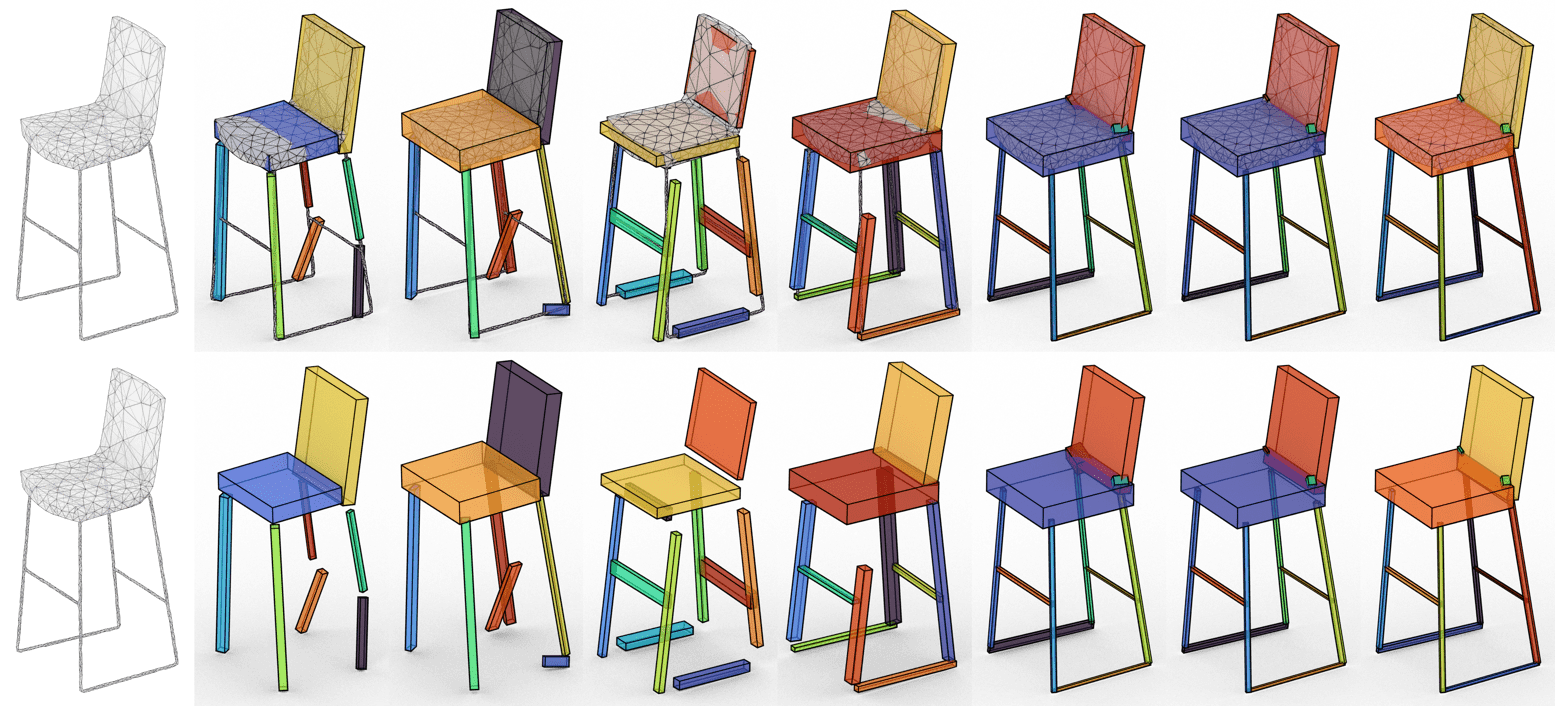}
\Image{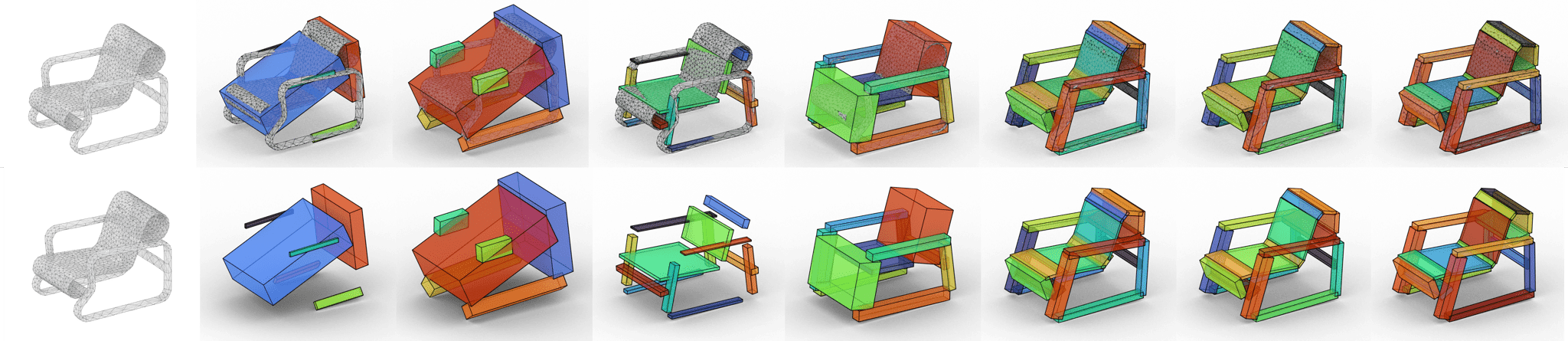}
\Image{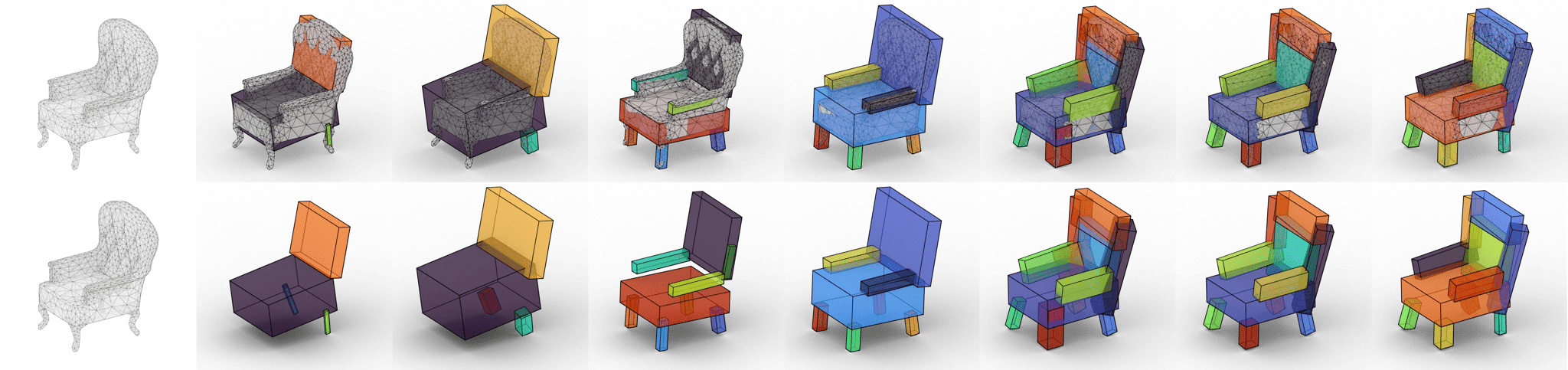}
\Image{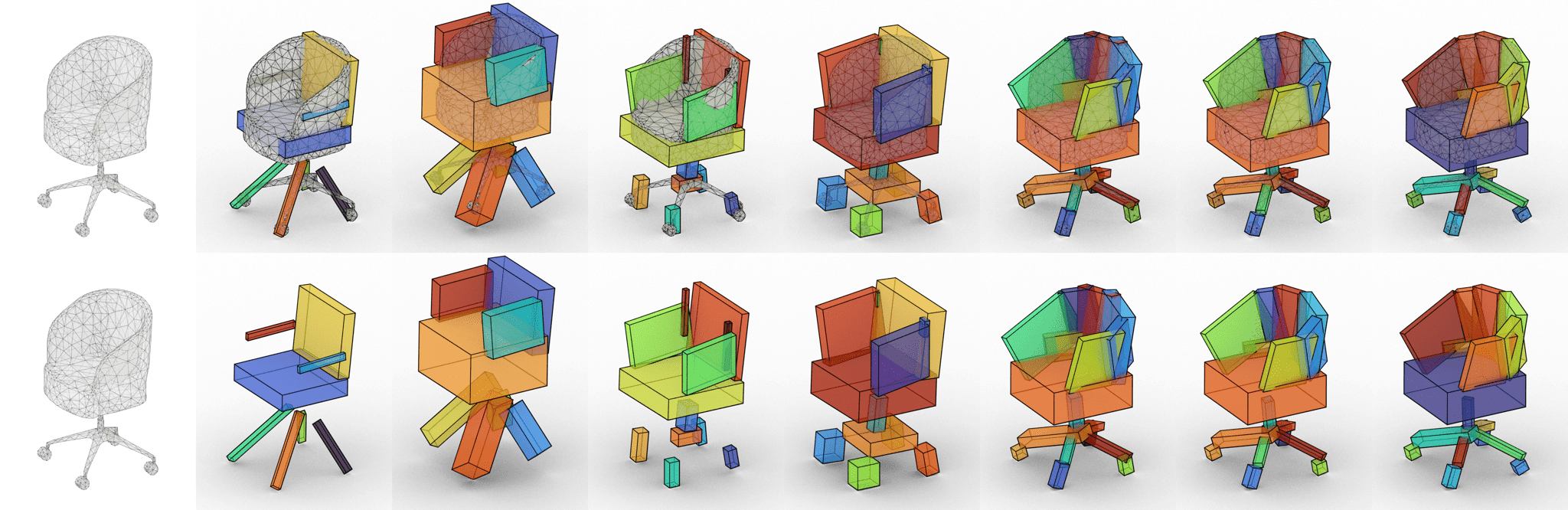}
\Image{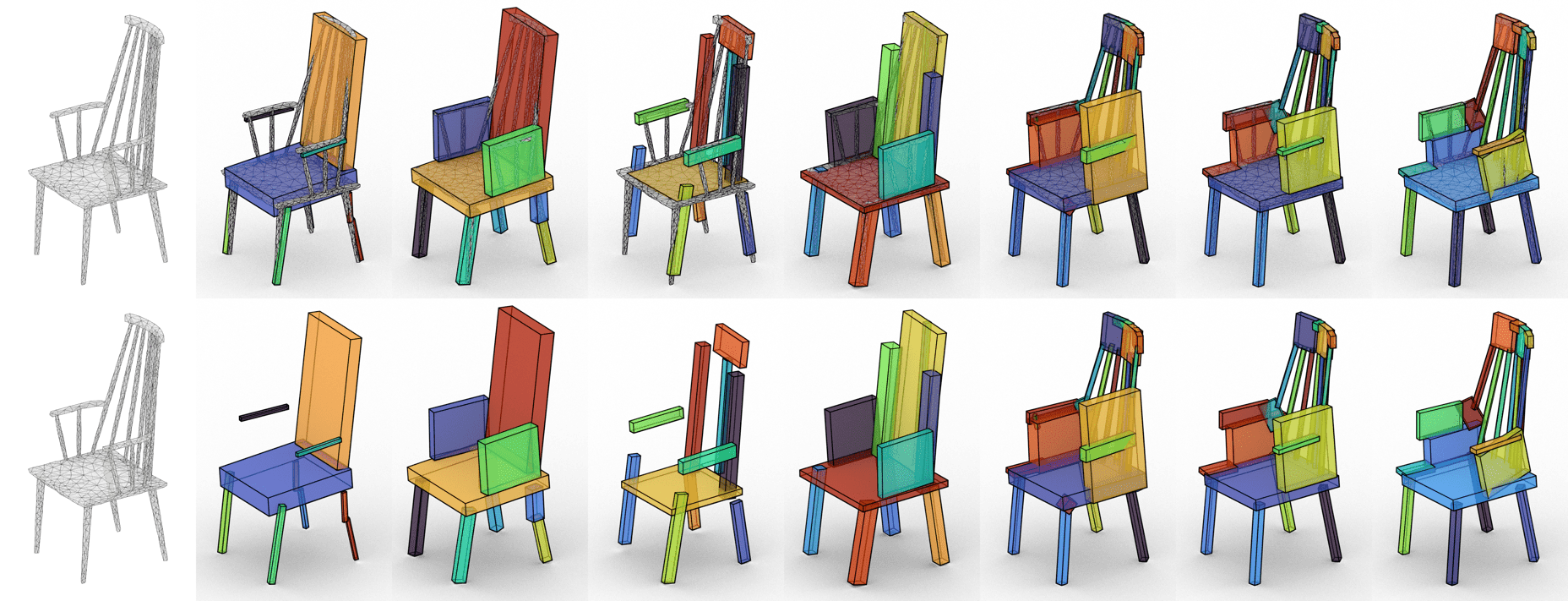}
\Image{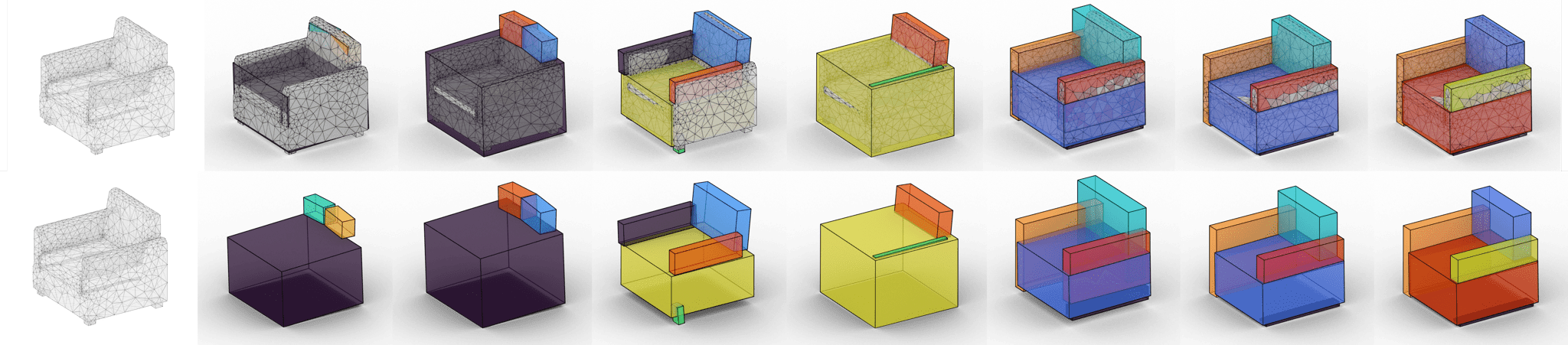}
\Image{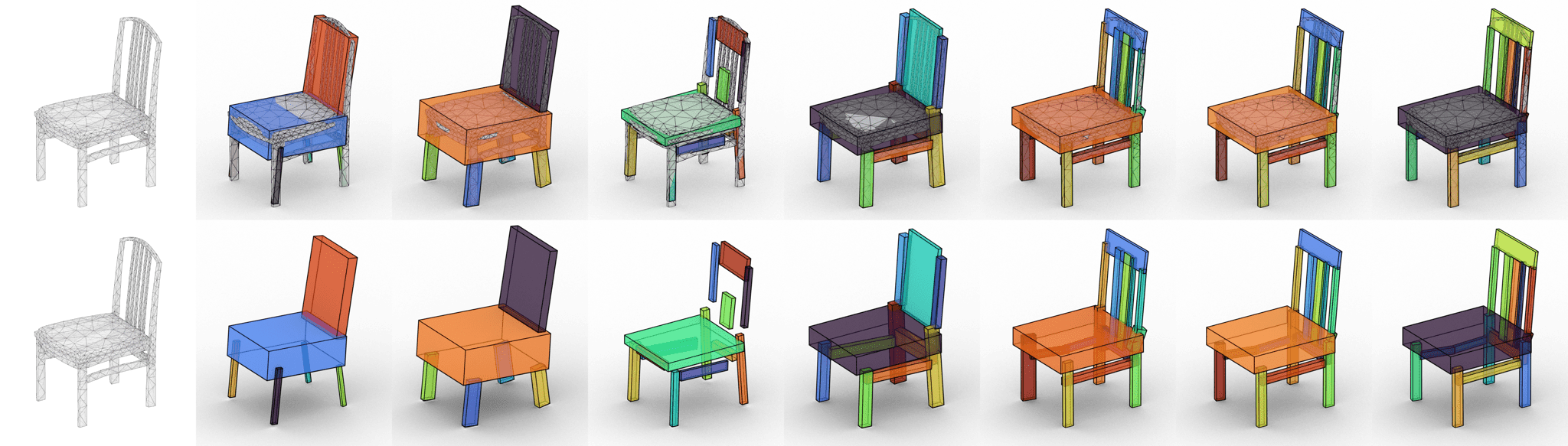}
\Image{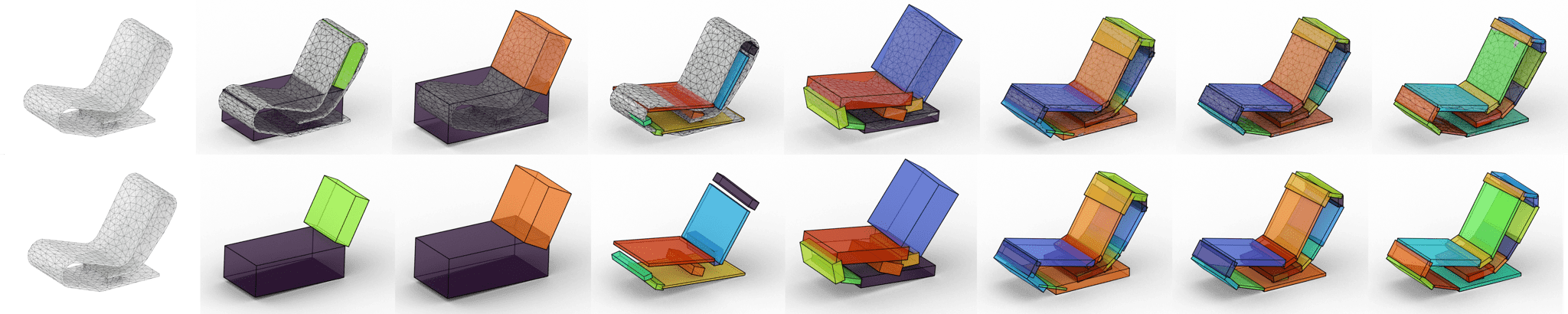}
\Image{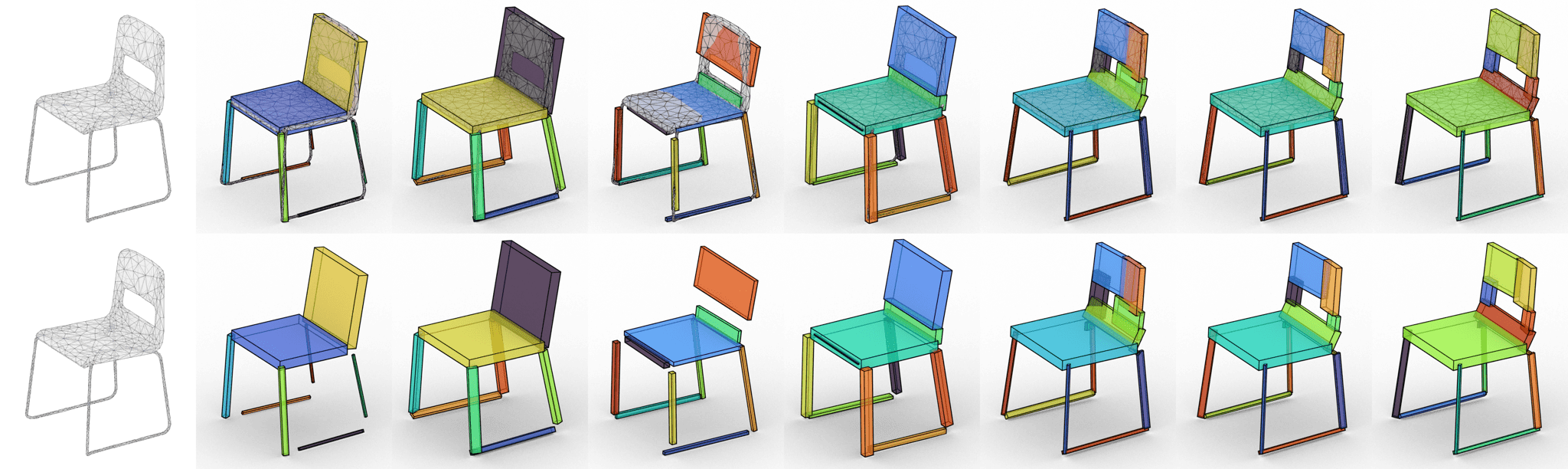}
\Image{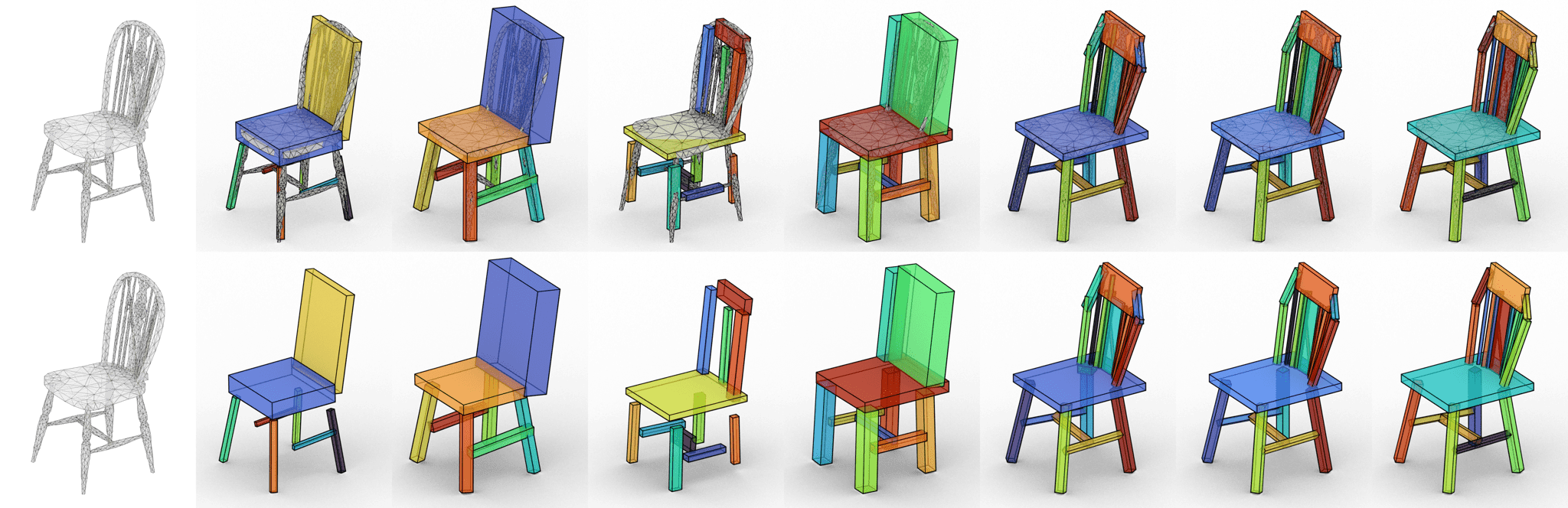}
\Image{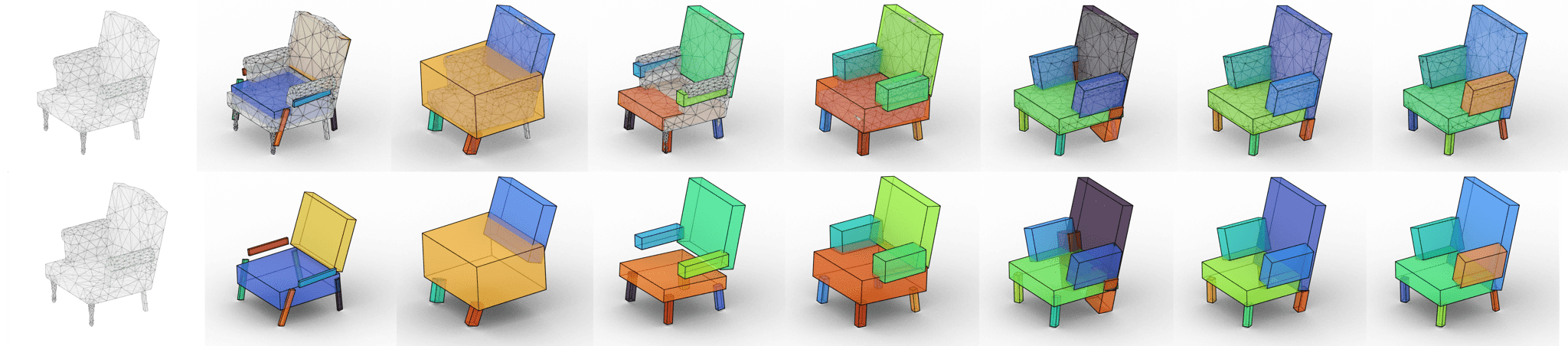}
\Image{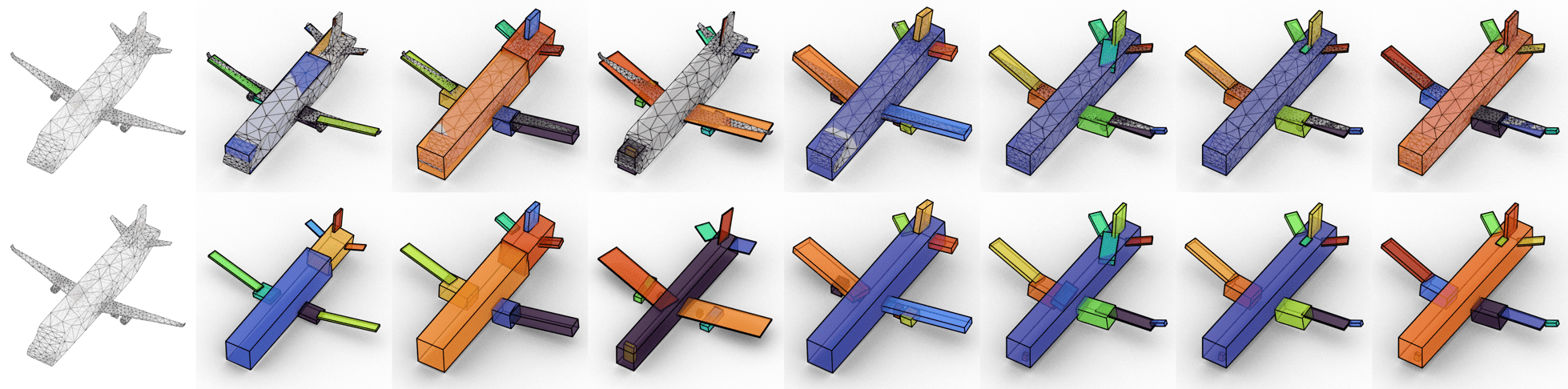}
\Image{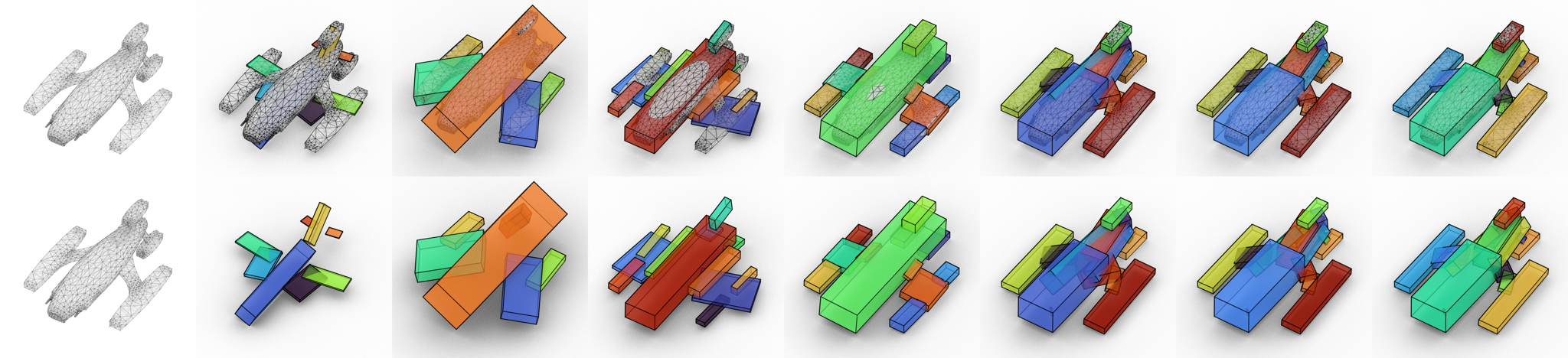}
\Image{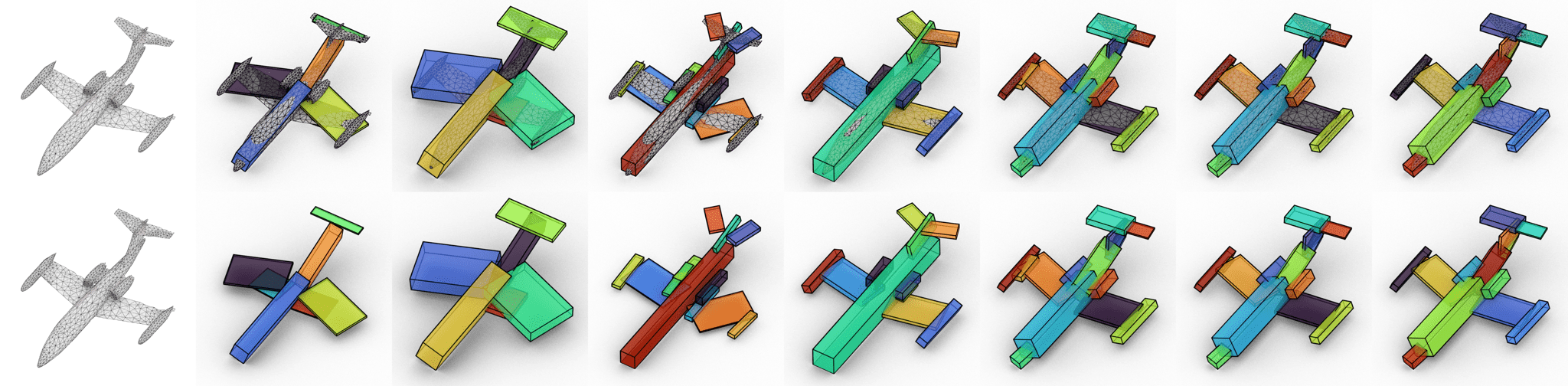}
\Image{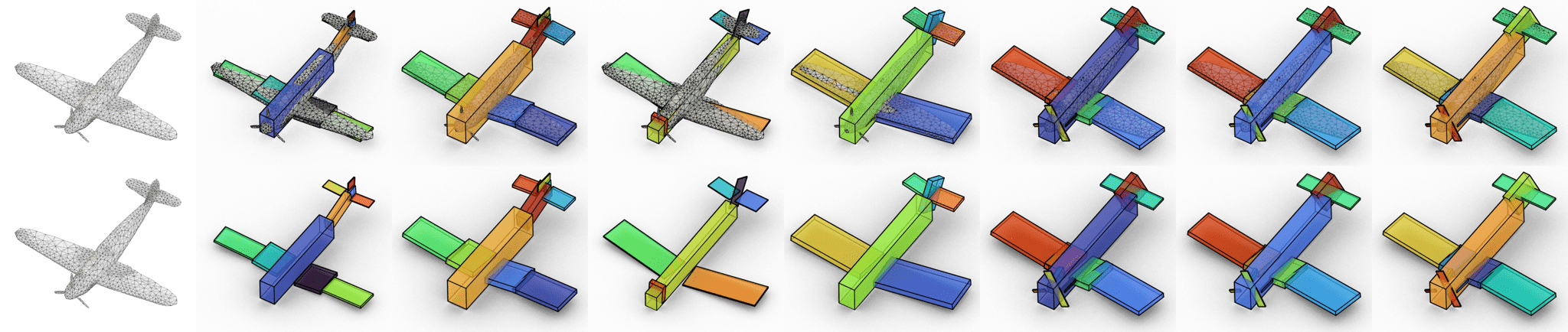}
\Image{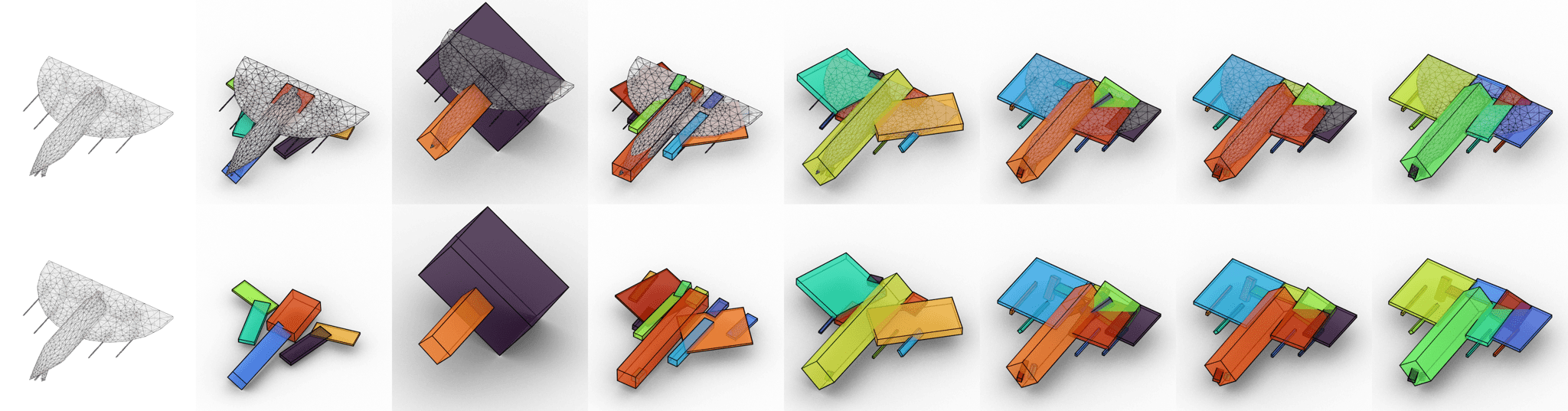}
\Image{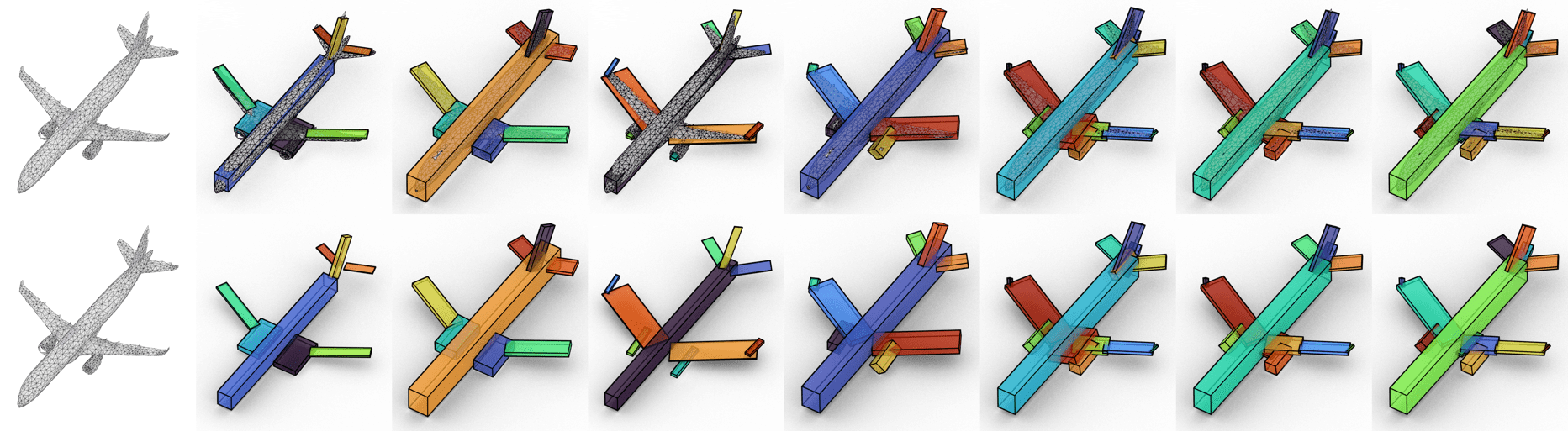}
\Image{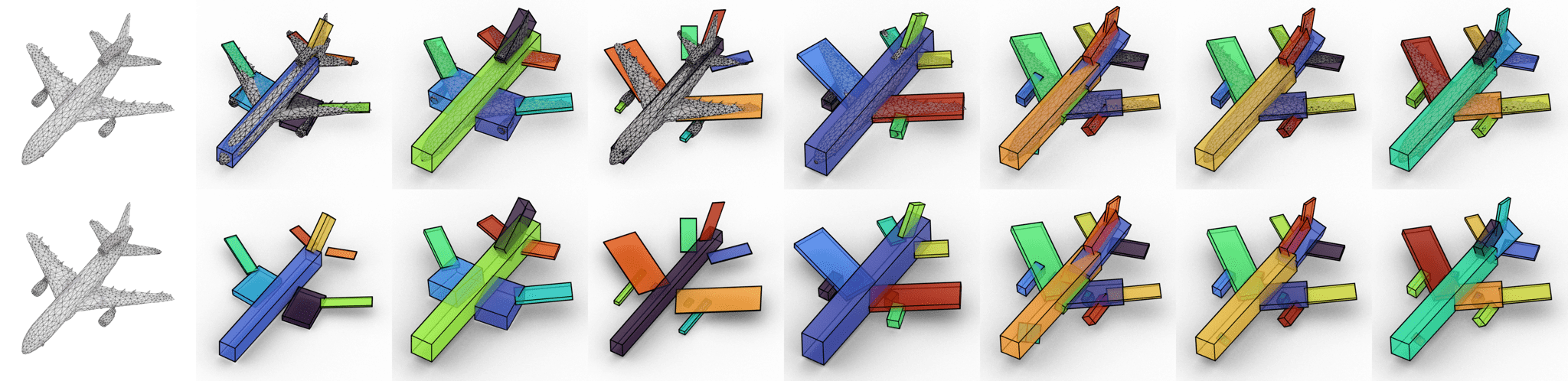}
\Image{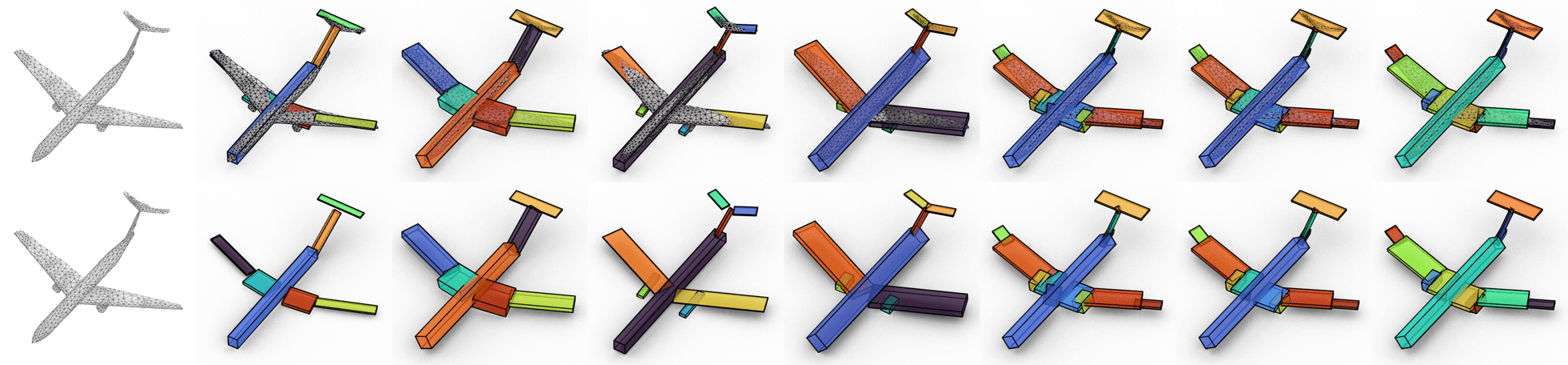}
\Image{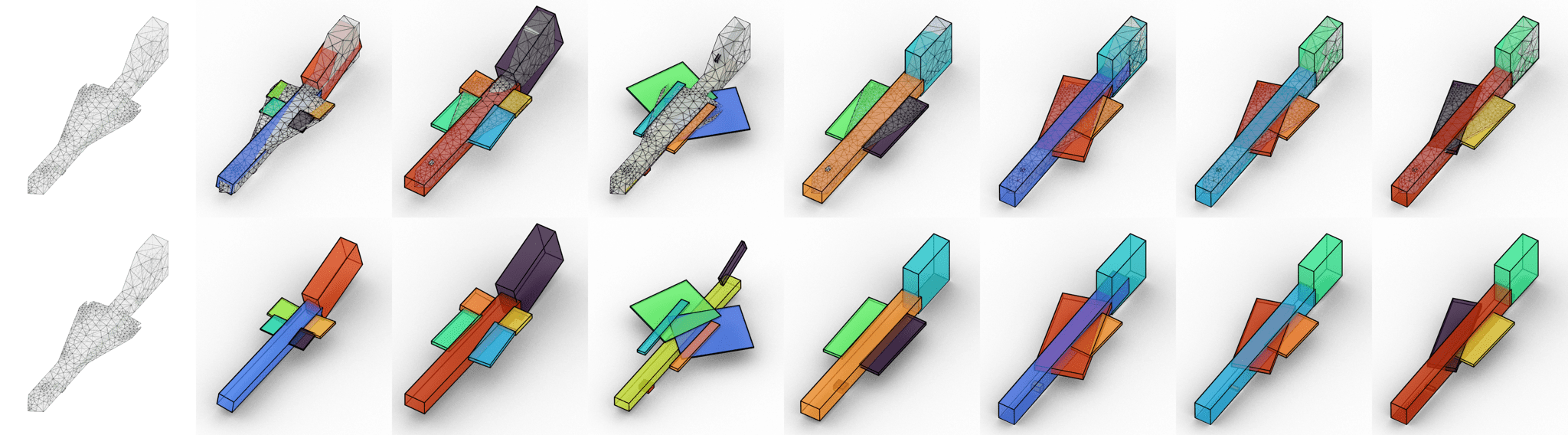}
\Image{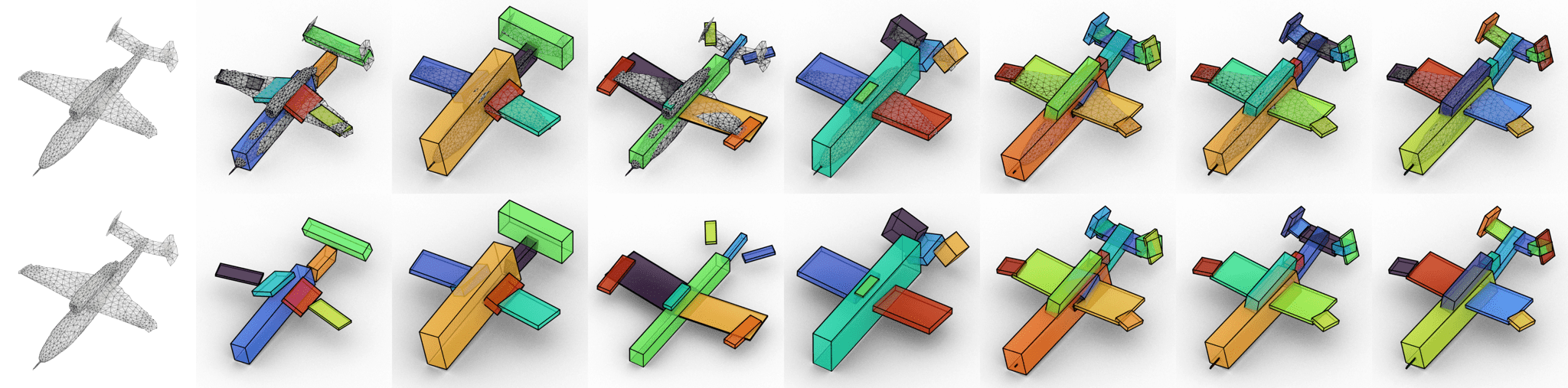}
\Image{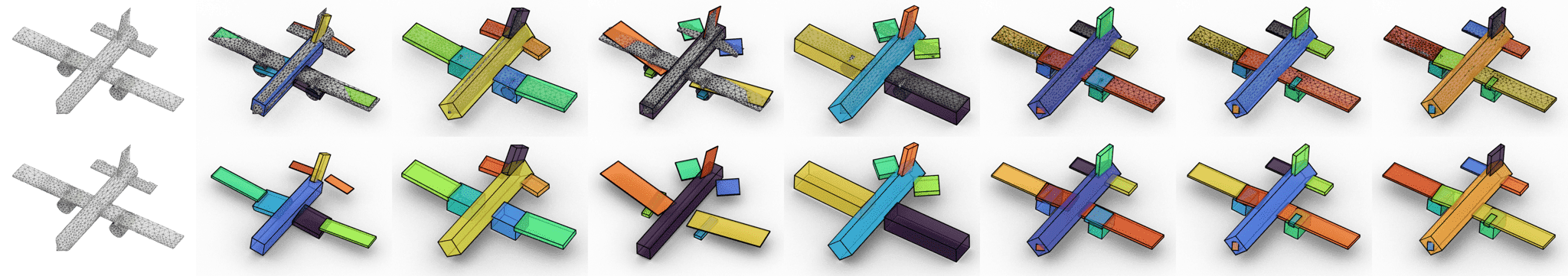}
\Image{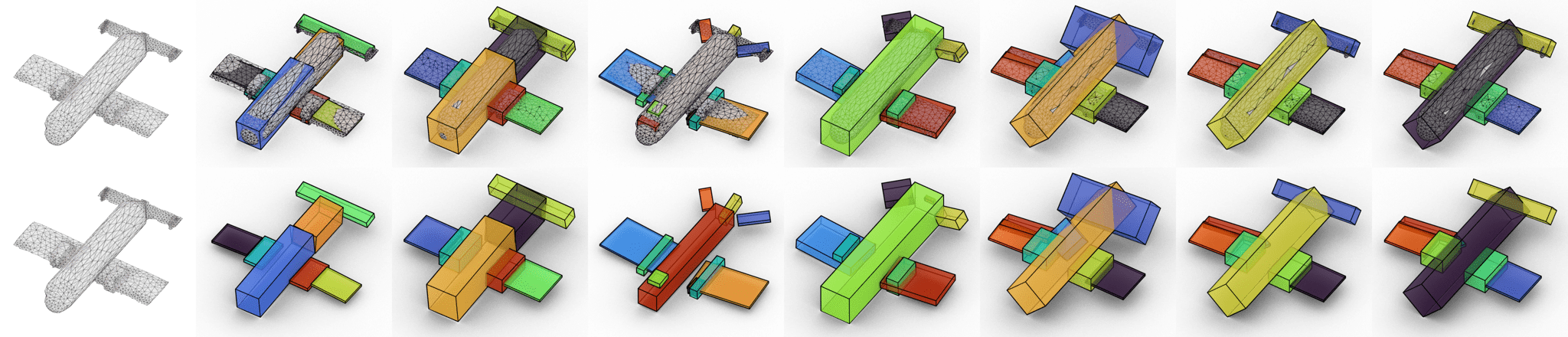}
\Image{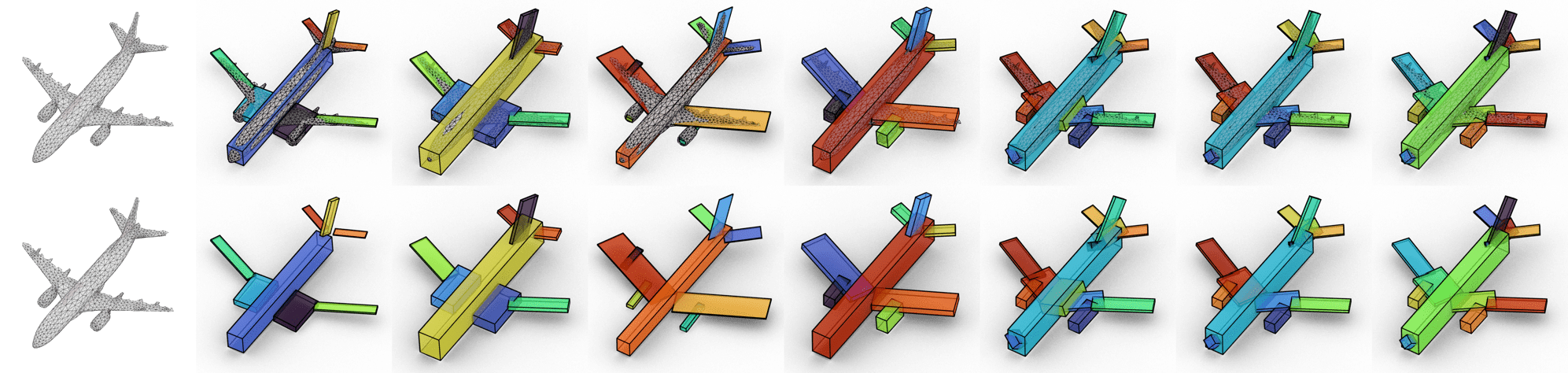}
\Image{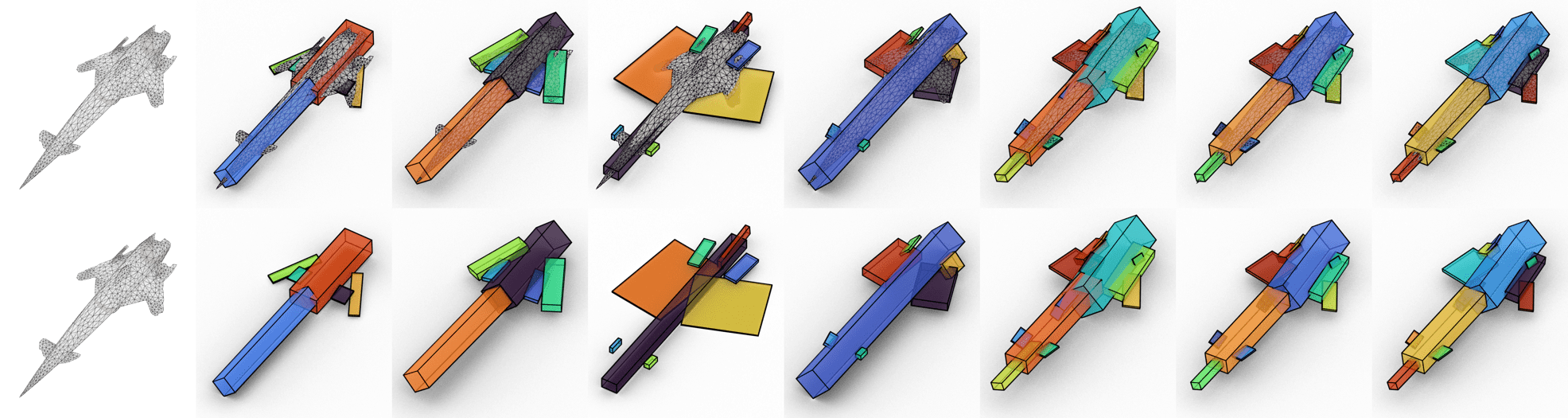}
\Image{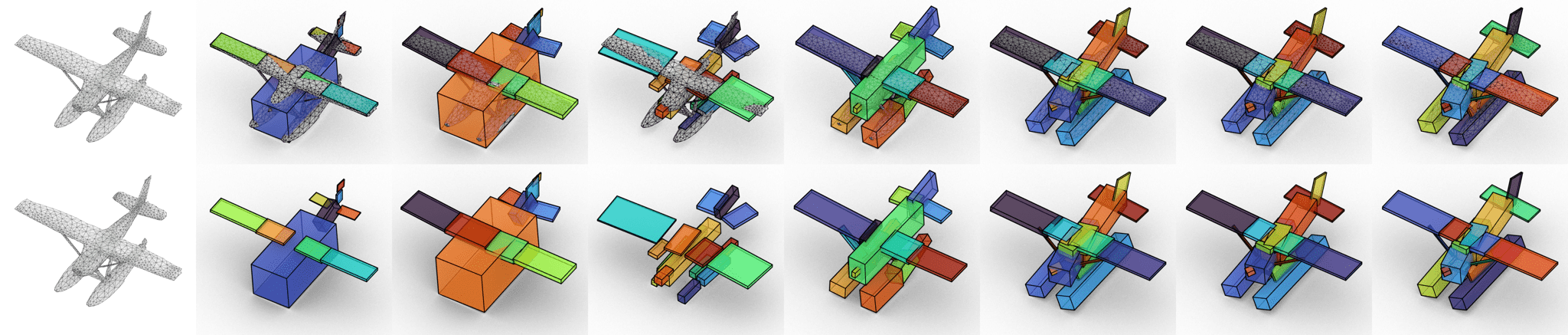}
\Image{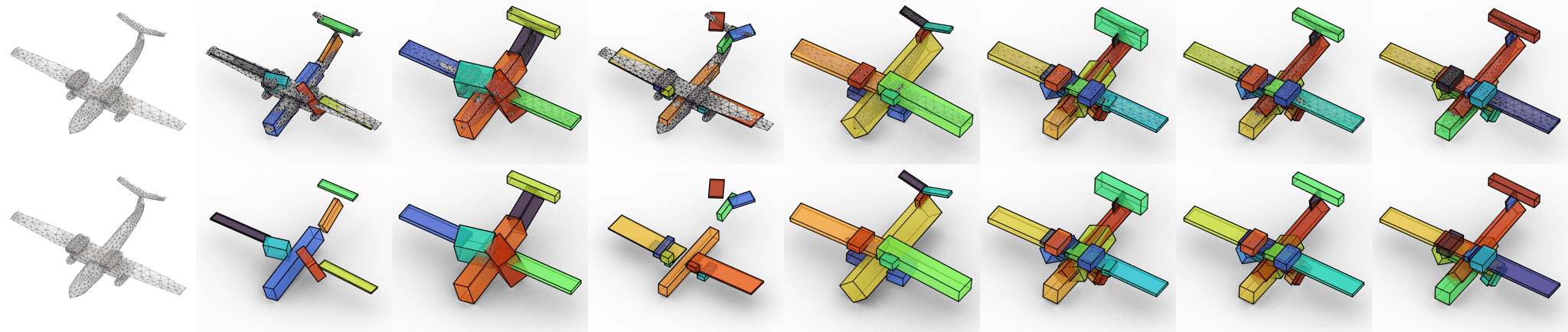}
\Image{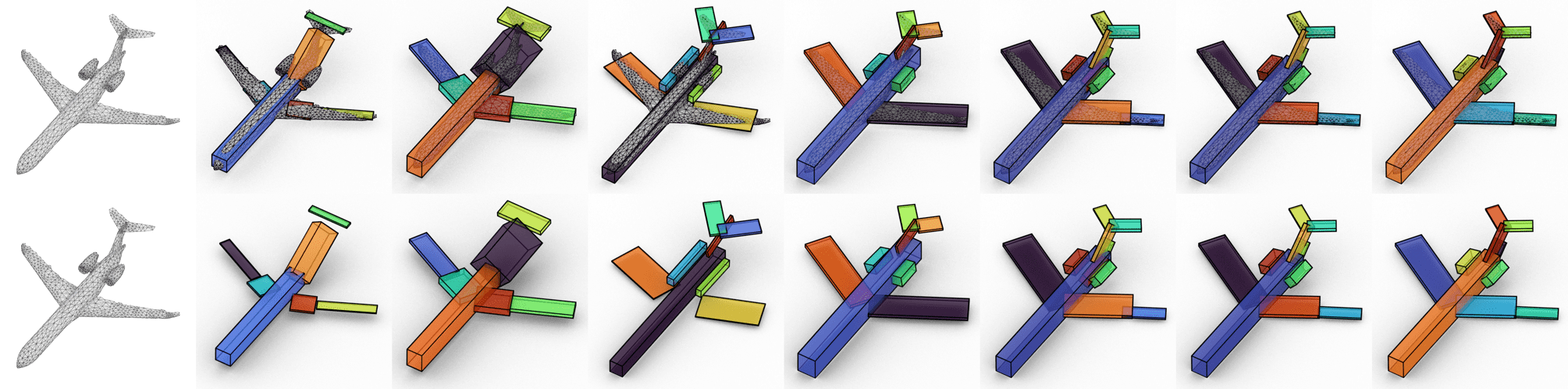}
\Image{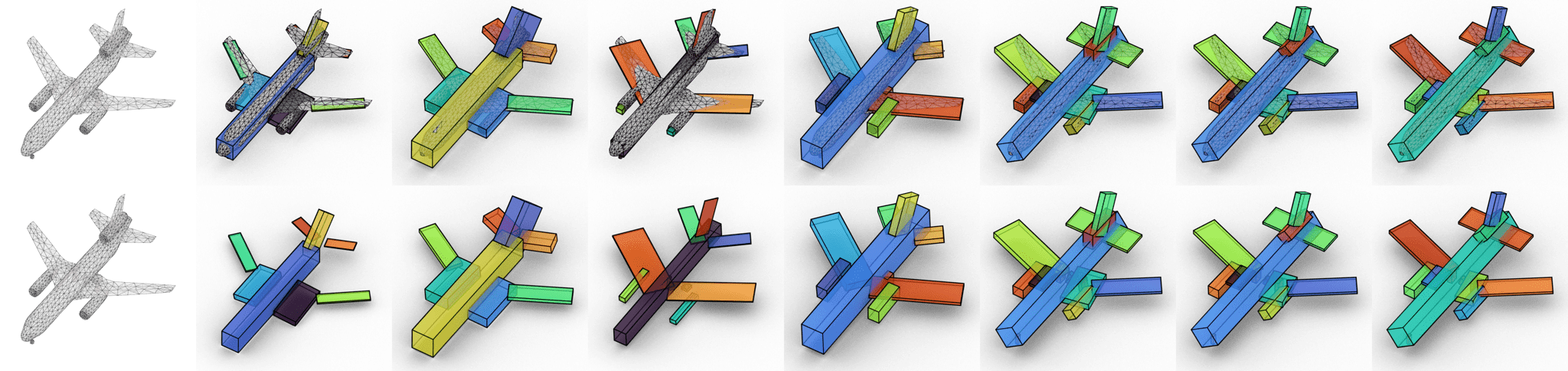}